%% file: casado-sca2022.tex
\documentclass{egpubl}
\usepackage{sca2022}

\SpecialIssuePaper         %

\usepackage[T1]{fontenc}
\usepackage{dfadobe}  

\usepackage{cite}  %
\BibtexOrBiblatex
\electronicVersion
\PrintedOrElectronic
\ifpdf \usepackage[pdftex]{graphicx} \pdfcompresslevel=9
\else \usepackage[dvips]{graphicx} \fi

\usepackage{egweblnk} 
\usepackage{upgreek}
\usepackage{mathtools}
\usepackage{amsfonts}
\usepackage{subcaption}
\usepackage{todonotes}
\title[PERGAMO: Personalized 3D Garments from Monocular Video]%
      {PERGAMO: Personalized 3D Garments from Monocular Video}

\author[A. Casado-Elvira, M. Comino Trinidad, \& D.Casas]
{\parbox{\textwidth}{\centering 
		Andrés Casado-Elvira \orcid{0000-0003-0021-3930
}~~~~~~~~~~~~~~
		Marc Comino Trinidad \orcid{0000-0001-5621-7565}~~~~~~~~~~~~~~   
		Dan Casas \orcid{0000-0002-3664-089X}
	}
	\\
	{\parbox{\textwidth}{\centering Universidad Rey Juan Carlos, Madrid, Spain.}
	}
}

\begin{document}

\definecolor{amethyst}{rgb}{0.6, 0.4, 0.8}
\definecolor{darkpastelgreen}{rgb}{0.01, 0.75, 0.24}
\definecolor{amber}{rgb}{1.0, 0.75, 0.0}
\definecolor{cadmiumorange}{rgb}{0.93, 0.53, 0.18}
\definecolor{lawngreen}{rgb}{0.49, 0.99, 0.0}
\definecolor{limegreen}{rgb}{0.2, 0.8, 0.2}
\definecolor{neongreen}{rgb}{0.22, 0.88, 0.08}
\definecolor{indigo}{rgb}{0.098, 0.008, 0.51}
\definecolor{rust}{rgb}{0.67, 0.235, 0.035}

\newcommand{\TODO}[1]{\textcolor{red}{[\textbf{TODO}: {#1}]}} 
\newcommand{\Change}[1]{#1}
\newcommand{\Dan}[1]{\textcolor{neongreen}{[\textbf{Dan}: {#1}]}} 
\newcommand{\Marc}[1]{\textcolor{cadmiumorange}{[\textbf{Marc}: {#1}]}}
\newcommand{\Andres}[1]{\textcolor{rust}{[\textbf{Andrés}: {#1}]}}
\newcommand{\TextMarc}[1]{\textcolor{cadmiumorange}{{#1}}}

\teaser{
 \includegraphics[width=1.0\linewidth]{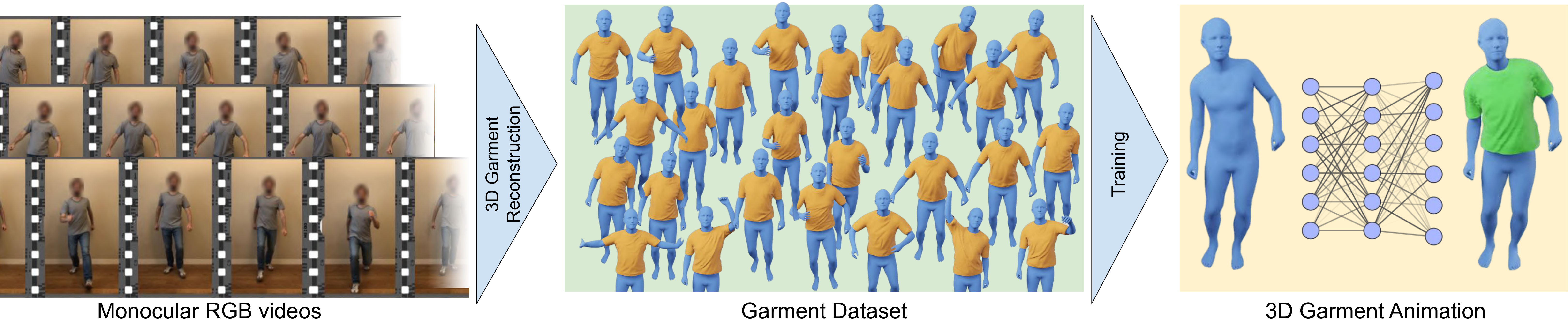}
 \centering
  \caption{Using just small a collection of monocular videos captured with a mobile phone (left), we propose a novel approach to reconstruct the 3D garment layer with fine-scale details (center), and use them to learn a deformable model of the captured apparel (right). Our learned model enables to animate the captured garment, exhibiting material-specific details, as a function of the underlying body pose.}
\label{fig:teaser}
}

\maketitle
\begin{abstract}
  Clothing plays a fundamental role in digital humans.
  Current approaches to animate 3D garments are mostly based on realistic physics simulation, however, they typically suffer from two main issues: high computational run-time cost, which hinders their deployment; and simulation-to-real gap, which impedes  the synthesis of specific real-world cloth samples.
  To circumvent both issues we propose PERGAMO, a data-driven approach to learn a deformable model for 3D garments from monocular images.
  To this end, we first introduce a novel method to reconstruct the 3D geometry of garments from a single image, and use it to build a dataset of clothing from monocular videos.
  We use these 3D reconstructions to train a regression model that accurately predicts how the garment deforms as a function of the underlying body pose.
  We show that our method is capable of producing garment animations that match the real-world behavior, and generalizes to unseen body motions extracted from motion capture dataset. 


\end{abstract}  
\input{sections/introduction}
\input{sections/related-work}

\input{sections/overview}
\input{sections/reconstruction}

\input{sections/regressor}
\input{sections/results}

\input{sections/conclussion}
\bibliographystyle{eg-alpha-doi}  
\bibliography{casado-sca2022}

\end{document}

%% file: sections/introduction.tex
\section{Introduction}
Synthesizing realistic 3D digital humans is a key topic in Computer Animation due to the large number of applications in industries such as video games, films, telecommunications, and more.
A fundamental challenge in digital humans is the modeling of dynamic clothing and garments, because it requires solutions capable of representing large diversity of designs, complex materials, and highly deformable surfaces.

The most well-established approach to model clothing is physics-based simulation, which is today a mature field capable of generating highly-detailed results \cite{narain2012arcsim,stuyck2018cloth,cirio2014yarn}.
However, simulation methods suffer from two main difficulties that hinder their deployment in plug-and-play use cases: 
first, the significant run-time computational cost, which hinders achieving real-time frame rates.
Current solutions for fast simulation need to simplify dynamics \cite{bender2014survey} or to adopt custom GPU solutions \cite{tang2018gpu,pcloth20}; 
and second, the simulation-to-real gap, which prevents to synthesize simulated cloth that behaves exactly as a desired real-world target.
Current methods attempt to fit simulation parameters to real-world observations, but require controlled setups and are limited to small patches of fabrics
\cite{miguel2012data}.

Alternatives to physics-based cloth methods have been proposed to \textit{independently} address these shortcomings. 
On one hand, impressive advances in data-driven methods based on neural networks \cite{santesteban2019virtualtryon,patel2020tailor,chentanez2020cnn} can highly-efficiently infer garment deformations for parametric avatars, generating much faster animations compared to physics simulations.
However, training data is usually obtained with simulation, hence the simulation-to-real gap remains. 
On the other hand, accurate 3D reconstruction methods \cite{pons2017clothcap,xiang2021modeling} can recover highly-detailed real-world garments, which can be potentially be used to train data-driven approaches. However, the need for multi-camera professional setups prevents to democratize such strategies.
All in all, modeling fast \textit{and} real-world realistic garment deformations remains an open challenge with existing methods.

To overcome all these limitations, we introduce PERGAMO, an approach to learn a deformation model for 3D garments from a single monocular video.
PERGAMO is based on two key features: it is learned from casual real-world images, hence there is no simulation-to-real gap or need for multi-camera setups;
and it is highly-efficient to evaluate, since at inference time it uses a shallow neural network that directly outputs garment deformations.
\Change{All in all, our main contribution is a 3D clothing reconstruction pipeline that is able to recover the explicit layer of a garment from just monocular RGB input. These reconstructions enable us to train a data-driven model to infer how a \textit{specifi}c garment deforms.}

To formulate PERGAMO, we use a novel two-stage approach where we first build a dataset by reconstructing the 3D geometry of deformed garments (Section \ref{sec:garment-reconstruction}), and then learn a nonlinear regressor from the reconstructed meshes. 
More specifically, we initially extract human-related features  such as body segmentation, body pose, and body normals, from the input images (Section \ref{sec:extracting-info}), which we leverage to deform a mesh template to reconstruct fine-scale detailed clothing using a differentiable rendering optimization (Sections \ref{sec:parametric-garment} and \ref{sec:fine-wrinkles}).
Then, we use the reconstructed garments as ground truth data to train a 3D garment deformation regressor (Section \ref{sec:regressor}).
We show that the learned model outputs pose-dependent garment surface details, such as folds and wrinkles, that closely match the real-word behaviour of the garment.

%% file: sections/related-work.tex
\section{Related Work}
Our work is related to the areas of 3D reconstruction and animation of garments and humans. In this section we discuss the existing literature in these areas.

\paragraph*{Multi-Camera 3D Garment Reconstruction.}
Many works on 3D garment reconstruction require multi-camera setups to capture the scene from different viewpoints. 
Initial approaches use studio setups, controlled environment, and color-coded fabrics
\cite{scholz2005garment,white2007capturing} to resolve depth and appearance ambiguities. 
Follow-up methods lifted the need for color patterns by, for example, using photometric stereo and controlled color lighting \cite{hernandezICCV07}, or using multi-view stereo ~\cite{bradley2008markerless}.
This enabled the 3D reconstruction of markerless clothing in controlled multi-camera settings, but output meshes lack high-frequency details such as folds and wrinkles.
To mitigate this, some methods use postprocessing steps to add wrinkle detail.
For example, Popa \textit{et al.} \cite{popa2009wrinkling} add spatio-temporal coherent wrinkle details to the 3D reconstructed meshes by detecting edges in the original images.
Wu \textit{et al.} \cite{wu2011shading} leverage shading information in general unconstrained lighting conditions to add fine-scale dynamic wrinkles. 
Alternatively, L\"{a}hner \textit{et al.} \cite{lahner2018deepwrinkles} use an image-to-image translation network based on a Generative Adversarial Network (GAN) to add high-frequency detail to a normal map.

Assuming a detailed point cloud, typically obtained from sophisticated multi-camera setup, it greatly facilitates the 3D garment reconstruction task. 
For example, ClothCap \cite{pons2017clothcap} reconstructs fine 3D garment geometry as well as the underlying body.
Similarly, Xiang \textit{et al.} ~\cite{xiang2021modeling} leverage an expensive 140 camera setup to reconstruct dressed humans, including an explicit clothing layer.
Bhatnagar \textit{et al.} \cite{bhatnagar2019multi} demonstrate that, given a large dataset of 3D scans, it is also possible to learn to infer 3D garments geometry from images.
Bang \textit{et al.} \cite{bang2021estimating} estimate accurate garment 2D patterns from full body scans, which enables to re-animate the recovered clothing using simulation.
We follow a similar goal, but propose a method that only requires a single uncalibrated camera.

\paragraph*{Monocular 3D Garment Reconstruction.}
To democratize the 3D reconstruction of garments, some methods have focused on relaxing the capture requirements and enable the reconstruction from single RGB image \cite{zhou2013garment,chen2015garmentmodeling,danvevrek2017deepgarment,jiang2020bcnet,yang2018garmentrecovery,su2022mulaycap}.
Zhou \textit{et al.} \cite{zhou2013garment} combine human pose estimation, garment outline prediction, and shape-from-shading cues to reconstruct garments from monocular images.
Dan{\v{e}}{\v{r}}ek \textit{et al.} \cite{danvevrek2017deepgarment} propose a learning-based solution to directly estimate garment vertex displacements from single images. 
Similarly, Jiang \textit{et al.} \cite{jiang2020bcnet} also learn to infer parametric garments ready to be worn by a SMPL \cite{loper2015smpl} human model.
Other methods require complex physics simulation steps \cite{yang2018garmentrecovery,su2022mulaycap} to drive the surface reconstruction such that it matches the input image.
We also reconstruct a garment layer from single image input, but we achieve significantly higher wrinkle detail and do not require physics simulation.
Key to our success is the combination of state-of-the-art normal map prediction with a differentiable rendering scheme to optimize vertex positions.

To reduce the inherent depth ambiguity in monocular RGB images, some methods resort to depth or RGB-D input to reconstruct garments.
For example, Chen \textit{et al.} \cite{chen2015garmentmodeling} combine garment parsing and contextual priors to reconstruct garments in rest pose from Kinect-based input.
Yu \textit{et al.} \cite{yu2019simulcap} combine RGB-D input with physics-based simulation to reconstruct 3D garments.
Despite the promising results, requiring depth input hinders the use of these approaches in domestic environments.

\paragraph*{Multi-Camera 3D Human Reconstruction.} Our work is also related to the methods that, instead of reconstructing an explicit mesh to represent the garment, aim at reconstructing a full dressed body using a single mesh \cite{starck2007surface,vlasic2008articulated,de2008performance,stoll2010video,robertini2016model}.
This line of research, often referred to \textit{performance capture}, usually employs multi-camera setups, and enables the replay of captured performances from any viewpoint.
Methods within this category can be split into model-based~\cite{de2008performance,vlasic2008articulated,stoll2010video}, which deform a template to fit it into the images, and model-free \cite{starck2007surface,collet2015high,wang2016capturing}, which have no prior knowledge of the shape and estimate it using the visual hull given by multi-view silhouettes.
Despite the high realism of the reconstructions achieved by these methods, they suffer from two main limitations: first, clothing and body are reconstructed in a single mesh, which precludes using the garment for animation; and second, the requirement for a multi-camera setup hinders their use in uncontrolled setups.
Our work addresses both limitations.

\paragraph*{Monocular 3D Human Reconstruction.} Many methods have been proposed to directly reconstruct dressed 3D humans as a single 3D mesh (\textit{i.e.,} no explicit modeling of a garment layer) from monocular RGB \cite{zheng2019DeepHuman,habermann2020deepcap,xiang2020monocloth,ma2020dressing3d,alldieck19cvpr} or RGB-D input \cite{burov2021dsfn,yu2017bodyfusion}.
Most of these works build on top of the impressive advances in deep learning for human pose estimation \cite{mehtaSIGGRAPH2017} and parametric model fitting \cite{hmrKanazawa17,pavlakos2018learning} in single images.

A common strategy in RGB methods is to use a 3D template of the subject, and deformed it to match visual cues (\textit{e.g.}, silhouette, 2D joints, etc.) extracted from the input image \cite{xu2018monoperfcap}.
Custom GPU-based optimizers have enabled such methods to run at interactive frame rates
\cite{habermann2019livecap} and, more recently, data-driven methods have been train to directly infer the geometry of dressed humans \cite{habermann2020deepcap}.
Despite the progress made by these methods, many challenges remain: requiring a subject template is not ideal, output meshes tend to exhibit baked-in garment details that remain static across the poses \cite{guo2021performance} and, most importantly, they do not explicitly separate garment from body.

Closest to ours is MonoClothCap~\cite{xiang2020monocloth}, an RGB method that does not require a subject-specific template. Instead, a generic parametric human model \cite{loper2015smpl} is deformed using a differentiable rendering scheme to fit into per-frame estimated normals of an actor.
We follow a similar path, but we are able to reconstruct an explicit layer of the garment, which subsequently allows us to train a regressor to predict how such garment deforms given arbitrary body motions.

As an alternative representation for 3D humans, recent works have explored the use of learning-based implicit functions \cite{park2019deepsdf}.
This representation is continuous, compact, and differentiable, which has enabled to efficiently and accurately reconstruct humans from single RGB images \cite{saito2019pifu,saito2020pifuhd,zheng2021pamir}
or partial point clouds \cite{bhatnagar2020ipnet,palafox2021npms}.
Despite the impressive advances in this domain, we focus our work on mesh-based representations and explicit garments, which directly enables animation of clothing in well-established character animation pipelines.

\paragraph*{3D Garment Animation.} Many works exist that address the problem of creating 3D animatable garments and dressed avatars. 
The key underlying challenge is to learn a model that is able to infer how the surface of the garment deforms as a function of the body pose and/or shape.
Earlier works assume multi-camera footage \cite{xu2011videobased,casas2014video4d}, and do not model clothing layer independently.
Assuming accurate 3D reconstructed meshes of dressed humans, some methods have tackled the modeling of garment deformations as a function of the underlying body 
\cite{yang2018analyzing,neophytou2014layered,pons2017clothcap}.
Our work tackles a similar problem, but learns to deform a garment layer from monocular RGB. 

With the raise of data-driven methods to learn highly nonlinear problems, recent works have attempted to model the deformation of garments leveraging large datasets of simulated data \cite{chentanez2020cnn,jin2020pixel}.
Such ground-truth dataset are usually generated offline using computationally expensive physics-based simulations \cite{narain2012arcsim}.
Closest to ours, Santesteban \textit{et al.} \cite{santesteban2019virtualtryon} use a recurrent neural network to learn shape-and-pose dynamic deformations of a single garment.
Similarly, TailorNet\cite{patel2020tailor} predicts garment deformations also as a function of clothing style, and decompose deformation into a high frequency and a low frequency component to improve the detail. 
Vidaurre \textit{et al.} \cite{vidaurre2020fcgnn} show better generalization capabilities to garment design by leveraging a fully convolutional architecture capable of infering deformations for any mesh topology.
Despite great results, these methods require loss supervision at the vertex level. Alternatively, very recent methods demonstrate that it is possible to learn deformations directly from physics-based losses \cite{bertiche2021pbns,santesteban2022snug} or using distance fields \cite{corona2021smplicit}.

Animation of 3D garments has also been tackled using clothed humans represented as a single mesh.  
CAPE \cite{ma2020dressing3d} uses detailed 4D scans of dressed avatars to learn a graph convolutional neural network to deform a full body mesh template.
More recent works use implicit representations \cite{Saito:CVPR:2021,wang2021metaavatar} and are capable of encoding finer details.
Nonetheless, using a single mesh to encode body and garment is not ideal for animation.
Instead, we focus on learning an model for garments using an explicit mesh 3D layer. As we demonstrate, this enables to dress animated characters using existing motion capture datasets such as AMASS \cite{AMASS:ICCV:2019}.

%% file: sections/overview.tex
\begin{figure*}
    \centering
    \includegraphics[width=\textwidth]{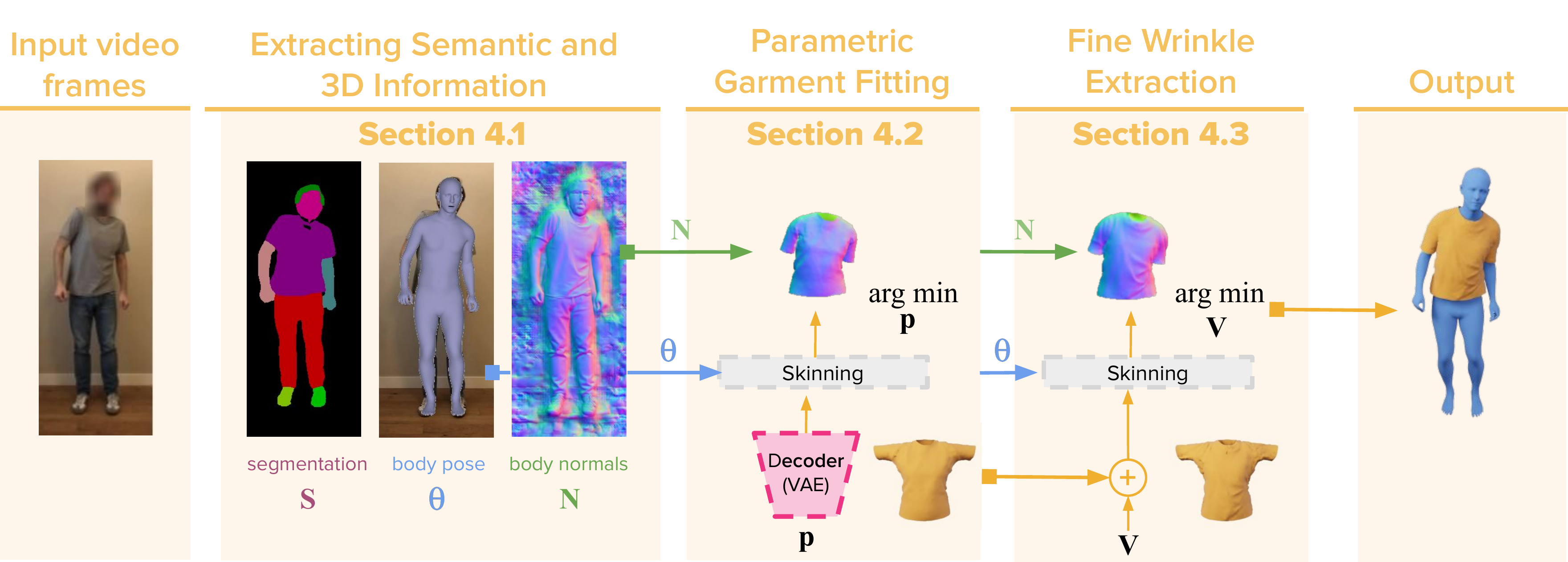}
    \caption{Overview of our 3D garment reconstructing approach. Starting from an RGB frame, we first extract human semantic and 3D information. We then fit a coarse garment template parameterized by a latent vector $\mathbf{p}$ to the estimated body normals, and finally we add fine-scale wrinkles by optimizing per-vertex displacement $\mathbf{V}$ using differentiable rendering.}
    \label{fig:pipeline}
\end{figure*}
\section{Overview}
Our goal is to learn a deformable model for 3D garments that faithfully reproduces the behavior of a \textit{real} garment-subject pair.
In other words, we want to model how a specific garment deforms when worn by a particular subject. This effectively enables a \textit{personalized} animation of garments for many tasks including: virtual try-on, telepresence, VR, videogames, and more.

More formally, we want to learn a model $R$ that outputs a garment
\begin{align}
 	\mathbf{X}&=R(\uptheta),
 	\label{eq:regressor_in_model} 
\end{align}
where $\mathbf{X} \in \mathbb{R}^{N_{\text{G}} \times 3}$ encodes the vertices of the deformed garment in rest pose (\textit{i.e.,} it contains pose-dependent wrinkles and folds), and $\uptheta$ is the target pose.
In practice, the garment $\mathbf{X}$ is worn by a parametric human body such as a SMPL \cite{loper2015smpl} that represents the body surface of the target subject.
To rig the deformed garment $\mathbf{X}$, we borrow the skinning functionalities (\textit{e.g.}, skeleton, rigging weights) from the underlying body model.

Given our goal for personalized garments (\textit{i.e.}, specific real-world behavior), the ground-truth data to learn $R()$ cannot be obtained with physics-based cloth simulation tools as done in state-of-the-art methods \cite{santesteban2019virtualtryon,patel2020tailor,vidaurre2020fcgnn,bertiche2021pbns}, for two reasons: 
first, due the unknown mechanical properties of the target garment, which are hard to obtain even in complex laboratory setups \cite{miguel2012data};
and, second, due to the unknown soft-tissue model of the underlying body, which also effects how garment behave \cite{romero2020skinmechanics}.
These limitations preclude generating synthetic data that match specific real-world behavior.

Therefore, instead of learning to deform garments using 3D simulations, we propose a image-driven approach that learns to infer the true behaviour of a garment from videos.
To this end, in Section \ref{sec:garment-reconstruction} we introduce a method to build a database of 3D meshes by directly reconstructing garments from a set of unconstrained monocular videos.
Then, in Section \ref{sec:regressor}, we describe a neural regressor that learns to faithfully deform a 3D garment given a target pose of the subject.

%% file: sections/reconstruction.tex
\section{3D Garment Reconstruction}
\label{sec:garment-reconstruction}
To obtain the dataset that we used to learn our personalized 3D deformation model, we directly reconstruct the surface of the target garment captured from monocular videos.
\Change{Figure \ref{fig:pipeline} illustrates our reconstruction pipeline.}

Since recovering 3D geometry from single view is an ill-posed problem, we tackle the task into three steps. First, in Section \ref{sec:extracting-info}, we extract a combination of semantic and 3D features from the input image. This information is key to reduce the complexity inherent to our problem.
Second, in Section \ref{sec:parametric-garment}, we fit a coarse parametric garment model into the input image by leveraging the extracted features.
And finally, in Section \ref{sec:fine-wrinkles}, we add fine wrinkle details into the garment using a differentiable rendering scheme.
\subsection{Extracting Semantic and 3D Information}
\label{sec:extracting-info}
Given a monocular video sequence $\mathcal{I}=\{ \mathbf{I}_{t}\}_{t=0}^{T}$, where $\mathbf{I}_{t}$ is an RGB frame, captured in a uncontrolled setting (\textit{i.e.}, unknown camera parameters, unknown illumination, unknown subject pose), we initially extract a set of human-related image features for each frame $\mathbf{I}_{t}$.
The extracted information will ease our ill-posed 3D reconstruction task that we describe later in the rest of this section.

First, we leverage a state-of-the-art human parsing approach \cite{li2020self} to assign a per-pixel label to the input image $\mathbf{I}$ (to simplify notation, we drop the subindex $t$ in the rest of this section, but note that all steps are done per each frame).
As output, we obtain a set of binary segmentation masks $\mathbf{S}_i$, one for each label type, that encodes what body part or garment type (\textit{e.g.}, t-shirt, trousers, head, right/left arm, etc.) each pixel contains. 
For the results shown in this work, we use the binary mask $\mathbf{S}_{\text{g}}$, which corresponds to the upper body clothing.

Second, we use an image-to-image translation network, based on the work of Saito \textit{et al.} \cite{saito2020pifuhd}, to estimate the surface normals $\mathbf{N}$ for each pixel of the input image $\mathbf{I}$.
Notice that estimated normals for pixels in the background can be unreliable or noisy, but since our goal focuses only on garment regions, we do not suffer from these artifacts. %

Finally, we estimate the pose $\theta \in \mathbb{R}^{69}$ of the subject in the input image $\mathbf{I}$ using a state-of-the-art human pose estimation method \cite{choutas2020expose}. 
\subsection{Parametric Garment Fitting}
\label{sec:parametric-garment}
To reconstruct a \Change{coarse approximation} of the 3D geometry of a deformed garment that appears on an input frame $\mathbf{I}$, we first use a model-based strategy. To this end, we use a parametric garment model 
\begin{equation}
G_{\text{coarse}}(\mathbf{p},\theta) = W(D(\mathbf{p}),\theta,\mathcal{W}),
\label{eq:parametric-garment}
\end{equation}
where $W$ is a skinning function (\textit{e.g.}, linear blend skinning) that articulates a parametric garment template $D(\mathbf{p}) \in \mathbb{R}^{N_{\text{G}}\times3}$ based on weights $\mathcal{W}$.
Our parametric garment template  $D(\mathbf{p})$ is learned from a public dataset of 3D garments \cite{santesteban2019virtualtryon} using a variational
autoencoder network, where $\mathbf{p} \in \mathbb{R}^{25}$ %
is the latent space learned with the autoencoder, and $D()$ the decoder block.
\Change{Intuitively, $\mathbf{p}$ encodes a latent representation of T-shirt deformations, and $D()$ is the mapping from latent variable to vertices position of a template mesh.}
This parametric template encodes coarse (\textit{i.e.,} not specific to \Change{the material of} the target garment) deformations of garments in rest pose.
Hence, this first fitting step aims at recovering a coarse version of the garment visible in the input frame $\mathbf{I}$.

In order to fit the garment model of Equation \ref{eq:parametric-garment} to a frame $\mathbf{I}$, we leverage the human-related image features $\mathbf{S}_{\text{g}}$, $\mathbf{N}$, and $\uptheta$ described in Section \ref{sec:extracting-info}, and formulate the following optimization problem
\begin{equation}
    \underset{\mathbf{p}}{\arg\min}~~ \mathcal{E}_{\text{coarse}} + \mathcal{E}_{\text{sil}} + \mathcal{E}_{\text{temp}}
    + \mathcal{E}_{\text{reg}}.
\end{equation}

$\mathcal{E}_{\text{coarse}}$ is the main data term, and enforces the normals of the fitted parametric garment to match the predicted normals $\mathbf{N}$.
This term is formulated as
\begin{equation}
    \mathcal{E}_{\text{coarse}} = \uplambda_{\text{coarse}_\mathbf{p}}~\| \phi_{\text{N}} (G(\mathbf{p},\theta))- (\mathbf{N}~\odot~\mathbf{S}_{\text{g}})\|^2
    \label{eq:e_coarse}
\end{equation}
where $\phi_{\text{N}}$ \Change{is} a differentiable rendering function that outputs camera-space per-pixel normals of the garment $G(\mathbf{p},\theta)$ and $\odot$ is the Hadamard product (\textit{i.e.}, element-wise multiplication).

In practice, the function $\phi_{\text{N}}$ is implemented using the state-of-the-art differentiable rendering library Kaolin \cite{KaolinLibrary}, which enables the computation of gradients of the image error w.r.t vertices position.

$\mathcal{E}_{\text{sil}}$ is a data-term that enforces the silhouette of the fitted garment to match the predicted mask $\mathbf{S}_{\text{g}}$,  
\begin{equation}
    \mathcal{E}_{\text{sil}} = \uplambda_{\text{sil}_\mathbf{p}}~\| \phi_{\text{S}} (G(\mathbf{p},\theta))-\mathbf{S}_{\text{g}}\|^2,
\end{equation}
where $\phi_{\text{S}}$ is a differentiable rendering function that outputs \Change{a mask with the silhouette} of the garment  \Change{(with 1s indicating inside, 0s outside and bit of a smooth transition on the edges)} in camera space, \Change{also implemented using the rendering library Kaolin \cite{KaolinLibrary}}.
$\mathcal{E}_{\text{temp}}$ and $\mathcal{E}_{\text{reg}}$ are regularizers formulated as
\begin{eqnarray}
\mathcal{E}_{\text{temp}} &=& \uplambda_{\text{temp}_\mathbf{p}}~\|\mathbf{p}_{t-1} - \mathbf{p}_t\|^2\\
\mathcal{E}_{\text{reg}} &=& \uplambda_{\text{reg}_\mathbf{p}}~\|\mathbf{p}_t\|^2
\end{eqnarray}
that enforce temporal stability and avoid unnatural deformations, respectively.

\subsection{Fine Wrinkle Extraction}
\label{sec:fine-wrinkles}
The parametric garment model fitted in Section \ref{sec:parametric-garment} is only capable to represent coarse details of the garment.
In order to add personalized details (\textit{e.g.}, material-specific wrinkles, pose-and-shape-depending details, etc.), in this final reconstruction step we compute a vector $\mathbf{V} \in \mathbb{R}^{N_{\text{G}}\times3}$ of per-vertex 3D displacements to add fine details.
To this end, we formulate our final garment model as  
\begin{equation}
G_{\text{fine}}(\mathbf{p},\theta,\mathbf{V}) = W(D(\mathbf{p})+\mathbf{V},\theta,\mathcal{W}),
\label{eq:parametric-garment-fine}
\end{equation}
and our goal in this last reconstruction step is to find the vector of 3D displacements $\mathbf{V}$ for each frame we want to reconstruct.
Notice that the rest of parameters of the model are already known.

Similar to the parametric garment fitting step from Section \ref{sec:parametric-garment}, we find the 3D displacements vector $\mathbf{V}$ solving an optimization at each frame 
\begin{equation}
\begin{aligned}
    \underset{\mathbf{V}}{\arg\min}&~~\mathcal{E}_{\text{fine}} + \mathcal{E}_{\text{edge}} + \mathcal{E}_{\text{temp}}
    + \mathcal{E}_{\text{reg}}\\
    \textrm{s.t.}&~~\forall \mathbf{v}_i\in \mathbf{V}: \eta_{\text{max}} > \mathbf{v}_i > \eta_{\text{min}},
\end{aligned}
\end{equation}
where $\eta_{\text{min}}$ and $\eta_{\text{max}}$ are thresholds for minimum and maximum allowed displacements, respectively.
$\mathcal{E}_{\text{fine}}$ is the main data term, and it resemble to the term $\mathcal{E}_{\text{coarse}}$ from Equation \ref{eq:e_coarse}, but with two important differences.
Specifically,
\begin{equation}
\begin{aligned}
    &\mathcal{E}_{\text{fine}} = \uplambda_{\text{fine}_\mathbf{V}} \cdot \\
    & \; \; \; \| (\nabla~\phi_{\text{N}}  (G_{\text{fine}}(\mathbf{p},\theta, \mathbf{V}))- \nabla~(\mathbf{N}~\odot~\mathbf{S}_\text{g}))~\odot~\phi_{\text{S}} (G(\mathbf{p},\theta))\|^2
    \label{eq:e_fine}
\end{aligned}
\end{equation}
where $\nabla$ is the image gradient operator, $\odot$ is the Hadamard product (\textit{i.e.}, element-wise multiplication), and $\phi_{\text{S}} (G(\mathbf{p},\theta))$ a function that renders the silhouette of the parametric garment fitted in the previous section.
In practice, the $\mathcal{E}_{\text{fine}}$ term allows the vertices of the garment to move freely to match the image normal $\mathbf{N}$, up to fine details to reproduce wrinkles. 
The gradient image operator $\nabla$ ensures attention on the parts of the image normal where normals change (\textit{e.g.}, the wrinkles).
Finally, the element-wise multiplication constrains this optimization to focus only on the pixels of the image where the garment $G(\mathbf{p},\theta)$ appears.

Equation \ref{eq:e_fine} allows free vertex movement, however, many of the garment vertices are not visible from the input image due to occlusions and self-occlusions. 
Allowing occluded vertices to move can potentially lead to undesired reconstruction artefacts.  
To resolve this ill-pose problem, we rely on a set of regularizers to constrain the optimization. Namely, we use the following terms
\begin{eqnarray}
\mathcal{E}_{\text{edge}} &=& \uplambda_{\text{edge}_\mathbf{V}}~\|\mathbf{E}_{\text{rest}} - \mathbf{E}_{t}\|^2\\
\mathcal{E}_{\text{temp}} &=& \uplambda_{\text{temp}_\mathbf{V}}~\|\mathbf{V}_{t-1} - \mathbf{V}_t\|^2\\
\mathcal{E}_{\text{reg}} &=& \uplambda_{\text{reg}_\mathbf{V}}~\|\mathbf{V}\|^2,
\end{eqnarray}
where ${E}_{\text{rest}}$ and ${E}_t$ are the edge lengths of the garment in rest pose and the optimized garment, and $\mathbf{V}$ the optimized garment vertices.
These three terms temporally and spatially regularize the garment deformation, such that the reconstructed mesh does not exhibit unnatural deformations.

%% file: sections/regressor.tex
\section{3D Garment Regressor}
\label{sec:regressor}
The method introduced in Section \ref{sec:garment-reconstruction} allows us to create a dataset of 3D reconstructed meshes $M$, each of them with an associated pose parameter $\uptheta$.
In this section we describe how to use this data to learn the model defined in Equation \ref{eq:regressor_in_model}, capable of inferring pose-dependent deformations.
\Change{Notice that the model proposed here is independent of the reconstruction method, and it could potentially be used on other similar 3D garment datasets.}

\input{sections/figures/figure-dataset}

We follow the standard approach in data-driven garments \cite{santesteban2019virtualtryon}, and convert our dataset of reconstructed meshes into a dataset of per-vertex displacements with respect to a template mesh in T-pose.
Specifically, for each mesh $M$ of our dataset we compute ground truth displacements
\begin{equation}
\Delta_{\text{GT}} = W^{-1}(M, \uptheta, \mathcal{W}) - \mathbf{\bar{T}},
\end{equation}
where $W^{-1}$ is the inverse skinning transformation, $\uptheta$ the pose parameter, and $\mathbf{\bar{T}}$ is the average unposed garment of the dataset.
The final deformed garment is then defined as $\mathbf{X} = \mathbf{\bar{T}} + \Delta_{\text{GT}}$.

To formulate the model defined in Equation $\ref{eq:regressor_in_model}$, we use a regressor that is trained to predict per-vertex garment offsets $\Delta$ as a function of the target body pose $\uptheta$.
In practice, we implement it using a simple MLP network with 3 hidden layers, supervised with an L1 loss on vertex offsets and vertex normals:
\begin{equation}
    \mathcal{L} = L_{1}(\Delta, \Delta_{GT}) + L_{1}(N(\mathbf{\bar{T}} + \Delta), N(\mathbf{\bar{T}} + \Delta_{GT}))
\end{equation}
\Change{where $N()$ is a function that computes the per-vertex normals of the garment meshes. Check Section \ref{sec:implementation-details} for further implementation details.}

In contrast to existing data-driven methods that are trained on hundreds of simulated meshes \cite{patel2020tailor,santesteban2019virtualtryon,bertiche2020cloth3d}, we only have a reduced set of reconstruced sequences which significantly complicates the learning.
To ease the generalization capabilities of our regressor despite relatively small training set, we encode the pose vector $\uptheta$ in a compact subspace that better captures key pose features. To this end, we leverage the multi-modal autoencoder introduced by SoftSMPL \cite{santesteban2020softsmpl}, which encodes poses in $\mathbb{R}^{10}$.
Hence, our final regressor is $R():\mathbb{R}^{10} \rightarrow \mathbb{R}^{N_{\text{G}} \times 3}$, where $N_{\text{G}}$ is 4,424 for the garment showcased in our results.

\Change{Finally, notice that due to residual errors at inference time, a few garment vertices might collide with the underlaying body mesh.
Similar to existing works in data-driven garments \cite{santesteban2019virtualtryon,patel2020tailor}, we push problematic vertices the normal direction of the closest body point.}

%% file: sections/figures/figure-dataset.tex
{
\newcommand{\cropFrameL}{150}
\newcommand{\cropFrameB}{150}
\newcommand{\cropFrameR}{150}
\newcommand{\cropFrameT}{150}
\newcommand{\cropFrameBrownL}{300}
\newcommand{\cropFrameBrownB}{200}
\newcommand{\cropFrameBrownR}{150}
\newcommand{\cropFrameBrownT}{410}
\newcommand{\cropFrameYellowL}{240}
\newcommand{\cropFrameYellowB}{280}
\newcommand{\cropFrameYellowR}{250}
\newcommand{\cropFrameYellowT}{410}
\newcommand{\cropRenderL}{722}
\newcommand{\cropRenderB}{70}
\newcommand{\cropRenderR}{722}
\newcommand{\cropRenderT}{20}
\newcommand{\imWidth}{0.118}
\begin{figure*}
 \begin{subfigure}{{\imWidth}\linewidth}
    \includegraphics[trim={\cropFrameL} {\cropFrameB} {\cropFrameR} {\cropFrameT}, clip, width=\linewidth]{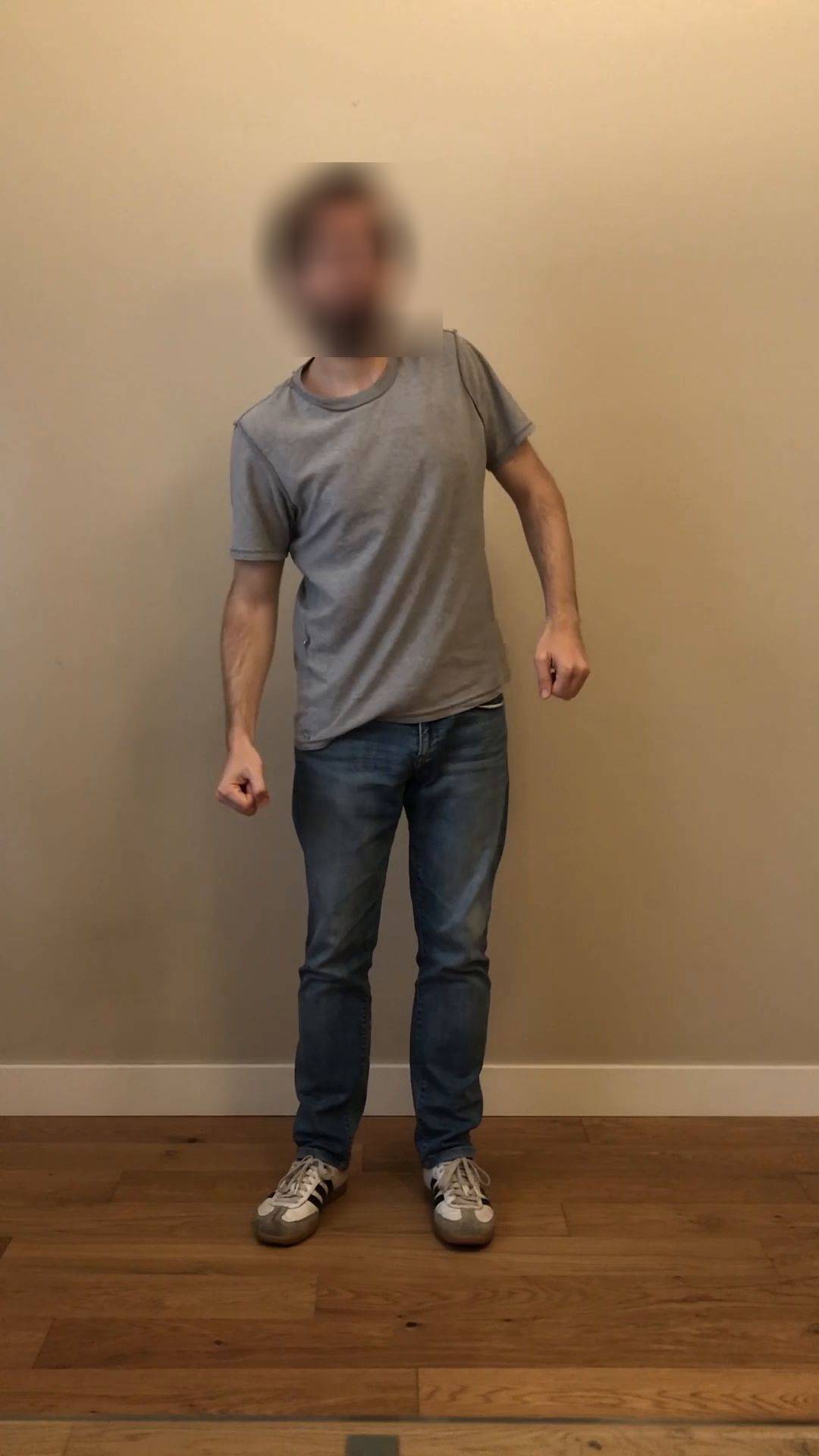}
  \end{subfigure}
  \begin{subfigure}{{\imWidth}\linewidth}
    \includegraphics[trim={\cropRenderL} {\cropRenderB} {\cropRenderR} {\cropRenderT}, clip,width=\linewidth]{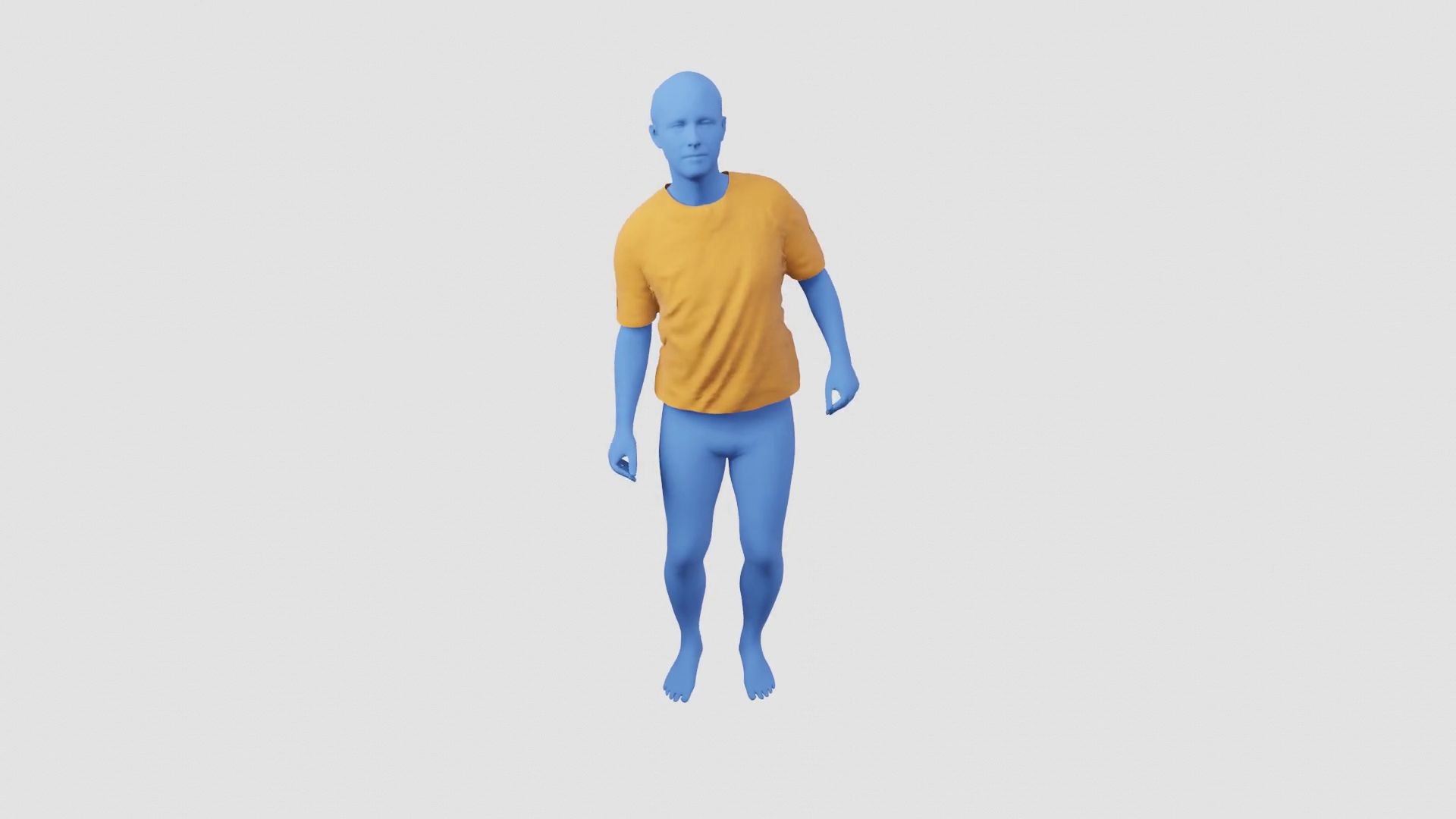}
  \end{subfigure}
  \hspace*{\fill}
  \begin{subfigure}{{\imWidth}\linewidth}
    \includegraphics[trim={\cropFrameL} {\cropFrameB} {\cropFrameR} {\cropFrameT}, clip, width=\linewidth]{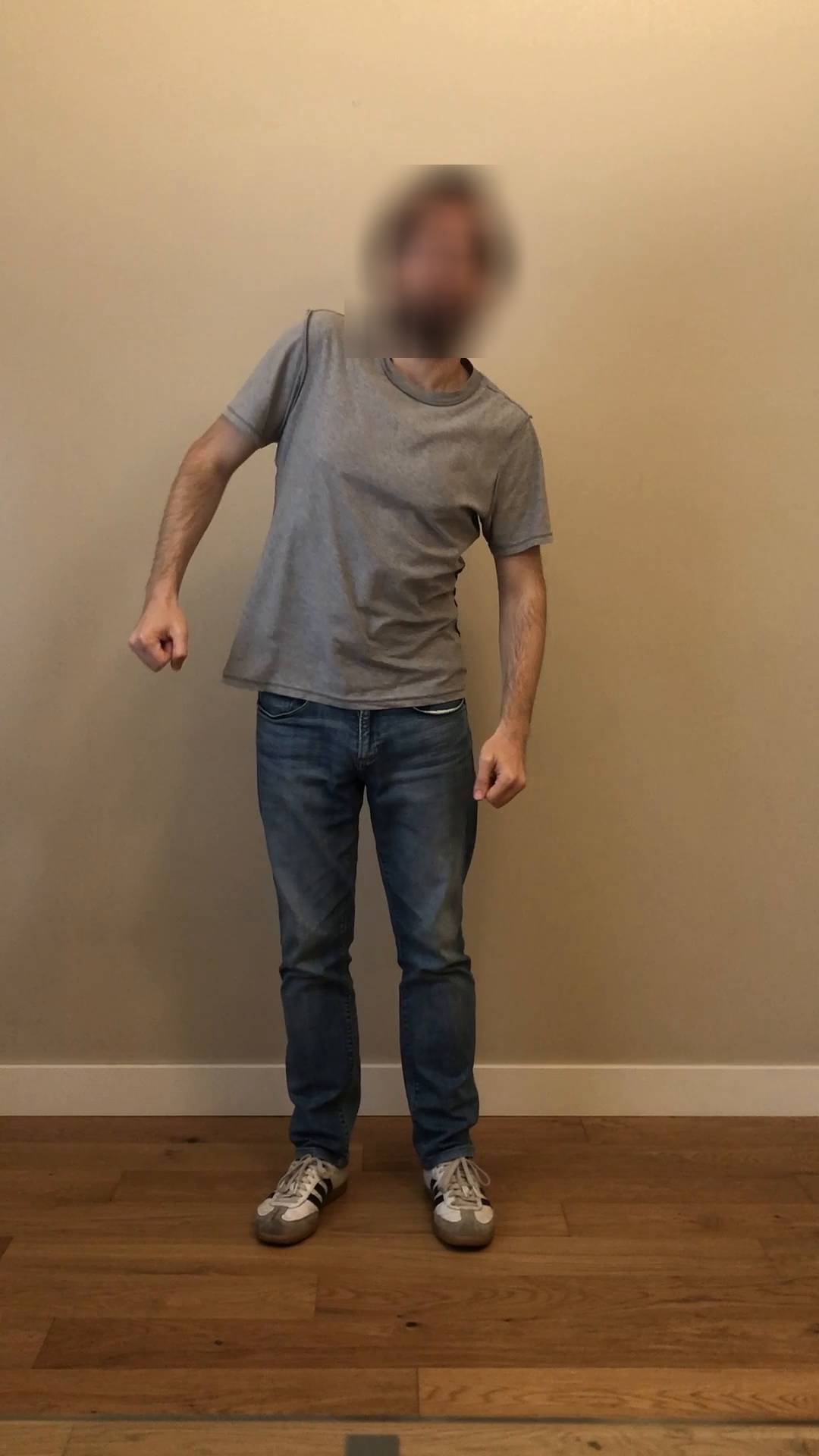}
  \end{subfigure}
  \begin{subfigure}{{\imWidth}\linewidth}
    \includegraphics[trim={\cropRenderL} {\cropRenderB} {\cropRenderR} {\cropRenderT}, clip,width=\linewidth]{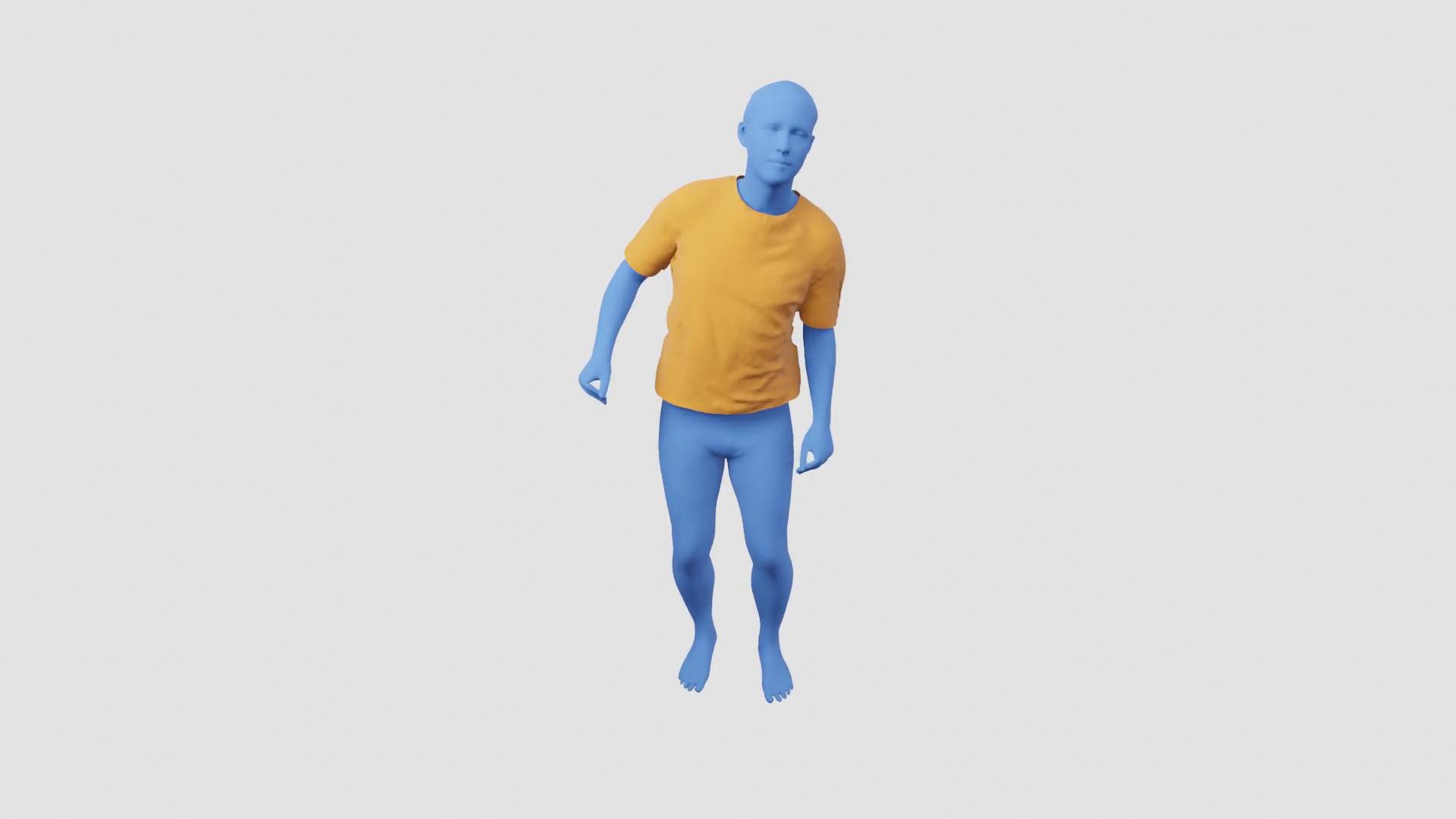}
  \end{subfigure}
  \hspace*{\fill}
   \begin{subfigure}{{\imWidth}\linewidth}
    \includegraphics[trim=140 0 140 260, clip, width=\linewidth]{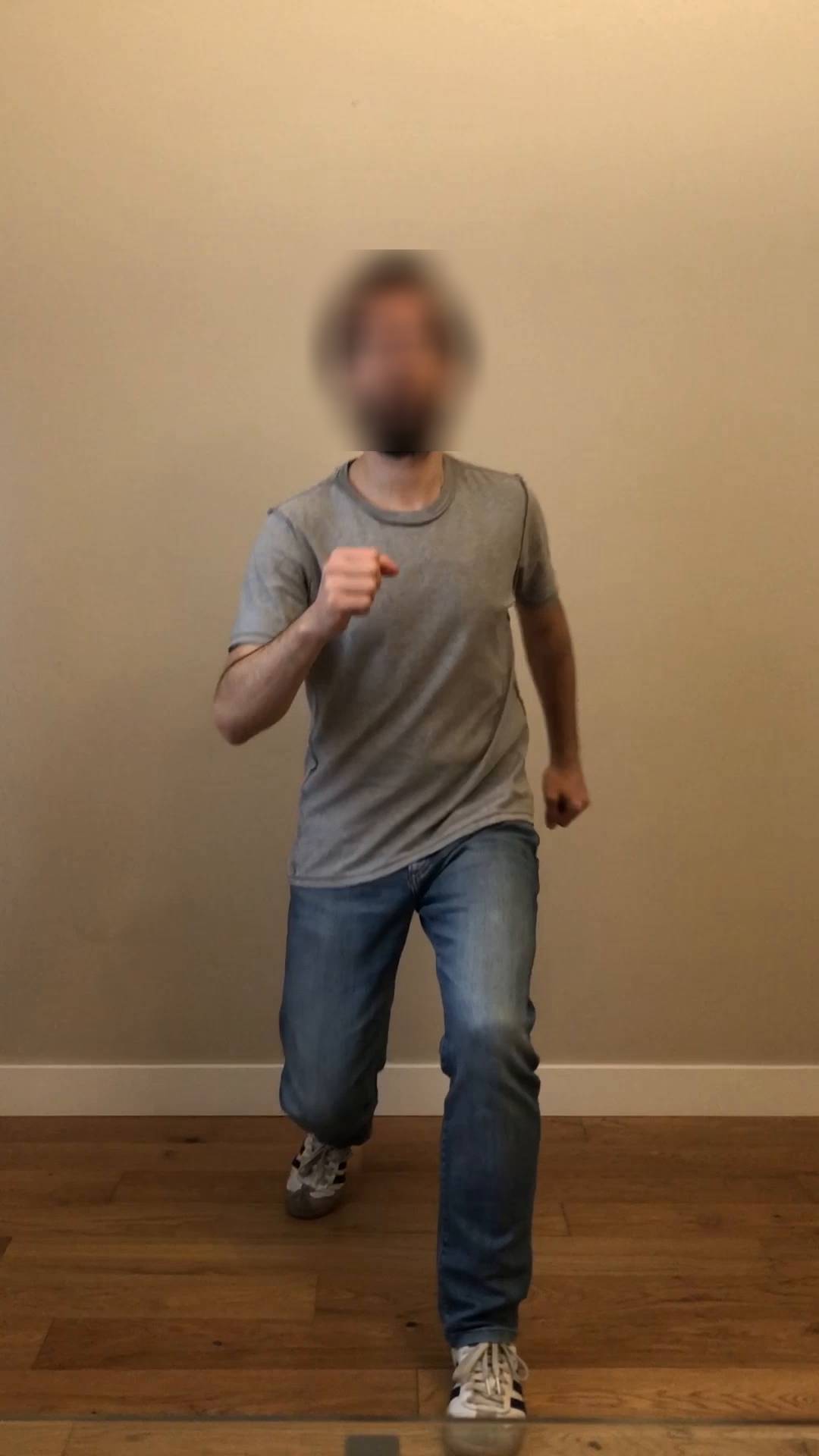}
  \end{subfigure}
  \begin{subfigure}{{\imWidth}\linewidth}
    \includegraphics[trim={\cropRenderL} {\cropRenderB} {\cropRenderR} {\cropRenderT}, clip,width=\linewidth]{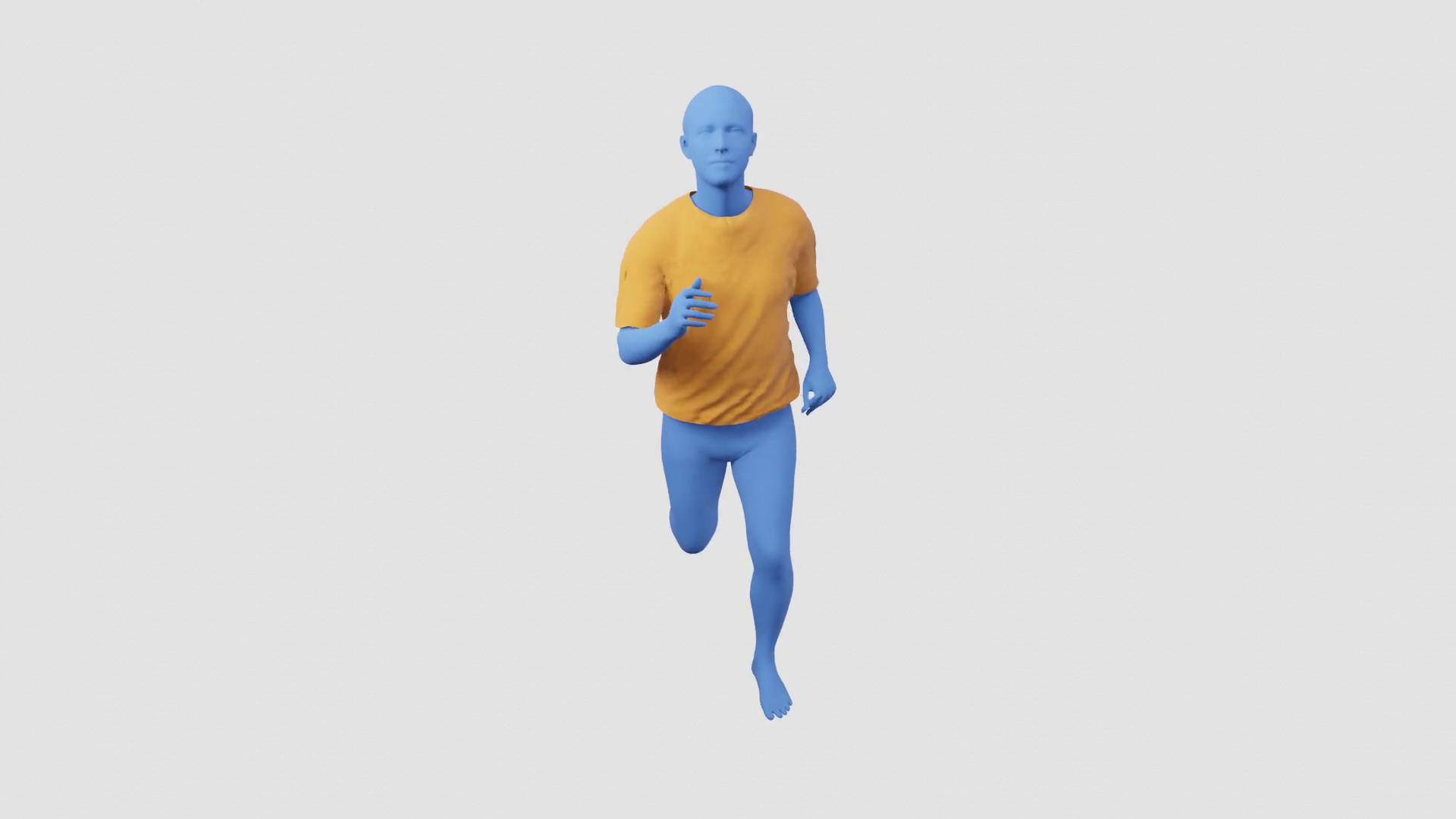}
  \end{subfigure}
  \hspace*{\fill}
  \begin{subfigure}{{\imWidth}\linewidth}
    \includegraphics[trim={\cropFrameL} {\cropFrameB} {\cropFrameR} {\cropFrameT}, clip, width=\linewidth]{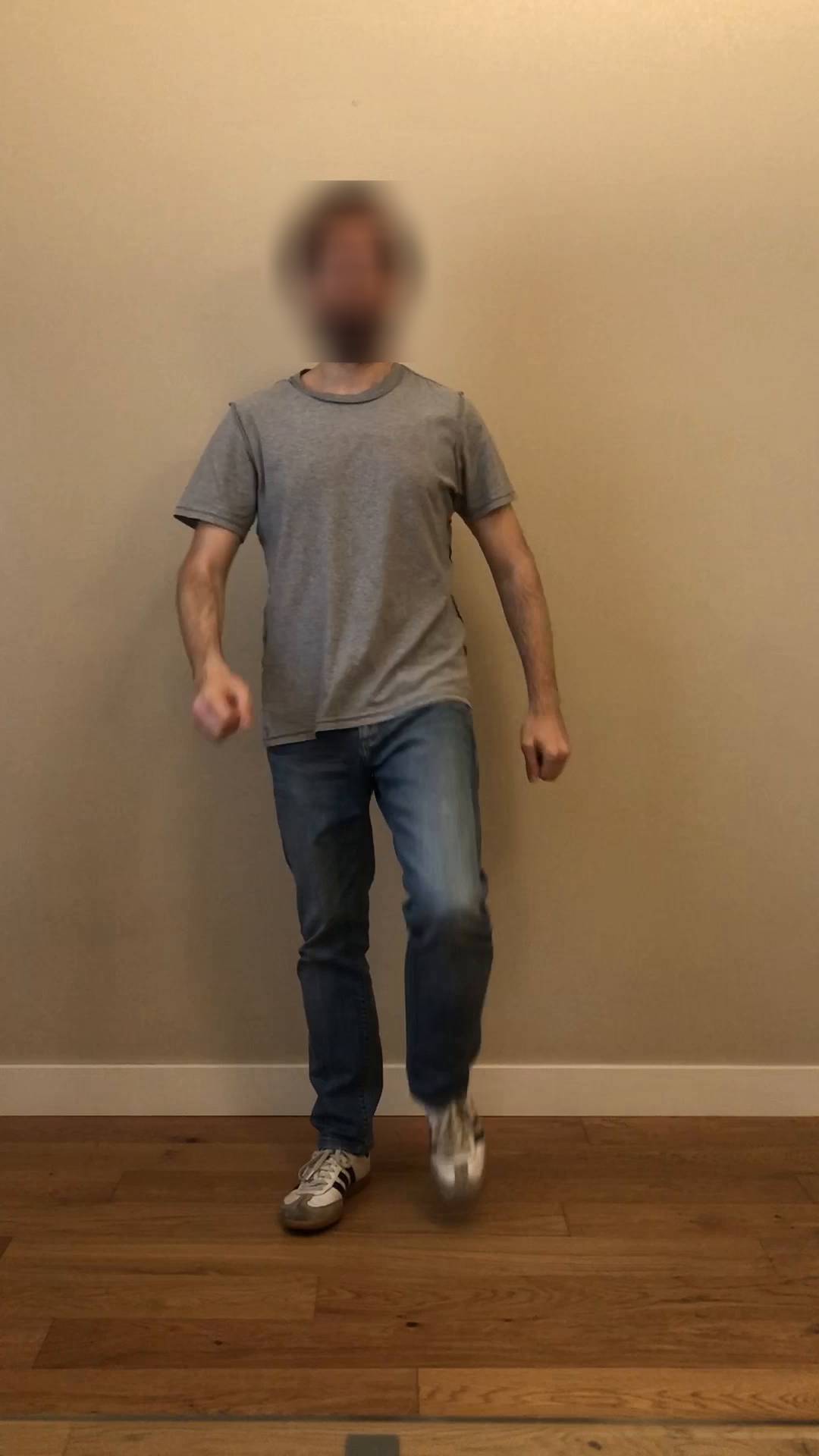}
  \end{subfigure}
  \begin{subfigure}{{\imWidth}\linewidth}
    \includegraphics[trim={\cropRenderL} {\cropRenderB} {\cropRenderR} {\cropRenderT}, clip,width=\linewidth]{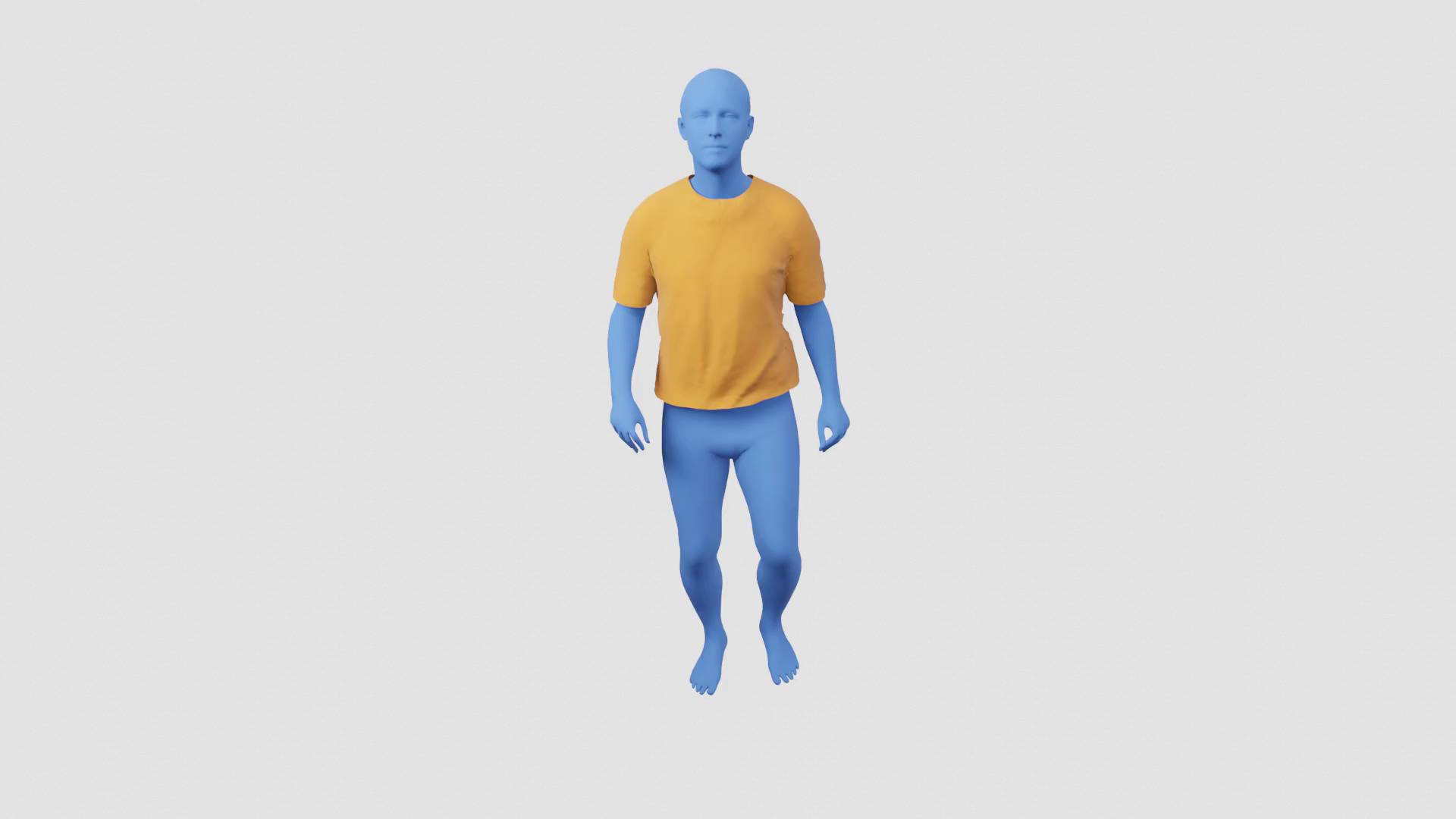}
  \end{subfigure}
  
  \vspace{0.2cm}
   \begin{subfigure}{{\imWidth}\linewidth}
    \includegraphics[trim={\cropFrameYellowL} {\cropFrameYellowB} {\cropFrameYellowR} {\cropFrameYellowT}, clip, width=\linewidth]{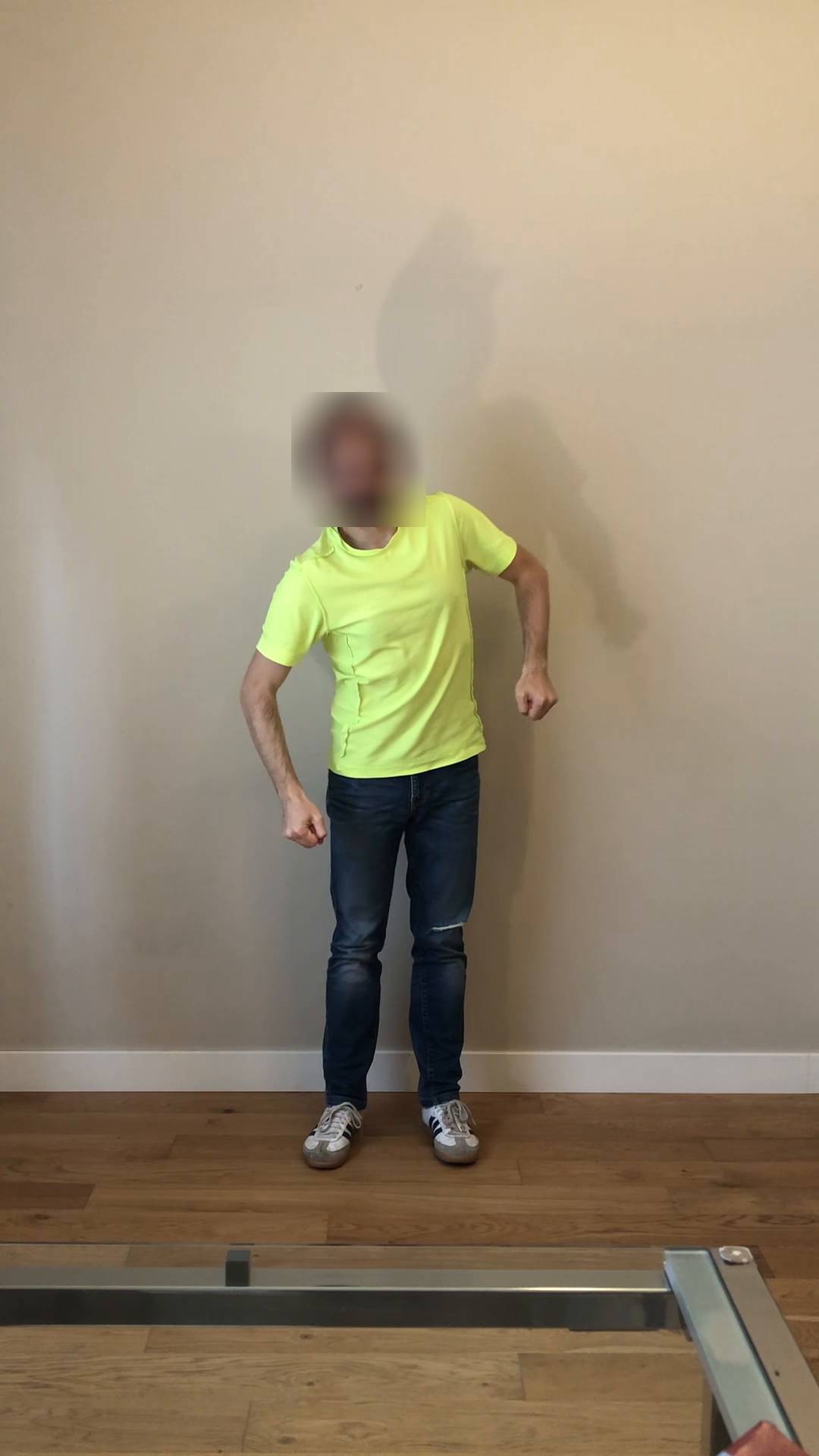}
  \end{subfigure}
  \begin{subfigure}{{\imWidth}\linewidth}
    \includegraphics[trim={\cropRenderL} {\cropRenderB} {\cropRenderR} {\cropRenderT}, clip,width=\linewidth]{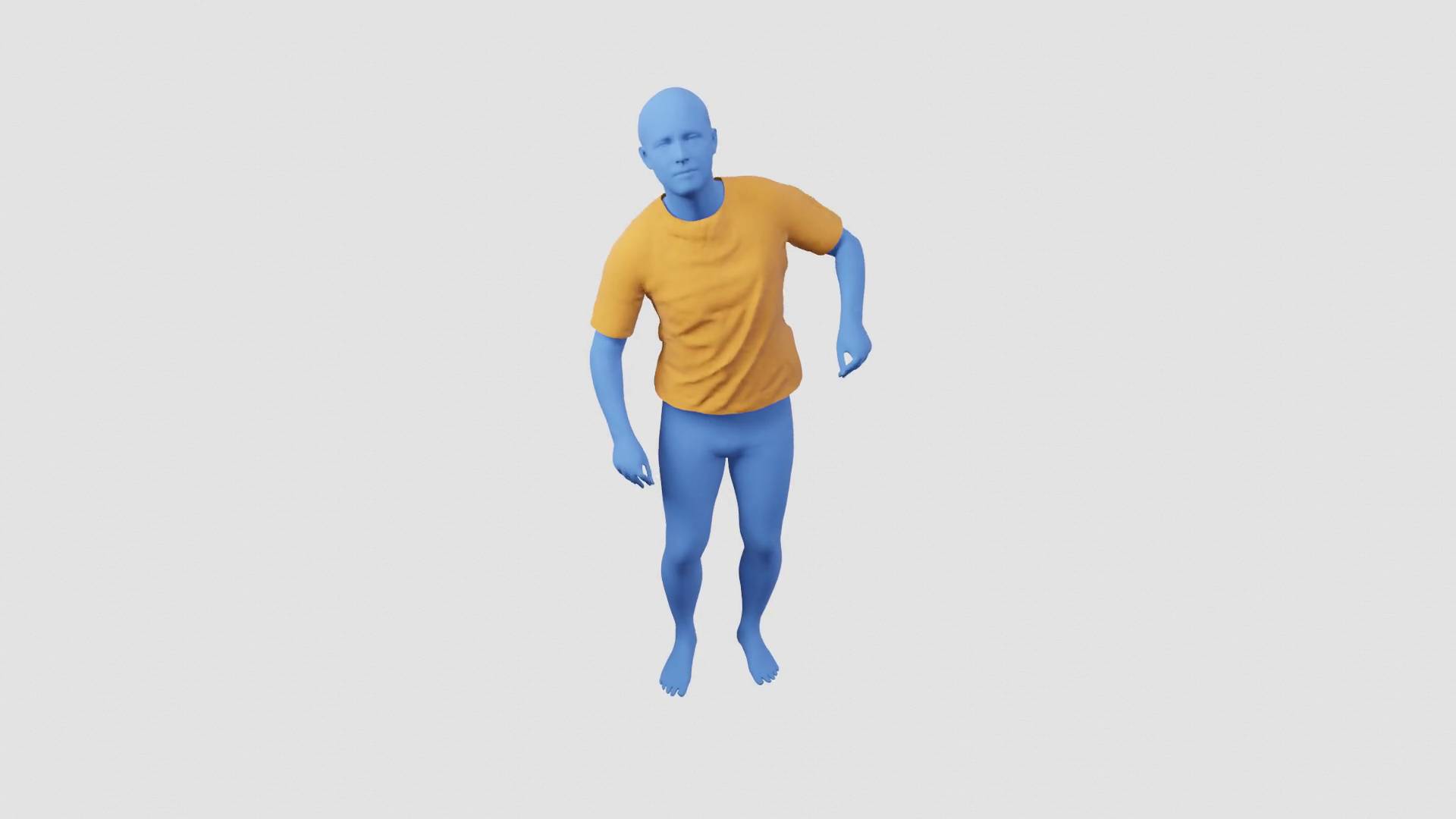}
  \end{subfigure}
  \hspace*{\fill}
  \begin{subfigure}{{\imWidth}\linewidth}
    \includegraphics[trim={\cropFrameYellowL} {\cropFrameYellowB} {\cropFrameYellowR} {\cropFrameYellowT}, clip, width=\linewidth]{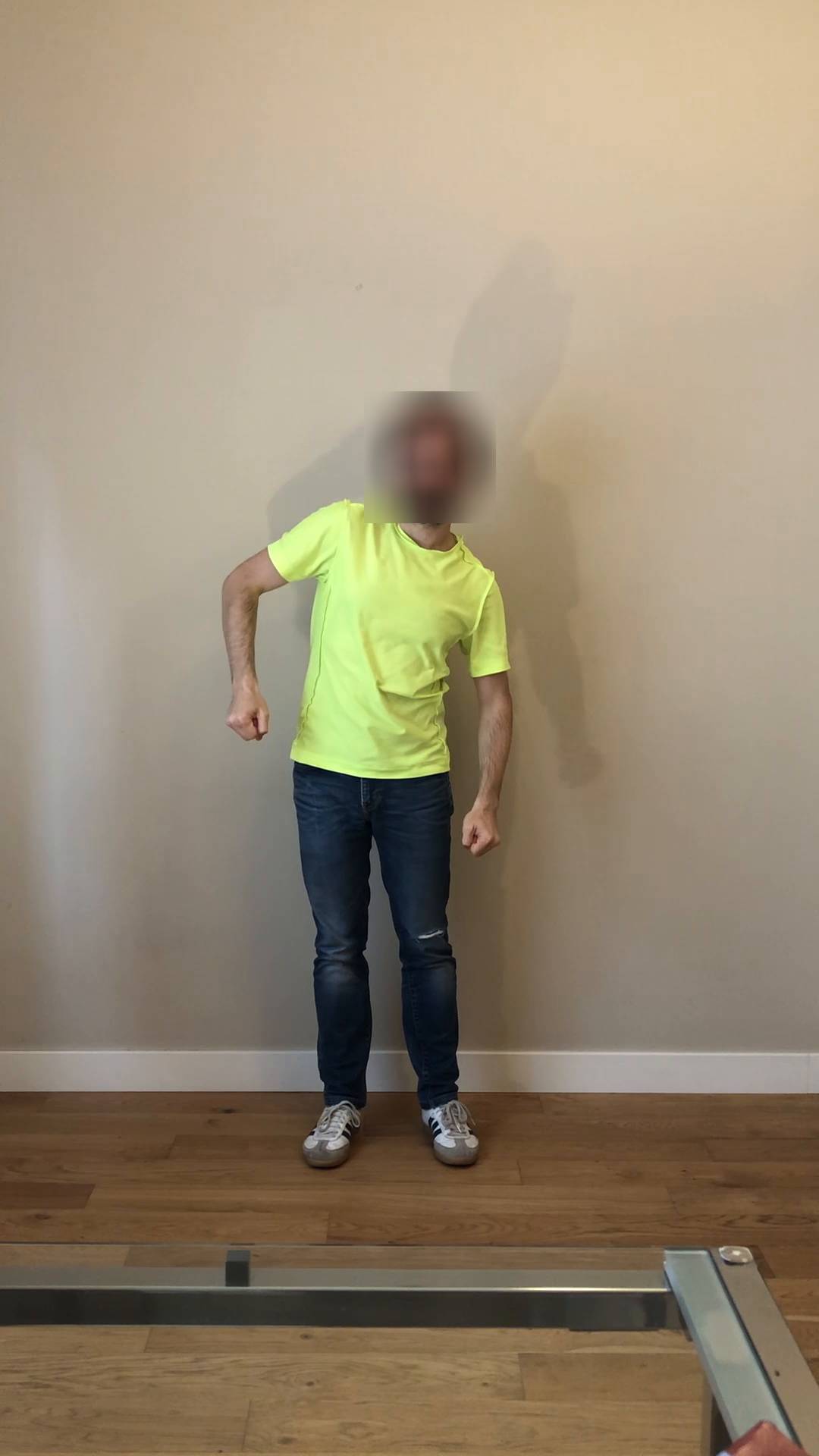}
  \end{subfigure}
  \begin{subfigure}{{\imWidth}\linewidth}
    \includegraphics[trim={\cropRenderL} {\cropRenderB} {\cropRenderR} {\cropRenderT}, clip,width=\linewidth]{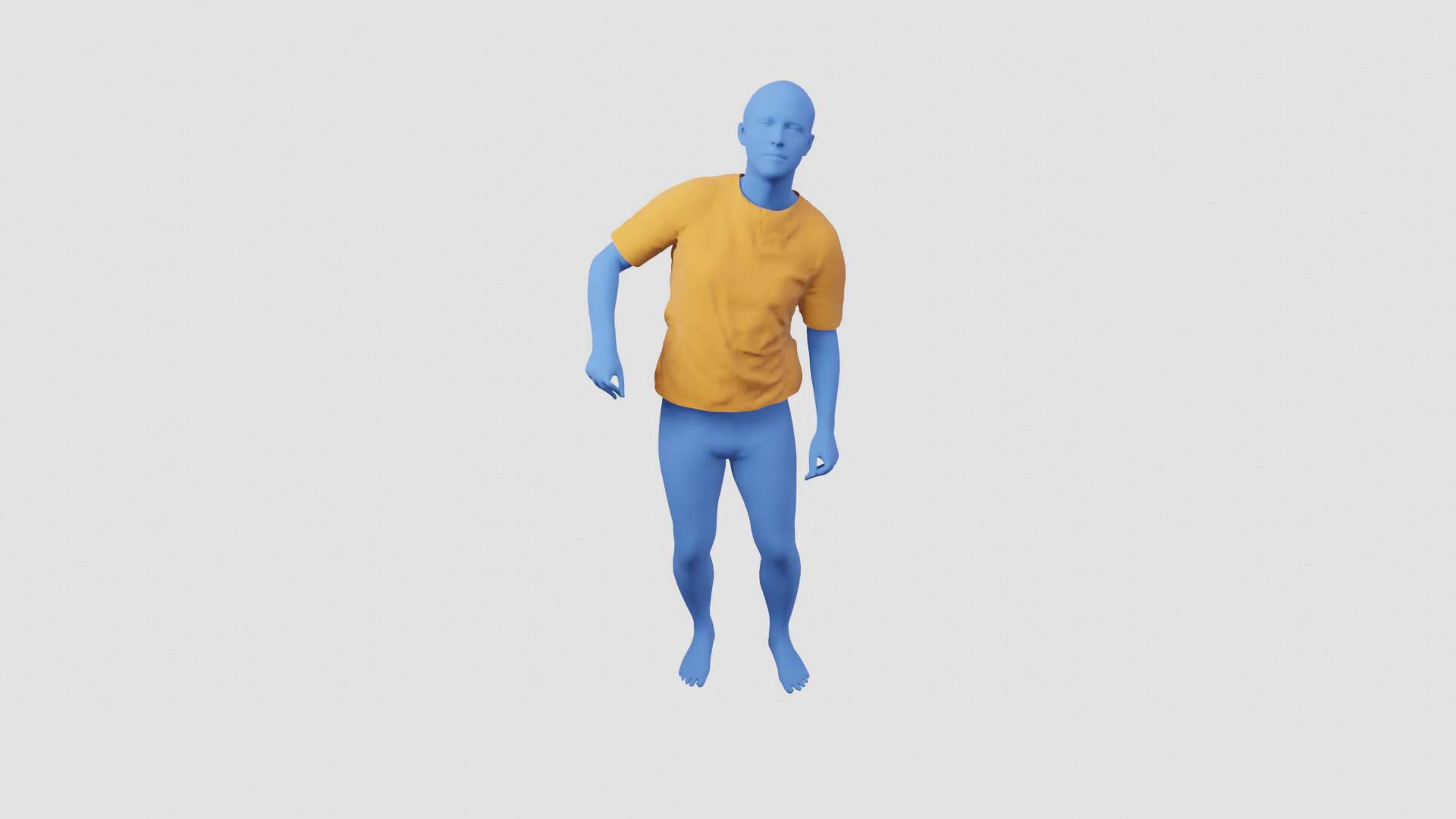}
  \end{subfigure}
  \hspace*{\fill}
  \begin{subfigure}{{\imWidth}\linewidth}
    \includegraphics[trim={\cropFrameYellowL} {\cropFrameYellowB} {\cropFrameYellowR} {\cropFrameYellowT}, clip, width=\linewidth]{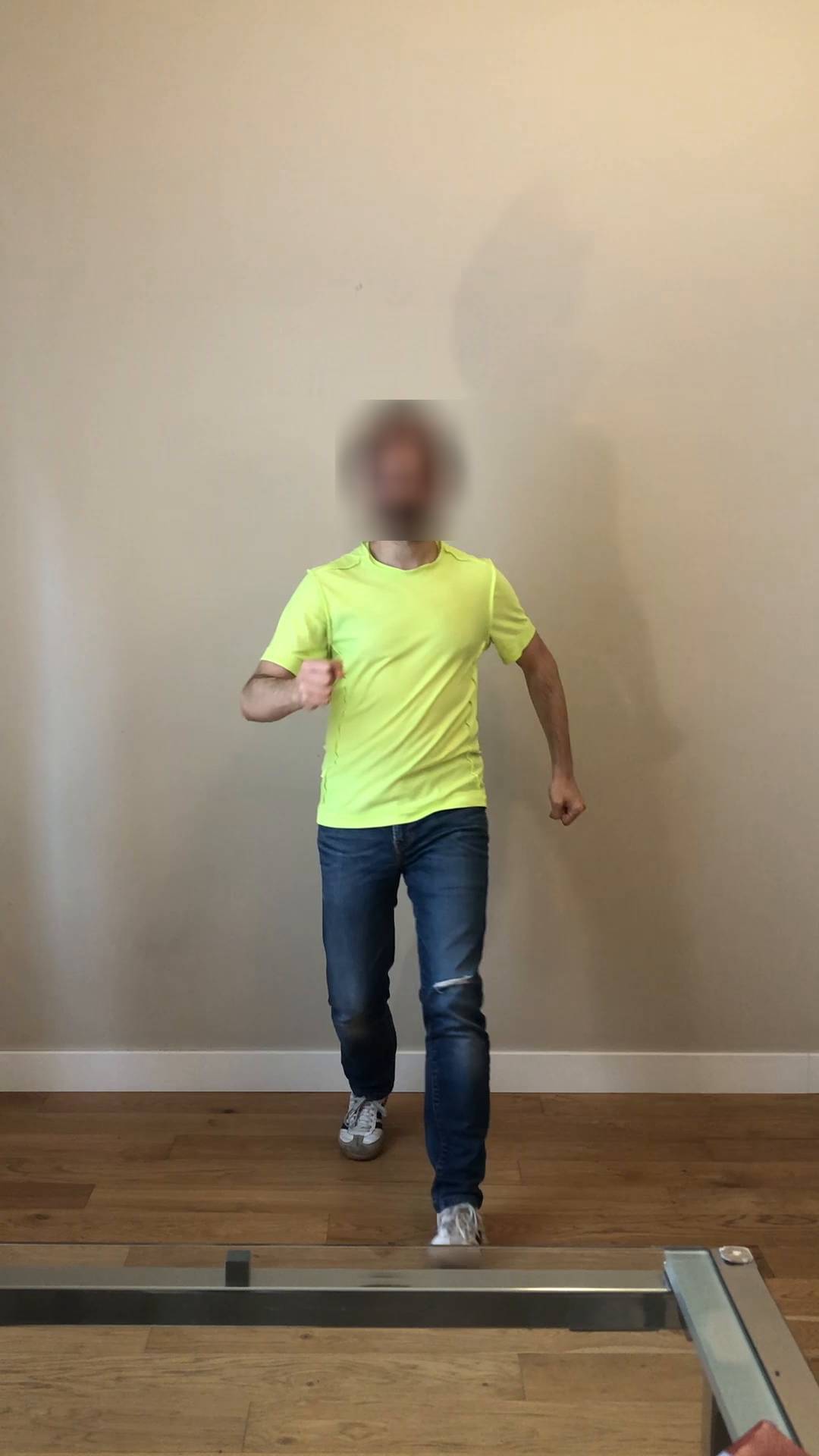}
  \end{subfigure}
  \begin{subfigure}{{\imWidth}\linewidth}
    \includegraphics[trim={\cropRenderL} {\cropRenderB} {\cropRenderR} {\cropRenderT}, clip,width=\linewidth]{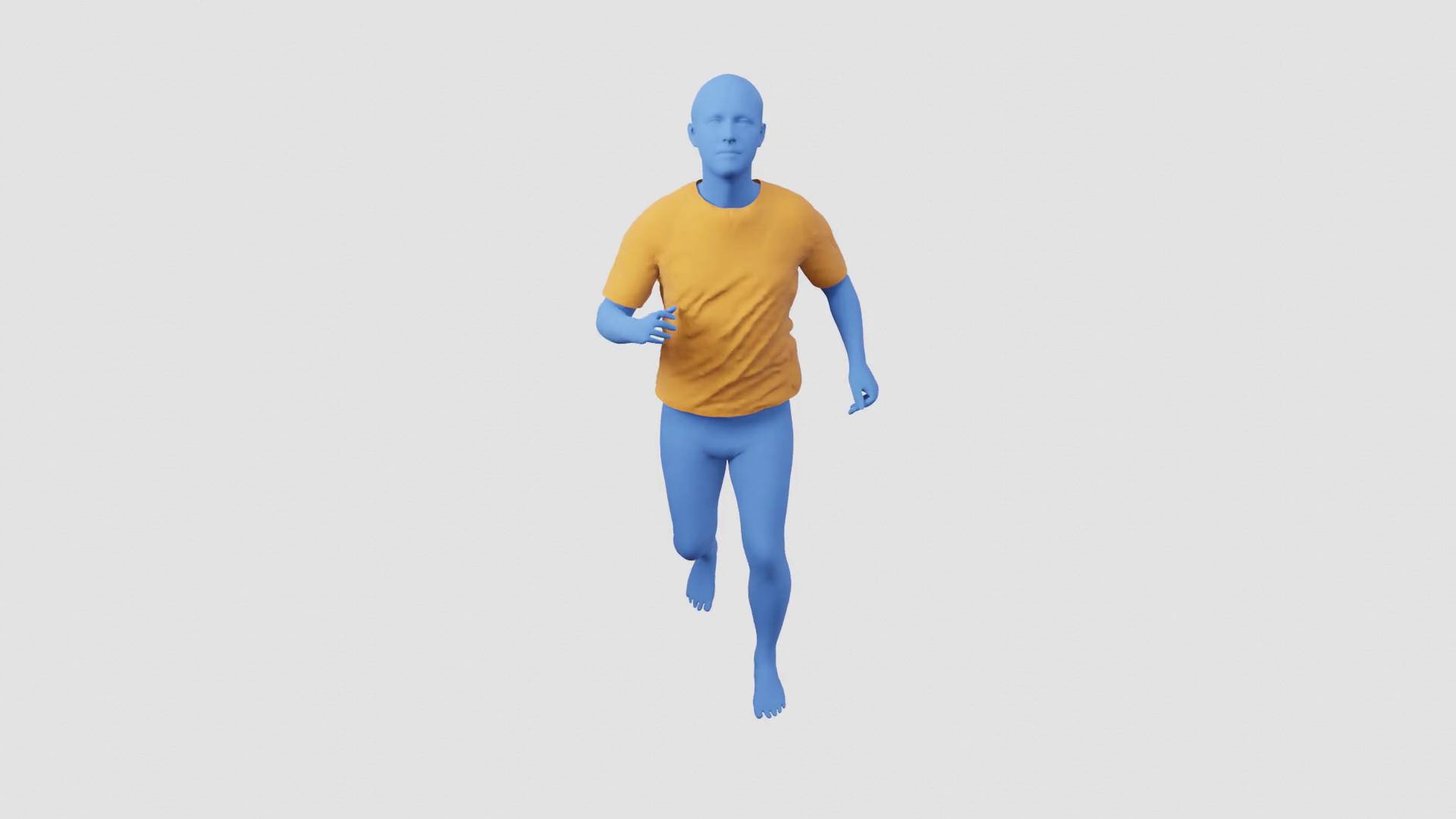}
  \end{subfigure}
  \hspace*{\fill}
  \begin{subfigure}{{\imWidth}\linewidth}
    \includegraphics[trim={\cropFrameYellowL} {\cropFrameYellowB} {\cropFrameYellowR} {\cropFrameYellowT}, clip, width=\linewidth]{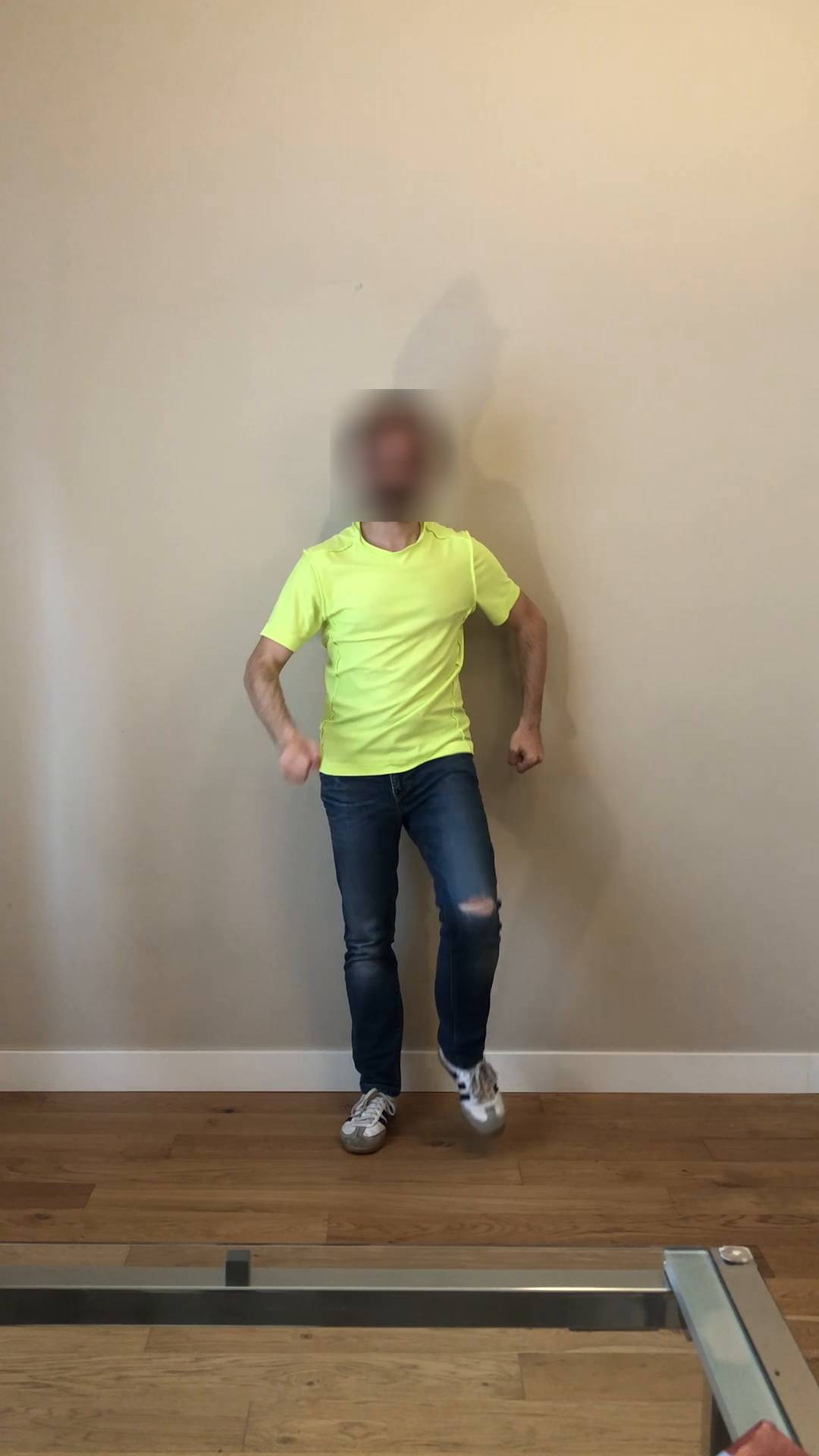}
  \end{subfigure}
  \begin{subfigure}{{\imWidth}\linewidth}
    \includegraphics[trim={\cropRenderL} {\cropRenderB} {\cropRenderR} {\cropRenderT}, clip,width=\linewidth]{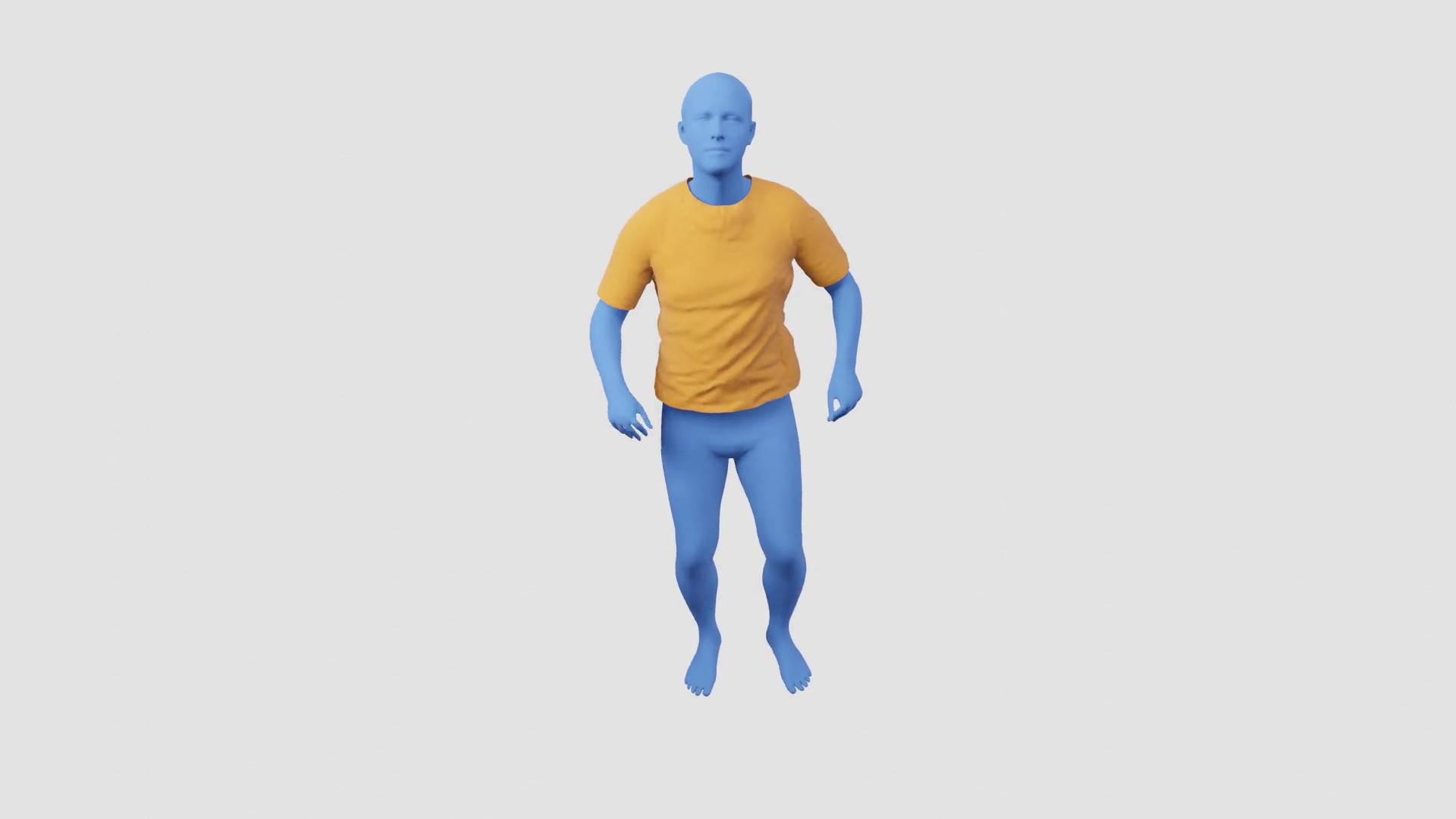}
  \end{subfigure}
  
  \vspace{0.2cm}
   \begin{subfigure}{{\imWidth}\linewidth}
    \includegraphics[trim={\cropFrameBrownL} {\cropFrameBrownB} {\cropFrameBrownR} {\cropFrameBrownT}, clip, width=\linewidth]{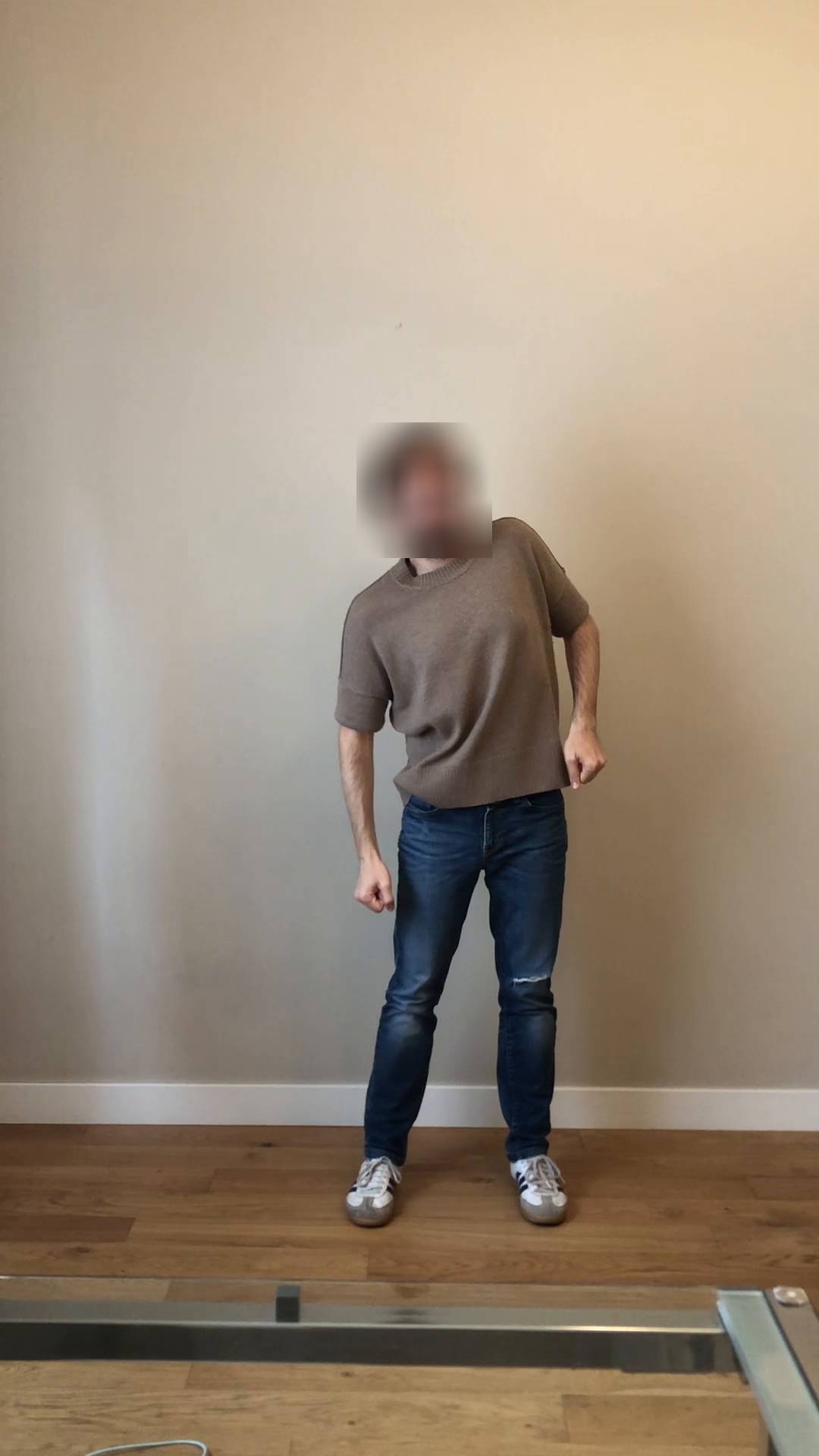}
  \end{subfigure}
  \begin{subfigure}{{\imWidth}\linewidth}
    \includegraphics[trim={\cropRenderL} {\cropRenderB} {\cropRenderR} {\cropRenderT}, clip,width=\linewidth]{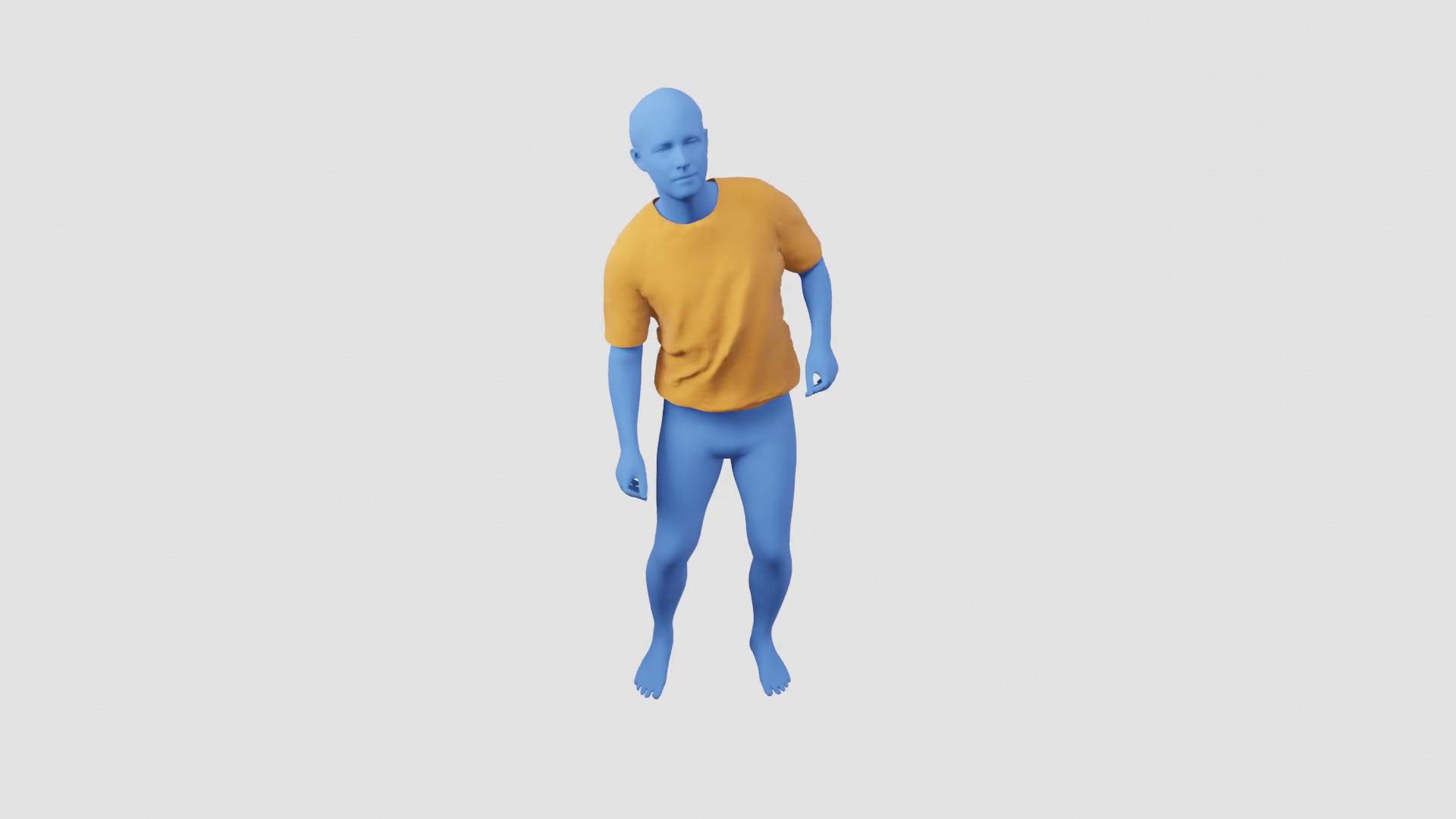}
  \end{subfigure}
  \hspace*{\fill}
  \begin{subfigure}{{\imWidth}\linewidth}
    \includegraphics[trim={\cropFrameBrownL} {\cropFrameBrownB} {\cropFrameBrownR} {\cropFrameBrownT}, clip, width=\linewidth]{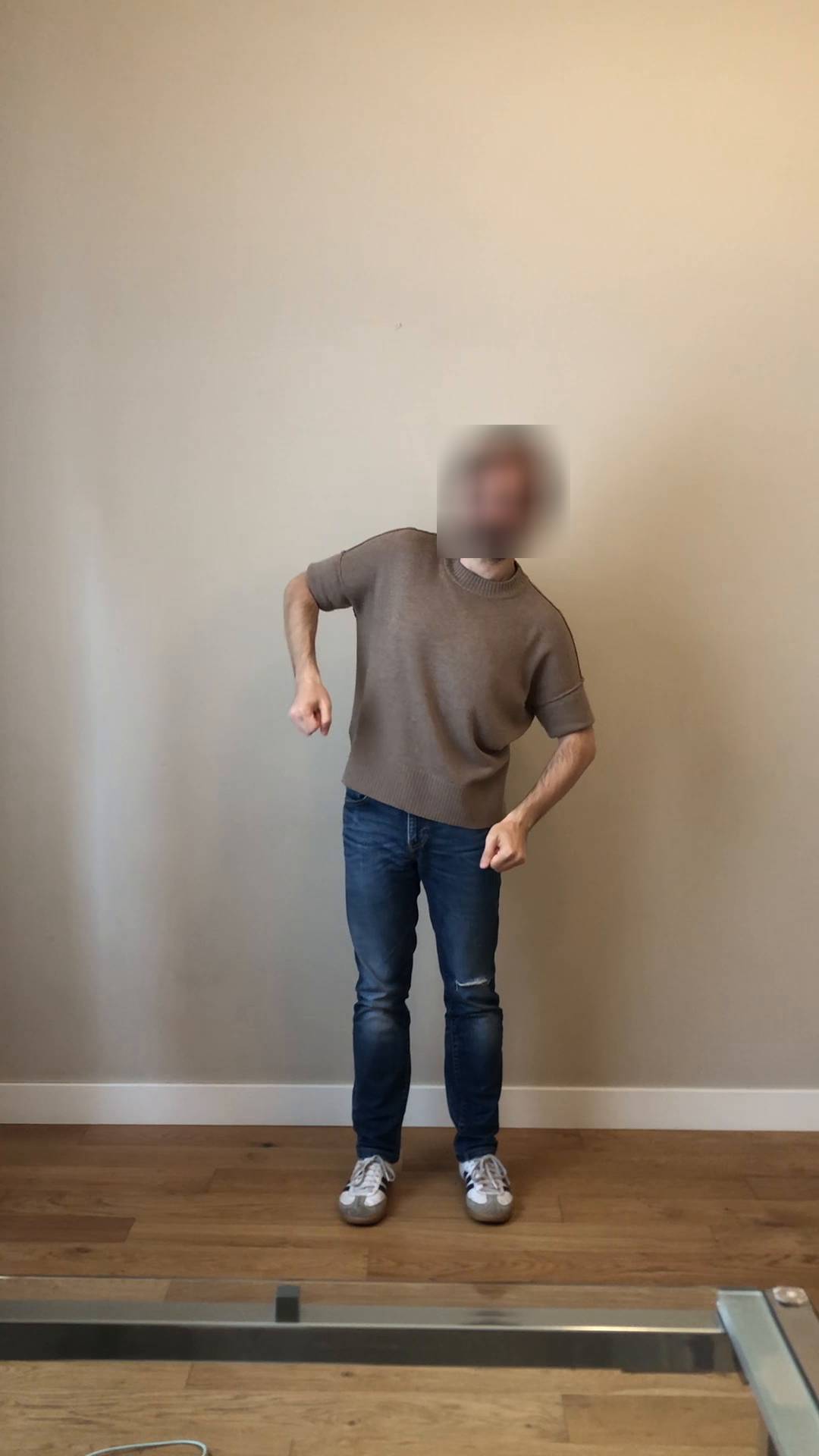}
  \end{subfigure}
  \begin{subfigure}{{\imWidth}\linewidth}
    \includegraphics[trim={\cropRenderL} {\cropRenderB} {\cropRenderR} {\cropRenderT}, clip,width=\linewidth]{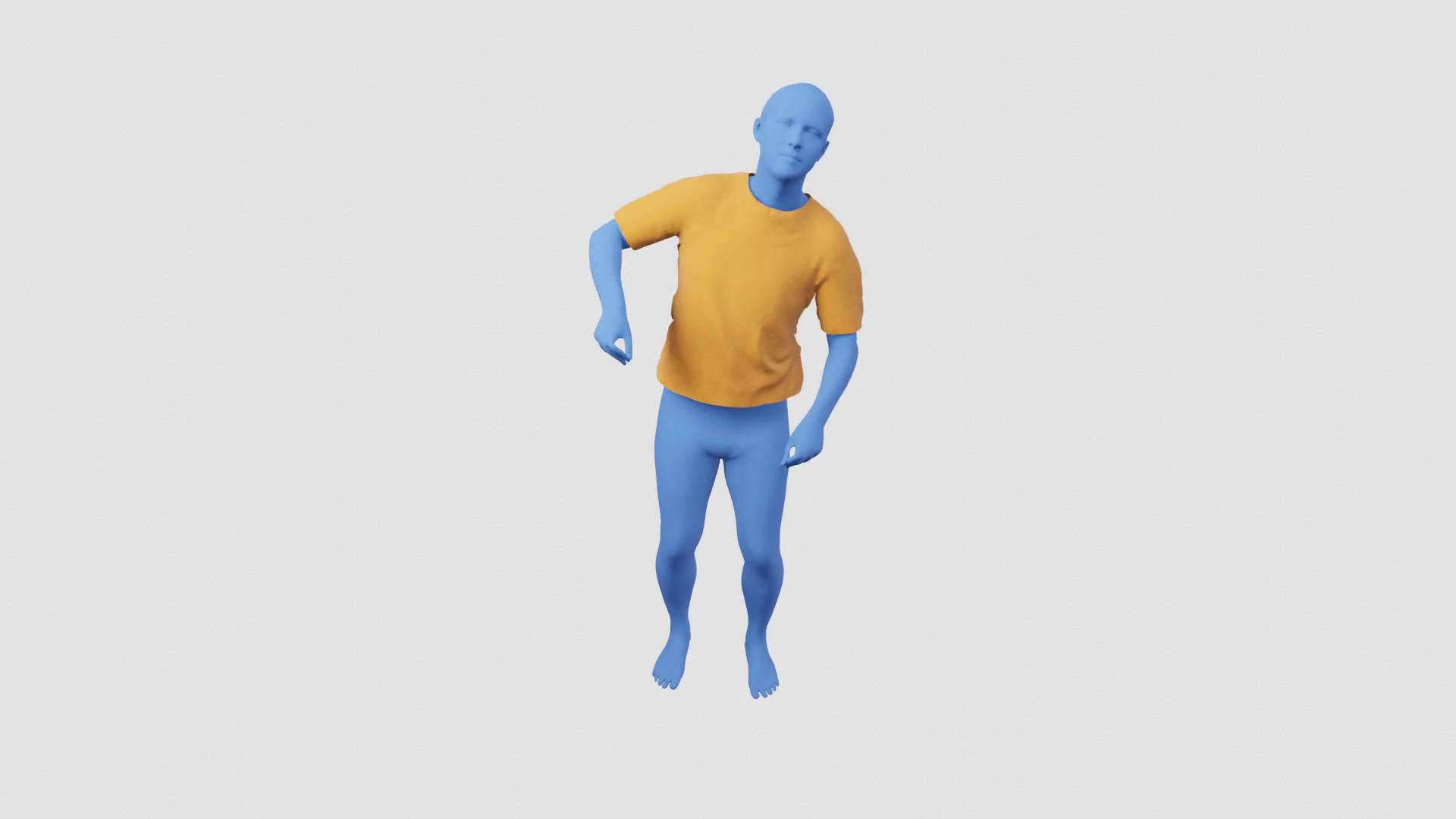}
  \end{subfigure}
  \hspace*{\fill}
   \begin{subfigure}{{\imWidth}\linewidth}
    \includegraphics[trim=250 160 200 450, clip, width=\linewidth]{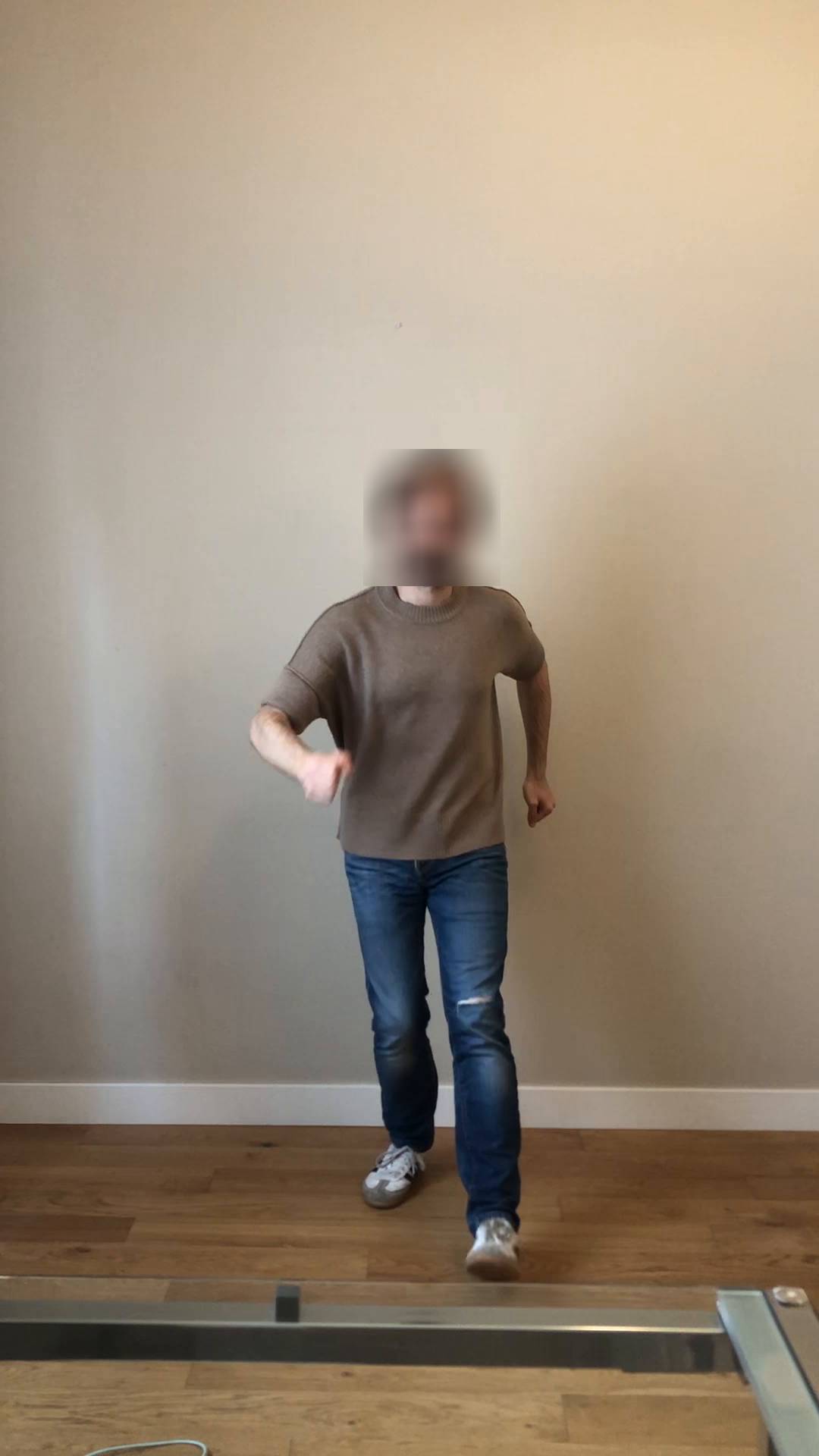}
  \end{subfigure}
  \begin{subfigure}{{\imWidth}\linewidth}
    \includegraphics[trim={\cropRenderL} {\cropRenderB} {\cropRenderR} {\cropRenderT}, clip,width=\linewidth]{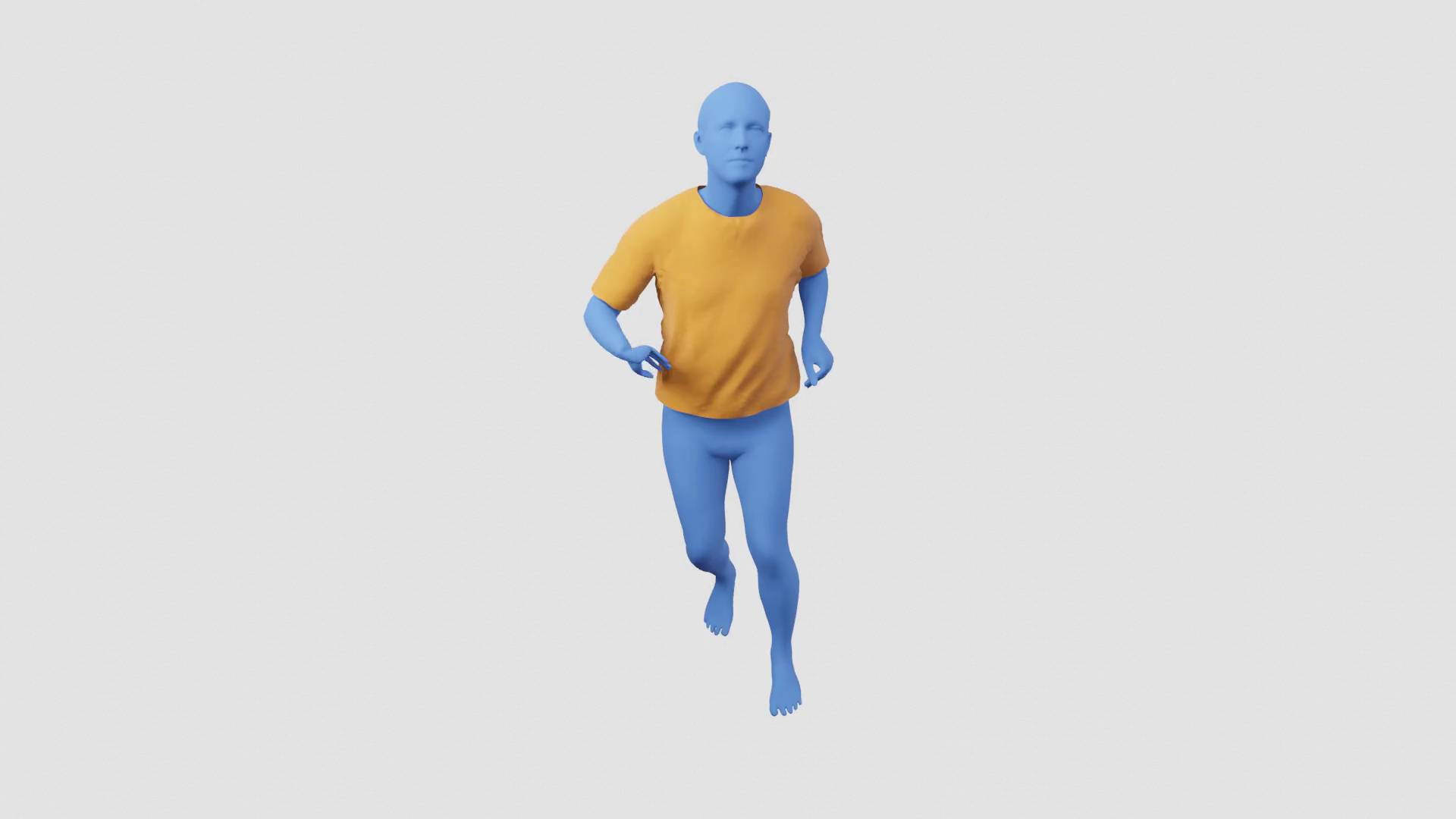}
  \end{subfigure}
  \hspace*{\fill}
  \begin{subfigure}{{\imWidth}\linewidth}
    \includegraphics[trim=250 160 200 450,clip, width=\linewidth]{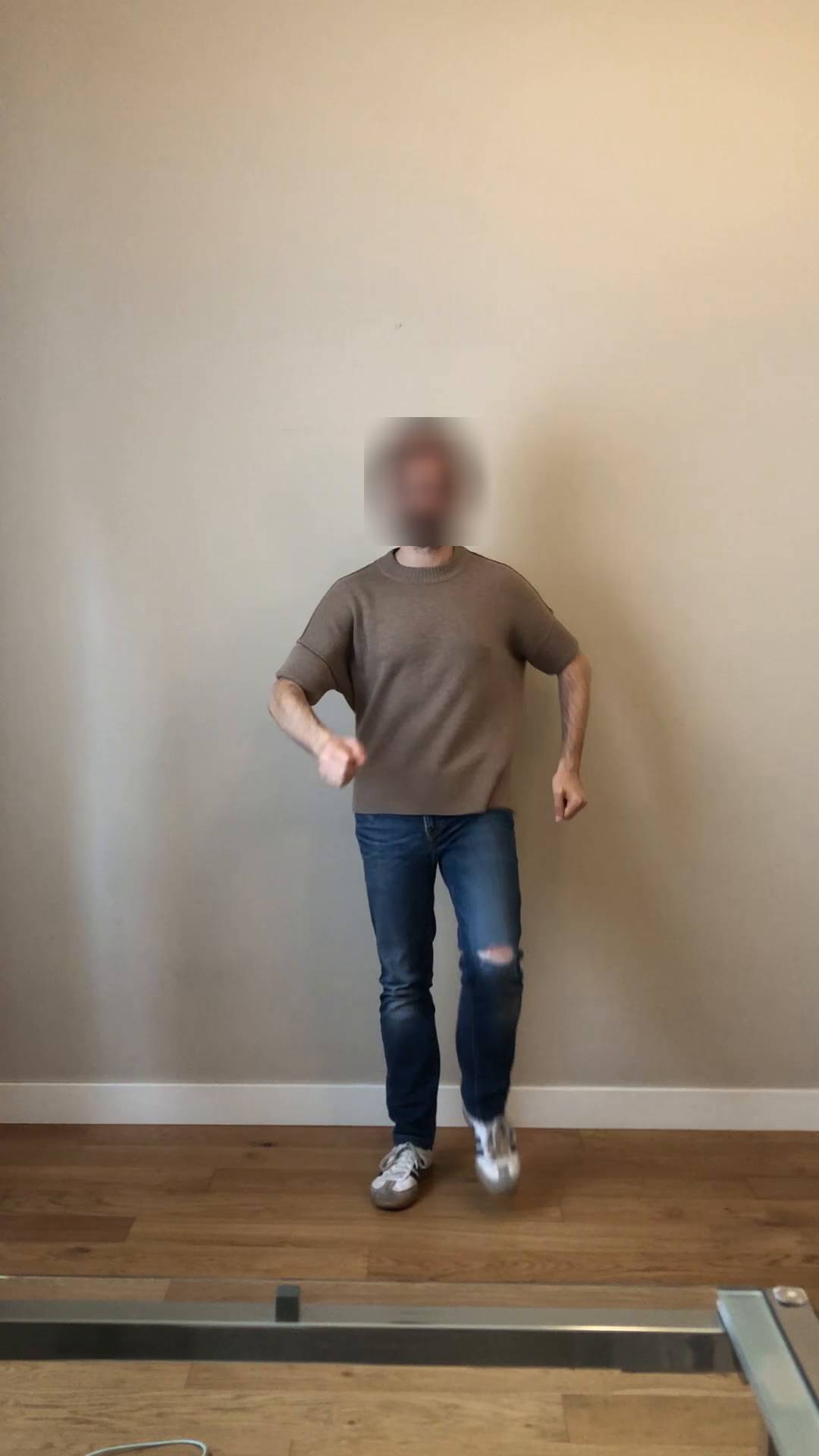}
  \end{subfigure}
  \begin{subfigure}{{\imWidth}\linewidth}
 \includegraphics[trim={\cropRenderL} {\cropRenderB} {\cropRenderR} {\cropRenderT},clip, width=\linewidth]{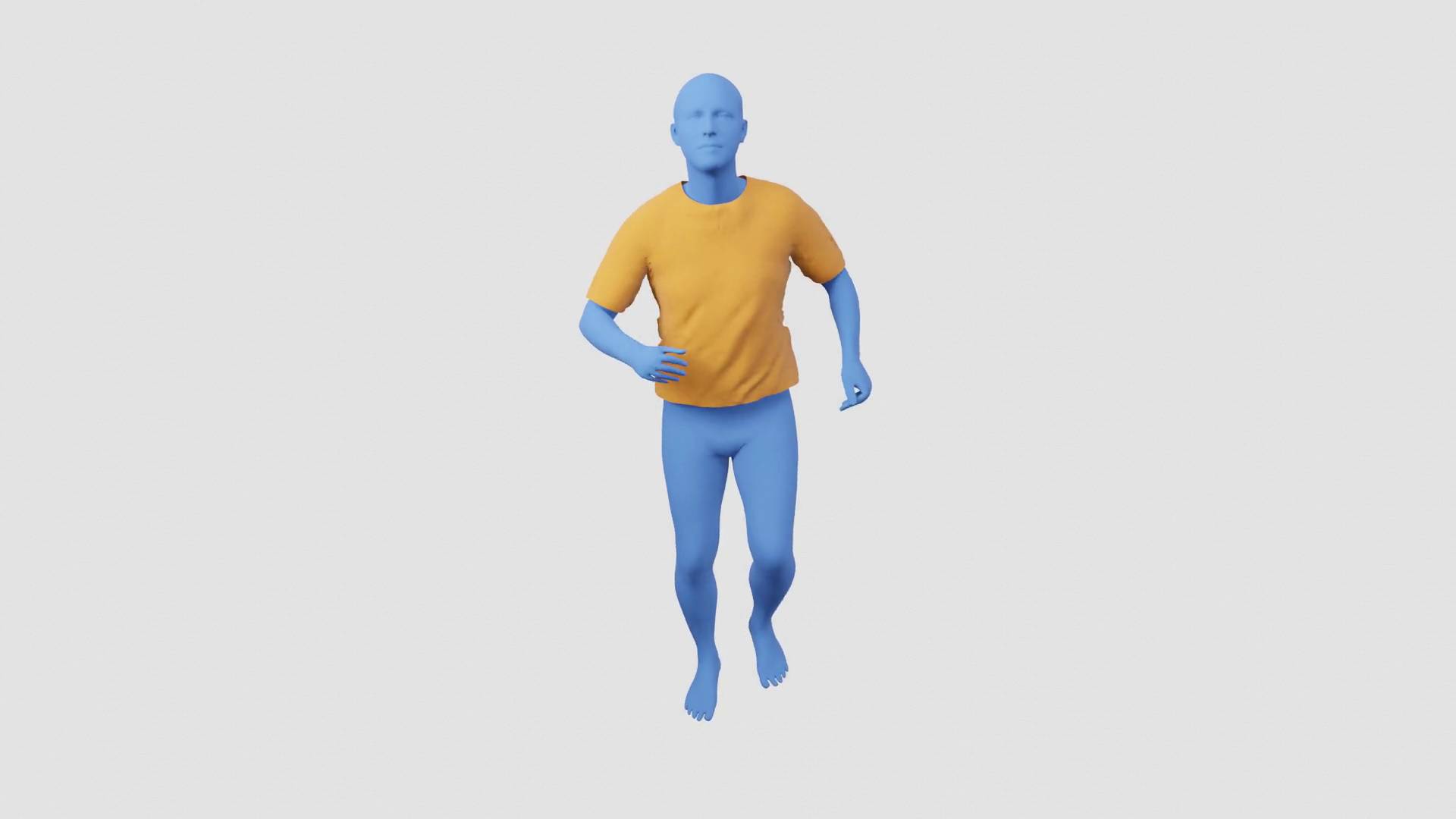}
  \end{subfigure}
	\caption{Qualitative results of our 3D Garment Reconstruction pipeline described in Section \ref{sec:garment-reconstruction}. Each row shows representative frames of a different garment of our dataset, and each column a similar pose. Notice how the brown t-shirt in the bottom row, made of a  thicker fabric, produces coarser wrinkles than the yellow t-shirt in the middle row. Our reconstructions faithfully capture these subtle differences, hence producing a garment-specific reconstruction details.}
	\label{fig:reconstruction}
\end{figure*}
}

%% file: sections/results.tex
\section{Results and Evaluation}
\subsection{Garment 3D Reconstruction}
\subsubsection*{Qualitative Evaluation}
Figure \ref{fig:reconstruction} and the supplementary video showcase a large variety of results of our 3D garment reconstruction pipeline introduced in Section \ref{sec:garment-reconstruction}.
To visually stress the quality of our results, we show side-by-side the corresponding input frame for each reconstruction. 

We have processed a dataset consisting in 3 different garments --grey, yellow, and brown t-shirts--, and 12 motion sequences for each garment. 
Importantly, each t-shirt is made of a different fabric, hence exhibiting different folds and wrinkles.
To highlight this, each row in Figure \ref{fig:reconstruction} depicts representative frames of a different garment, and each column similar poses.
It can be observed that the yellow t-shirt --a thin 100\% polyester sports t-shirt-- produces fine wrinkles; the grey t-shirt --100\% cotton t-shirt-- produces mid-scale wrinkles; and the brown t-shirt --made with thick fabric, similar to a sweater-- produces coarser wrinkles.

\input{sections/figures/figure-panoptic}
\input{sections/figures/figure-oleksander}

Additionally, in Figure \ref{fig:qualitative-panoptic} and in the supplementary video, we show qualitative results of our method in sequences of the Panoptic CMU dataset, and we compare to MonoClothCap \cite{xiang2020monocloth}. Notice that MonoClothCap is a monocular reconstruction approach that outputs clothed avatars encoded in a single mesh, which is not ideal to model garments.
Results demonstrate that our approach is capable of recovering finer wrinkle detail, while reconstructing an explicit garment layer.

Similarly, in Figure \ref{fig:oleks} we qualitatively compare our reconstruction results to both MonoClothCap \cite{xiang2020monocloth} and MonoPerfCap \cite{xu2018monoperfcap} in an outdoor sequence.
The visual quality of our reconstruction outperform these works, while we are able to also recover an explicit layer of the garment.

\subsubsection*{Quantitative Evaluation}
\label{sec:ablation}
\begin{figure}
    \centering
    \includegraphics[width=0.5\textwidth]{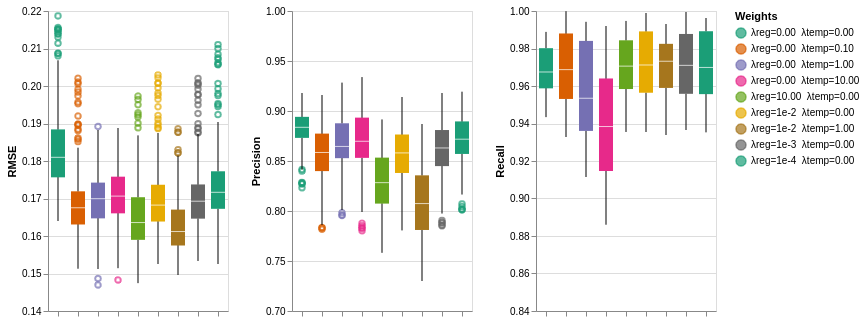}
    \caption{Quantitative evaluation for different values of weights $\uplambda_{\text{temp}_\mathbf{p}}$ and $\uplambda_{\text{reg}_\mathbf{p}}$ used in our garment reconstruction described in Section \ref{sec:parametric-garment}. We report RMSE, Precision and Recall scores.}
    \label{fig:ablationVaeCharts}
\end{figure}

Recovering 3D geometry from a single view is an ill-posed problem. To tackle it, we use multiple regularization terms in our reconstruction pipeline.
In Figure~\ref{fig:ablationVAE} and \ref{fig:ablationVaeCharts} we depict the effects of these terms, which we discuss below. 

We first analyze the effect of the terms $\mathcal{E}_{\text{temp}}$ and $\mathcal{E}_{\text{reg}}$ of our parametric garment fitting step (Section~\ref{sec:parametric-garment}). 
To this end, we use sequences of the BUFF dataset~\cite{zhang2017detailed} which provides sequences of 3D human textured meshes. To use such mesh data in our image-based pipeline, we render RGB and ground-truth surface normal in a $512 \times 512$ pixels image. 
We then process the resulting RGB frames using our method and a variety of values for $\uplambda_{\text{temp}_\mathbf{p}}$ and $\uplambda_{\text{reg}_\mathbf{p}}$. 
In Figure~\ref{fig:ablationVAE} we show representative frames of the sequence \texttt{shortlong\_hips\_96} reconstructed using different regularization choices (rows one to four). 
The first row shows how the mesh may degenerate when no regularization is enforced. The second row shows how a slight regularization of the VAE latent space cannot completely prevent the mesh degeneration, while inconsistencies between consecutive frames may appear causing flickering artifacts (see differences between frame 89 and 90). 
Temporal regularization (third row) can prevent flickering artifacts at the expense of a worse fit to the ground-truth silhouette. 
A careful combination of both regularization strategies may prevent degeneration and flickering while not greatly affecting the silhouette fitting.

We also analyze the effect of regularization mechanisms implemented at the fine wrinkle extraction step (Section~\ref{sec:fine-wrinkles}). 
Row 5 at Figure~\ref{fig:ablationVAE} shows how the optimization process can explode when no regularization on the free-vertex movements is enforced. 
Finally, row 6 show that our final choice for weights closely matches the ground truth normals.

{
\newcommand{\TitleHSepTeaser}{0.15cm}
\begin{figure}
\begin{tikzpicture}[every node/.style={inner sep=0,outer sep=0}]%
  \node (image) at (1,0) 
  {\includegraphics[trim={5cm 0 0 0},clip,width=0.4\textwidth]{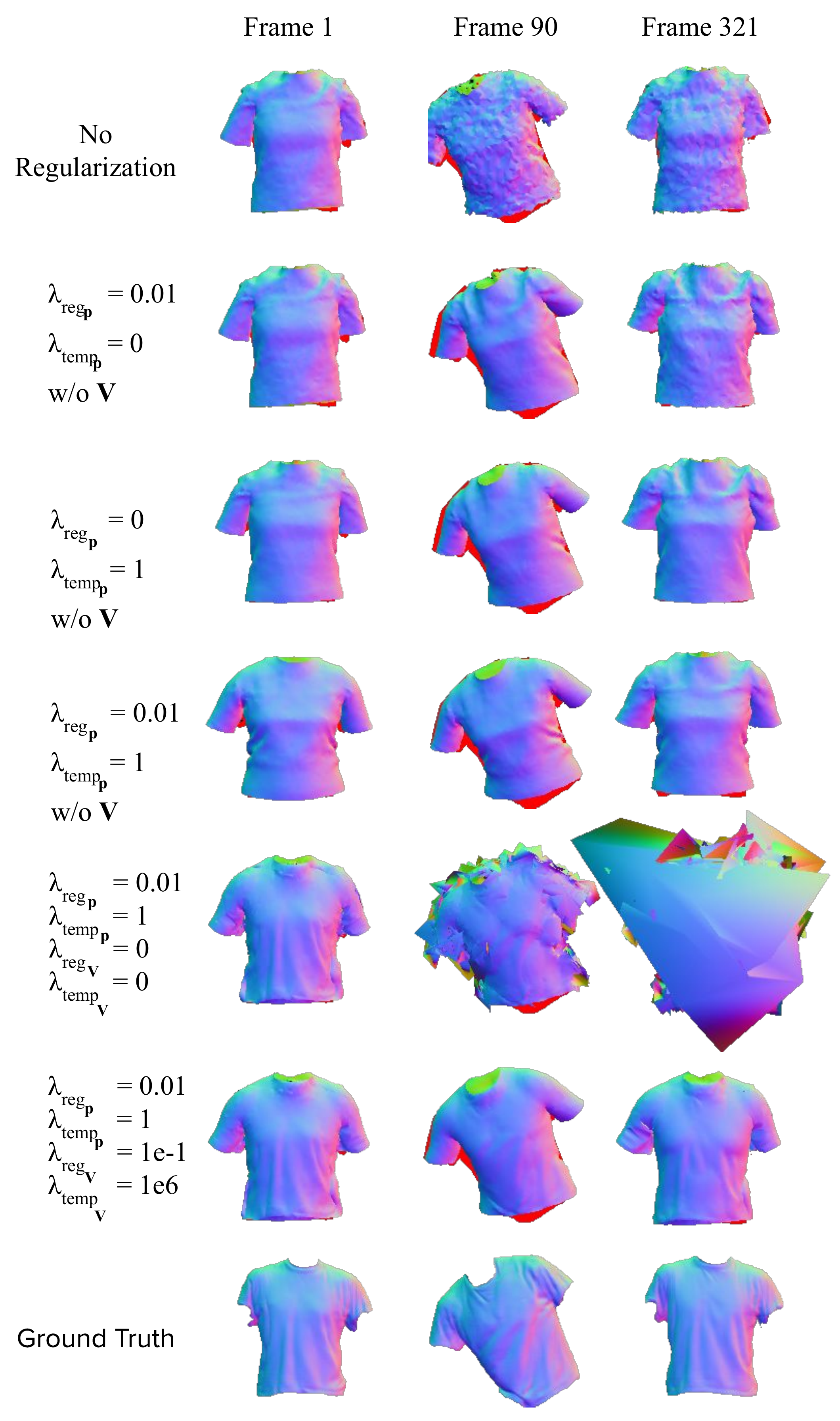}};
 \node at (-3.4,6.5) {\small{No}};
 \node at (-3.4,6.1) {\small{Regularization}};
 \node[align=left] (A) at (-3.4,4.5) {\small{$\uplambda_{\text{reg}_\mathbf{p}} ~~= 0.01$}};
  \node[align=left, below=0.4cm of A.west,anchor=west] (B) {\small{$\uplambda_{\text{temp}_\mathbf{p}} = 0.00$}};
  \node[align=left, below=0.4cm of B.west,anchor=west] (C) {\small{w/o $\mathbf{V}$}};
  \node[align=left, below=1.4cm of C.west,anchor=west] (D) {\small{$\uplambda_{\text{reg}_\mathbf{p}} ~~= 0.00$}};
  \node[align=left, below=0.4cm of D.west,anchor=west] (E) {\small{$\uplambda_{\text{temp}_\mathbf{p}} = 1.00$}};
  \node[align=left, below=0.4cm of E.west,anchor=west] (F) {\small{w/o $\mathbf{V}$}};
  \node[align=left, below=1.4cm of F.west,anchor=west] (G) {\small{$\uplambda_{\text{reg}_\mathbf{p}} ~~= 0.01$}};  
  \node[align=left, below=0.4cm of G.west,anchor=west] (H) {\small{$\uplambda_{\text{temp}_\mathbf{p}} = 1.00$}};
  \node[align=left, below=0.4cm of H.west,anchor=west] (I) {\small{w/o $\mathbf{V}$}};
  \node[align=left, below=1.2cm of I.west,anchor=west] (J) {\small{$\uplambda_{\text{reg}_\mathbf{p}} ~~= 0.01$}};  
  \node[align=left, below=0.4cm of J.west,anchor=west] (K) {\small{$\uplambda_{\text{temp}_\mathbf{p}} = 1.00$}};
  \node[align=left, below=0.4cm of K.west,anchor=west] (L) {\small{$\uplambda_{\text{reg}_\mathbf{V}} ~~= 0.00$}};  
  \node[align=left, below=0.4cm of L.west,anchor=west] (M) {\small{$\uplambda_{\text{temp}_\mathbf{V}} = 0.00$}};  
  \node[align=left, below=1.15cm of M.west,anchor=west] (N) {\small{$\uplambda_{\text{reg}_\mathbf{p}} ~~= 0.01$}};
  \node[align=left, below=0.4cm of N.west,anchor=west] (O) {\small{$\uplambda_{\text{temp}_\mathbf{p}} = 1.00$}};
  \node[align=left, below=0.4cm of O.west,anchor=west] (P) {\small{$\uplambda_{\text{reg}_\mathbf{V}} ~~= 0.10$}};
  \node[align=left, below=0.4cm of P.west,anchor=west] (Q) {\small{$\uplambda_{\text{temp}_\mathbf{V}} = 1$e$6$}};  
 \node at (-3.4,-6.8) {\small{Ground Truth}};
 \end{tikzpicture}
  \caption{Effect of the value choice for $\uplambda_{\text{temp}_\mathbf{p}}$ and $\uplambda_{\text{reg}_\mathbf{p}}$ (Rows 1 to 4). The lack of regularization causes the resulting meshes to degenerate. Effect of $\uplambda_{\text{temp}_\mathbf{V}}$ and $\uplambda_{\text{reg}_\mathbf{V}}$ (Rows 5 to 6). The lack of regularization can causes the vertices to explode. }
  \label{fig:ablationVAE}
  \vspace{-2mm}
\end{figure}
}
\input{sections/figures/figure-regression}

Furthermore, we use the rendered ground-truth normal maps to quantitatively evaluate our reconstructions. 
In Figure~\ref{fig:ablationVaeCharts} we provide results in terms of RMSE of the normal angle difference and precision and recall of the generated reconstructed silhouettes. 
Our choice of $\uplambda_{\text{temp}_\mathbf{p}} = 0.01$ and $\uplambda_{\text{reg}_\mathbf{p}} = 1$ achieves best results in terms of RMSE and competitive results of recall, meaning our reconstructions overlap well with the ground-truth silhouettes.

\subsection{Garment 3D Regressor}
To validate that the garment deformation regressor introduced in Section \ref{sec:regressor} generalizes to human poses not present in our dataset, we test the regressor using publicly available motion capture data.
Our goal is to demonstrate that our model can produce realistic clothing animations for arbitrary motion input.

To this end, we use motion sequences from the publicly available AMASS \cite{AMASS:ICCV:2019} and BUFF \cite{zhang2017detailed} datasets, and feed them into a regressor trained with the grey t-shirt dataset.
Figure \ref{fig:qualitative-regression} and the supplementary video show realistic animations generated with our approach.

\subsection{Implementation Details}
\label{sec:implementation-details}
\Change{In this section we provide additional implementation and performance details of our method.
We have implemented PERGAMO in a desktop machine with an Intel Core i7-5820K CPU, a Nvidia RTX 3080 Ti GPU and 16GB of DDR4 RAM.}

\noindent\textbf{Parametric Garment Fitting (Section \ref{sec:parametric-garment})}.
In our experiments we set $\uplambda_{\text{sil}_\mathbf{p}} = 1$ and start with $\uplambda_{\text{coarse}_\mathbf{p}} = 0$ and increase it to $1$ after half of the optimization iterations. In particular, we run the optimization loop for a total of 200 iterations for the first frame of a sequence using gradient descent. Afterwards, we always initialize $\mathbf{p}_t$ using $\mathbf{p}_{t-1}$ and, thanks to this, we can reduce the number of optimization iterations to 20.
We use regularization weights $\uplambda_{\text{temp}_\mathbf{p}} = 1$ and $\uplambda_{\text{reg}_\mathbf{p}} = 0.01$, and explore the effect of these in the ablation study conducted in Section~\ref{sec:ablation}.

\noindent\textbf{Fine Wrinkle Extraction (Section \ref{sec:fine-wrinkles}).}
We fix $\uplambda_{\text{fine}_\mathbf{V}} = 6.5 \cdot 10^2$ and use regularization weights $\uplambda_{\text{temp}_\mathbf{V}} = 10^6$, $\uplambda_{\text{edge}_\mathbf{V}} = 10^4$ and $\uplambda_{\text{reg}_\mathbf{V}} = 0.1$. We explore the effect of these in the ablation study conducted in Section~\ref{sec:ablation}.
In our experiments we run the optimization loop for a total of 200 iterations for the first frame of a sequence using gradient descent. Afterwards, we always initialize $\mathbf{V}_t$ using $\mathbf{V}_{t-1}$ which allows us to reduce the number of optimization iterations to 20.

\Change{The values for the different loss parameters $\uplambda$ have been set empirically using trade-off between the faithfulness and stability of the reconstructions.
Selected values have been used for all reconstructions shown in this paper, including the sequences for the 3 different types of shirts, the Panoptic CMU Dataset \cite{joo2018total}, and MonoPerfCap~\cite{xu2018monoperfcap} sequences used for providing comparisons.}

\Change{Our full reconstruction pipeline for one input frame takes around $2.615 \pm 0.039$ seconds.}%

\noindent\textbf{Garment Regressor (Section \ref{sec:regressor})}.
We use a vanilla MLP (\Change{$R():\mathbb{R}^{10} \rightarrow \mathbb{R}^{N_{\text{G}} \times 3}$}) with 3 hidden layers \Change{ ($HL_{1}():\mathbb{R}^{10} \rightarrow \mathbb{R}^{N_{\text{G}}}$, $HL_{2}():\mathbb{R}^{N_{\text{G}}} \rightarrow \mathbb{R}^{N_{\text{G}}}$ and $HL_{3}():\mathbb{R}^{N_{\text{G}}} \rightarrow \mathbb{R}^{N_{\text{G}} \times 3}$, with $N_{\text{G}} =4,424$ for the garment in our results). Each layer is fully-connected with the previous layer, and the first and second layers are followed by a LeakyRELU activation (slope = 0.1) and a Dropout (10\%) layer.}
We use an Adam optimizer for 100 epochs with batch size of 64, and decreasing learning rate from 5e-3 to 1e-5.
\Change{Finally, we have noticed that normalizing the inputs and outputs of our regressor using the training set statistics can ease the training process.}

\Change{After training, predicting the deformations for a single pose takes around $0.5 \pm 10^{-2}$ ms, which is coherent when compared to existing data-driven methods \cite{santesteban2019virtualtryon, patel2020tailor}. 
Posing the resulting vertices took $1 \pm 0.1$ ms. Lastly, the postprocess to remove potential residual collisions takes $33 \pm 3$ ms.}

%% file: sections/figures/figure-panoptic.tex
{
\newcommand{\imWidth}{0.3215}
\newcommand{\cropOursL}{90}
\newcommand{\cropOursB}{0}
\newcommand{\cropOursR}{100}
\newcommand{\cropOursT}{0}
\newcommand{\cropOursLL}{22}
\newcommand{\cropOursBB}{0}
\newcommand{\cropOursRR}{25}
\newcommand{\cropOursTT}{0}
\newcommand{\cropMCCL}{150}
\newcommand{\cropMCCB}{100}
\newcommand{\cropMCCR}{1030}
\newcommand{\cropMCCT}{80}
\newcommand{\cropOriginalL}{670}
\newcommand{\cropOriginalB}{95}
\newcommand{\cropOriginalR}{500}
\newcommand{\cropOriginalT}{80}
\begin{figure}[t]
 \begin{subfigure}{{\imWidth}\linewidth}
    \includegraphics[trim={\cropOriginalL} {\cropOriginalB} {\cropOriginalR} {\cropOriginalT}, clip, width=\linewidth]{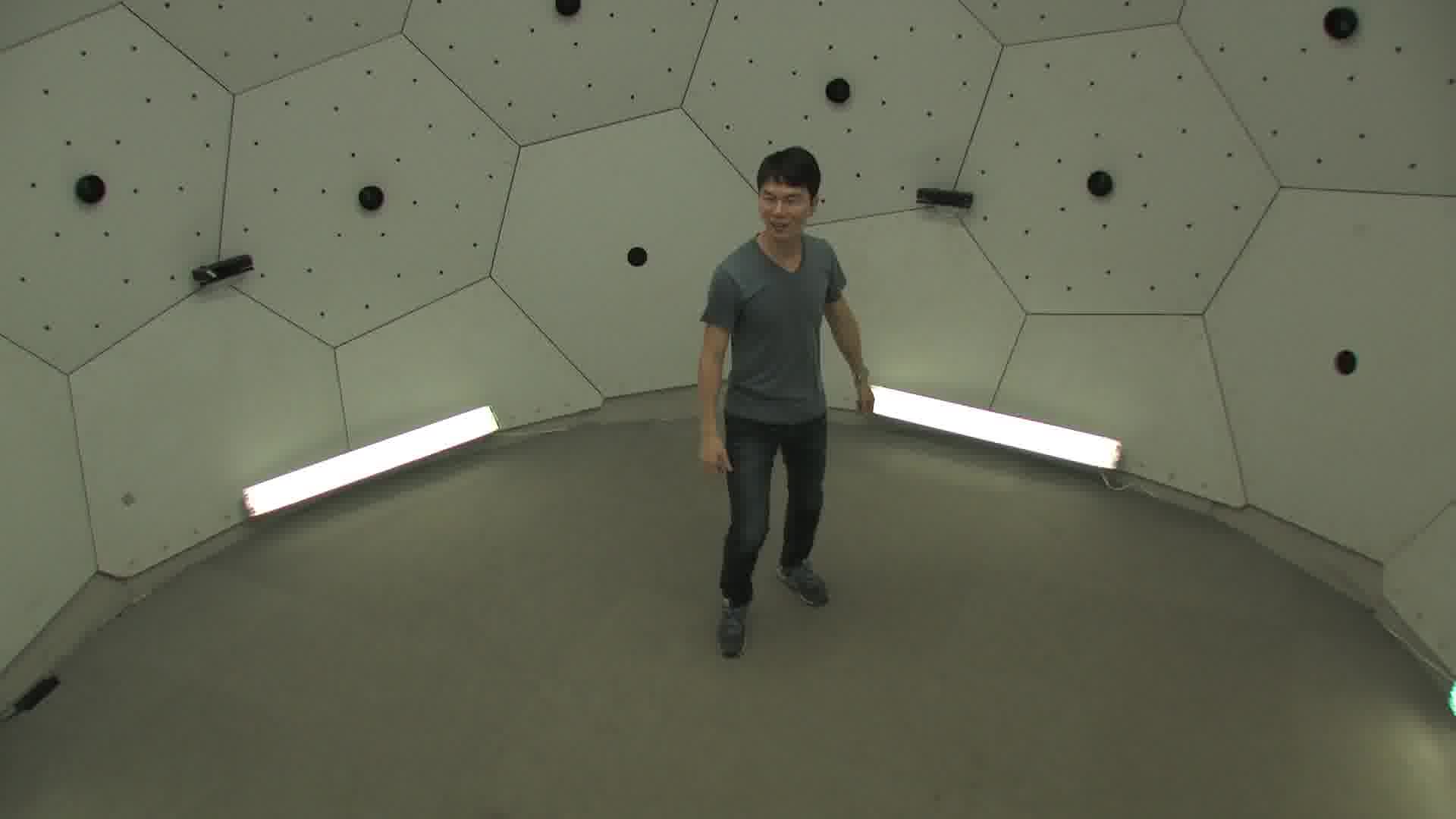}
  \end{subfigure}
  \hspace*{\fill}
   \begin{subfigure}{{\imWidth}\linewidth}
    \includegraphics[trim={\cropMCCL} {\cropMCCB} {\cropMCCR} {\cropMCCT}, clip, width=\linewidth]{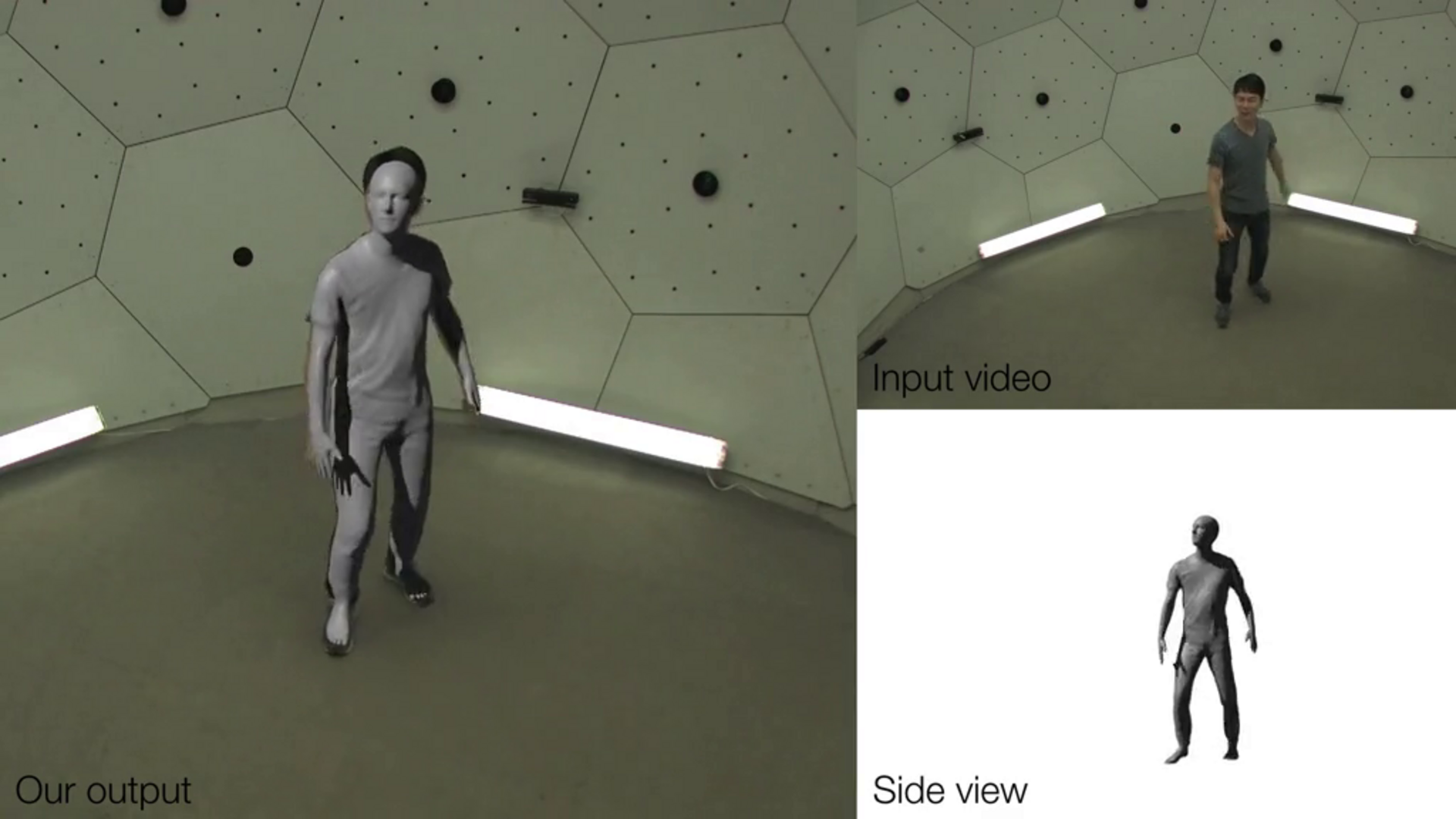}
  \end{subfigure}
  \hspace*{\fill}
 \begin{subfigure}{{\imWidth}\linewidth}
    \includegraphics[trim={\cropOursLL} {\cropOursBB} {\cropOursRR} {\cropOursTT}, clip, width=\linewidth]{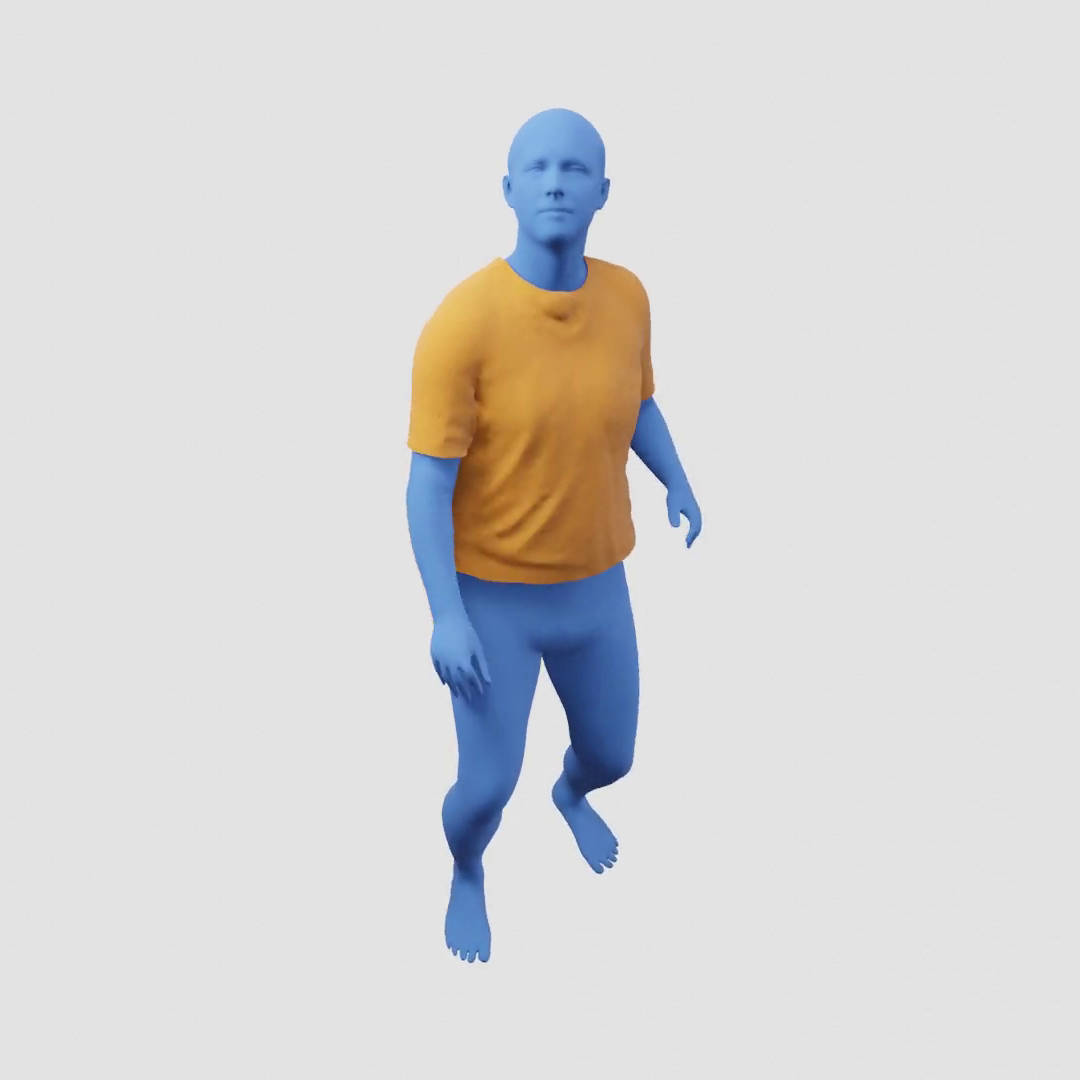}
  \end{subfigure}
  
   \vspace{0.1cm}
 \begin{subfigure}{{\imWidth}\linewidth}
    \includegraphics[trim={\cropOriginalL} {\cropOriginalB} {\cropOriginalR} {\cropOriginalT}, clip, width=\linewidth]{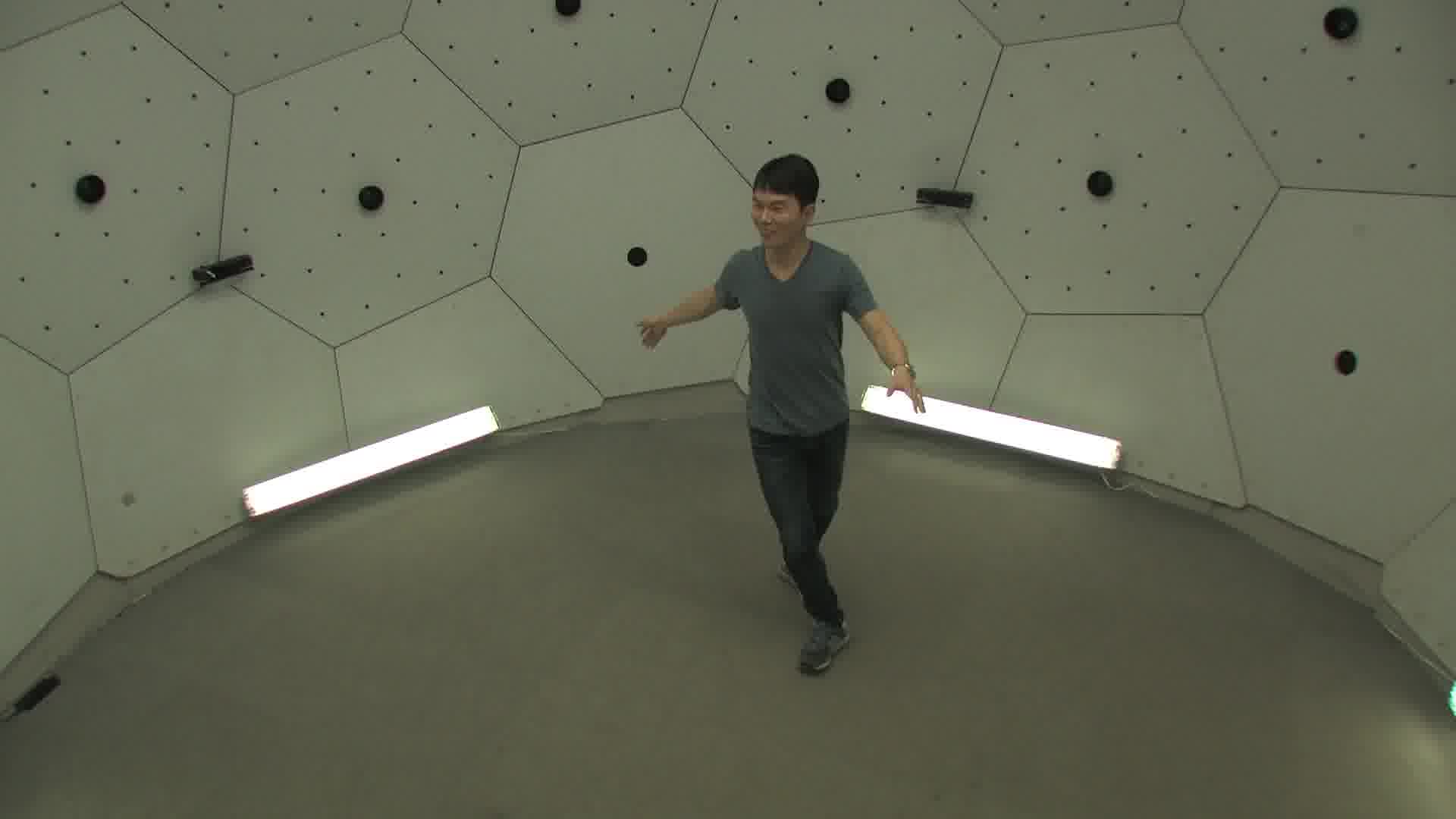}
  \end{subfigure}
    \hspace*{\fill}
   \begin{subfigure}{{\imWidth}\linewidth}
    \includegraphics[trim={\cropMCCL} {\cropMCCB} {\cropMCCR} {\cropMCCT}, clip, width=\linewidth]{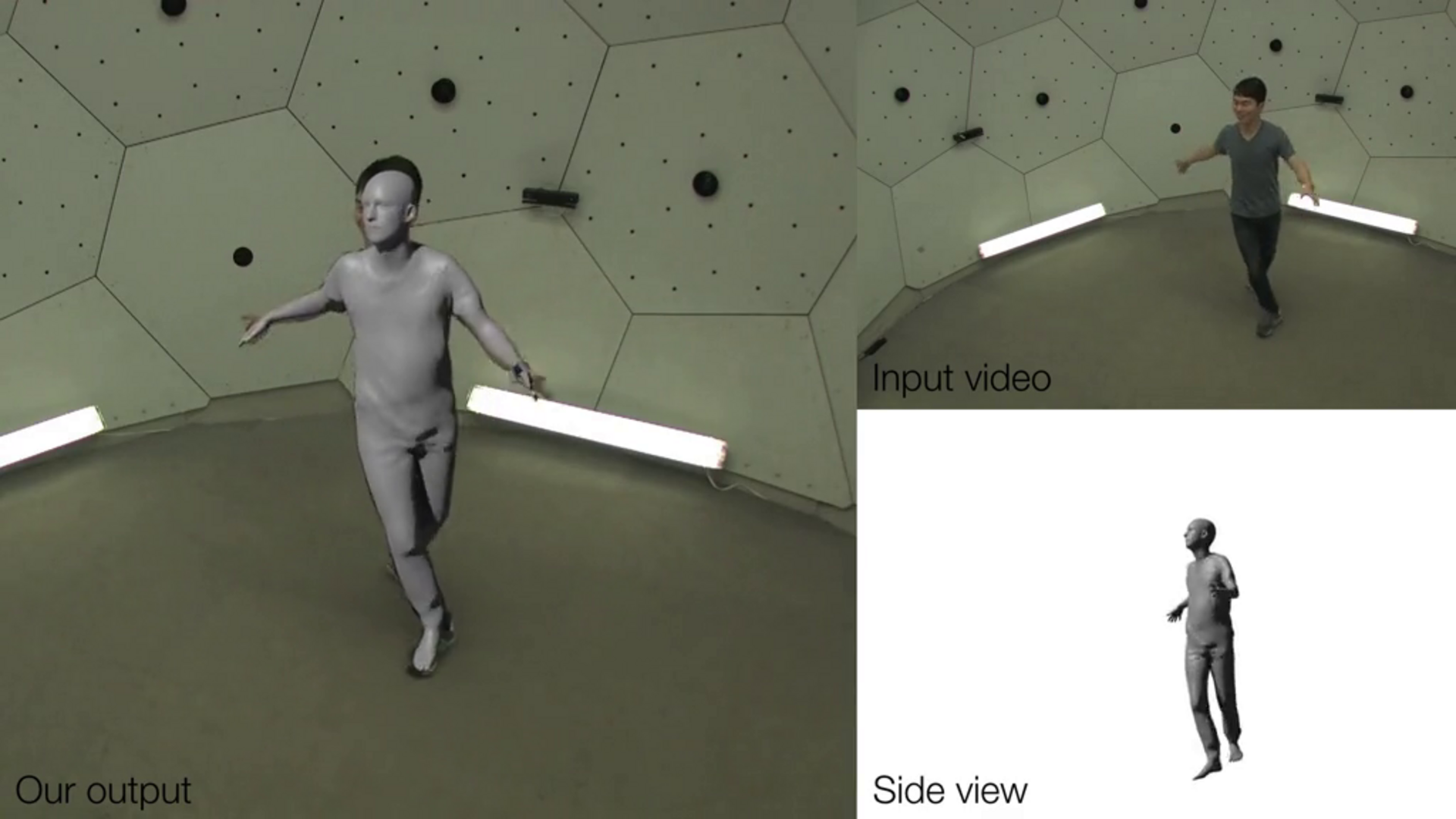}
  \end{subfigure}
  \hspace*{\fill}
   \begin{subfigure}{{\imWidth}\linewidth}
    \includegraphics[trim={\cropOursLL} {\cropOursBB} {\cropOursRR} {\cropOursTT}, clip, width=\linewidth]{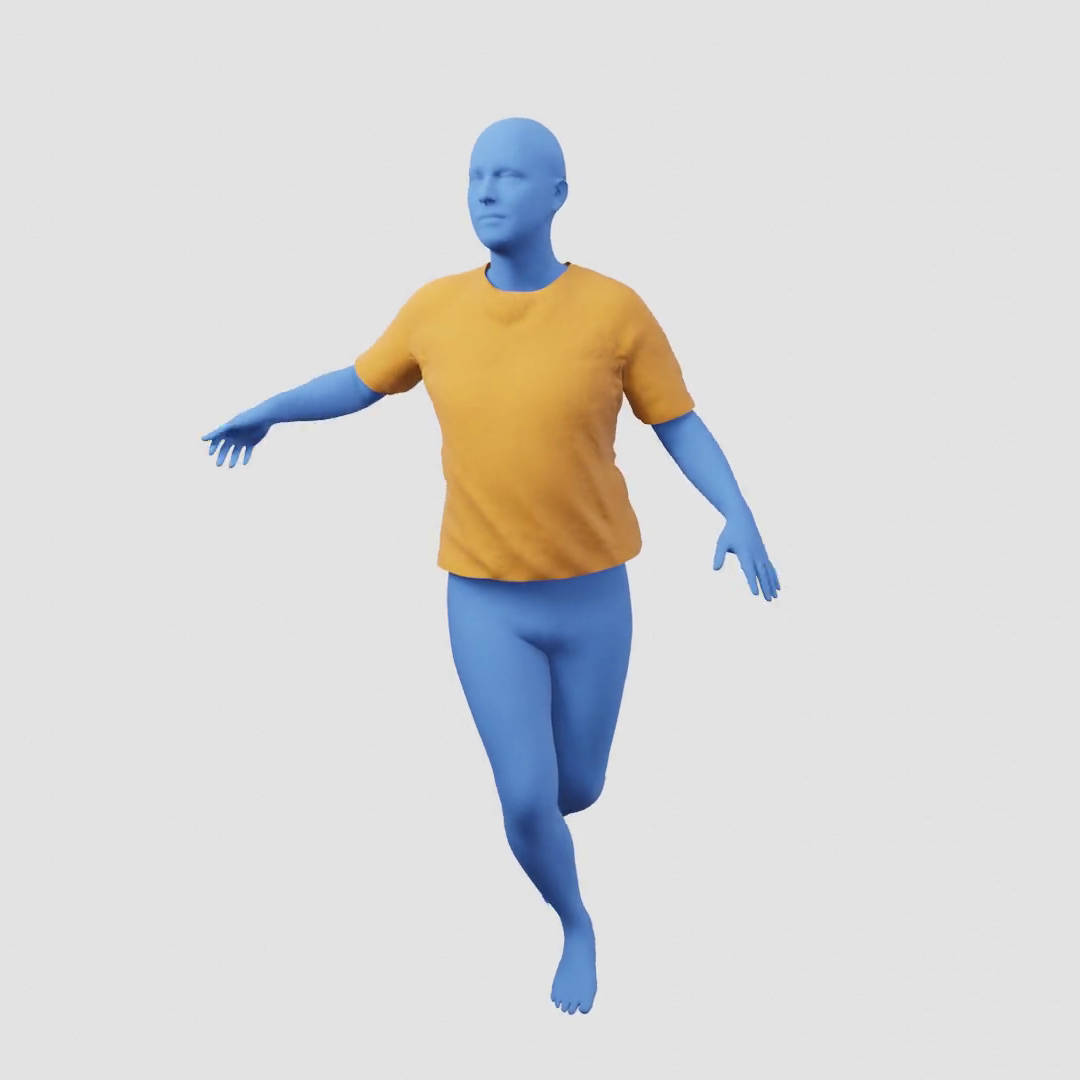}
  \end{subfigure}

 \vspace{0.1cm}
 \begin{subfigure}{{\imWidth}\linewidth}
    \includegraphics[trim={\cropOriginalL} {\cropOriginalB} {\cropOriginalR} {\cropOriginalT}, clip, width=\linewidth]{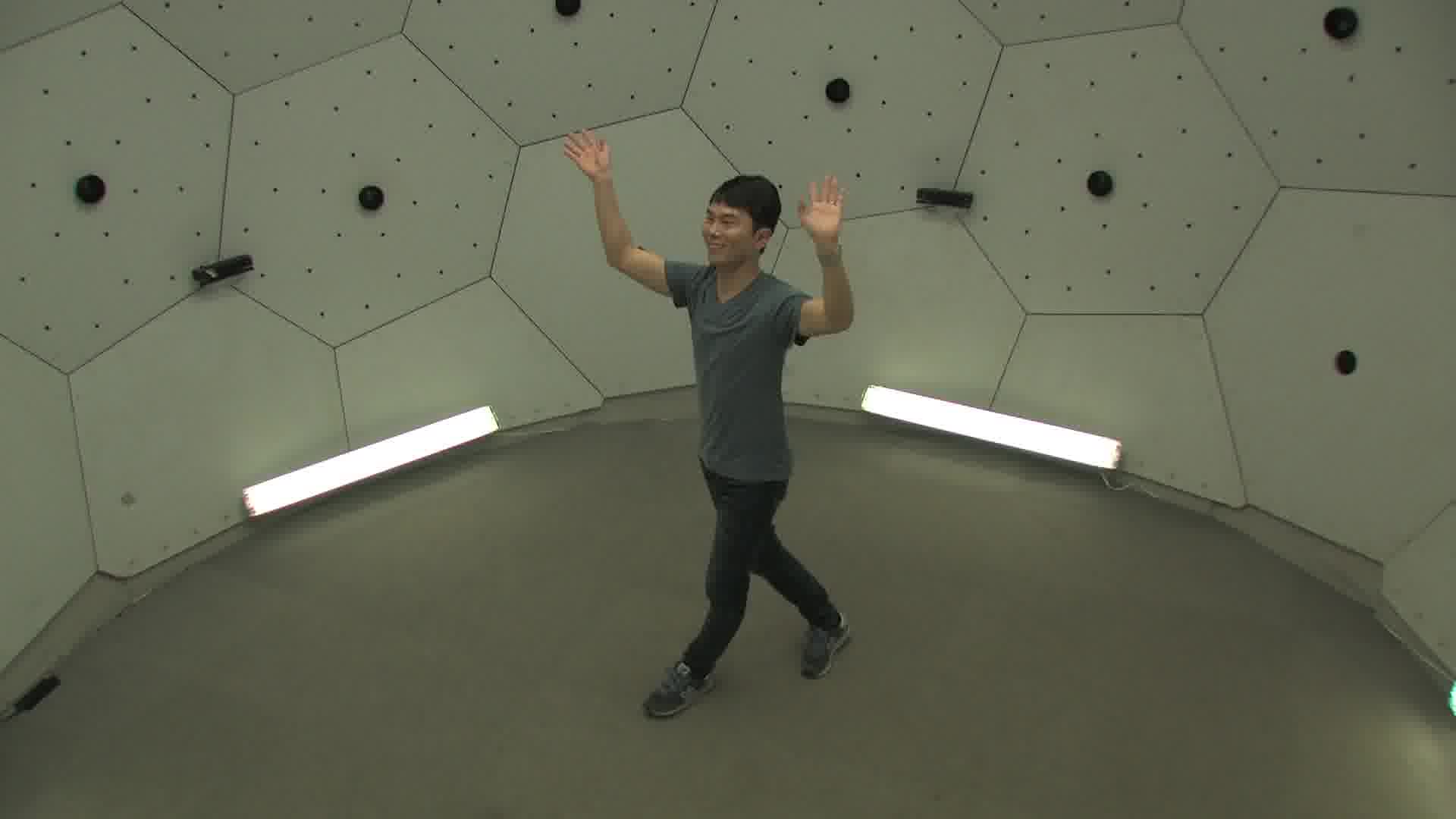}
    \caption*{\scriptsize{Input}}
  \end{subfigure}
    \hspace*{\fill}
   \begin{subfigure}{{\imWidth}\linewidth}
    \includegraphics[trim={\cropMCCL} {\cropMCCB} {\cropMCCR} {\cropMCCT}, clip, width=\linewidth]{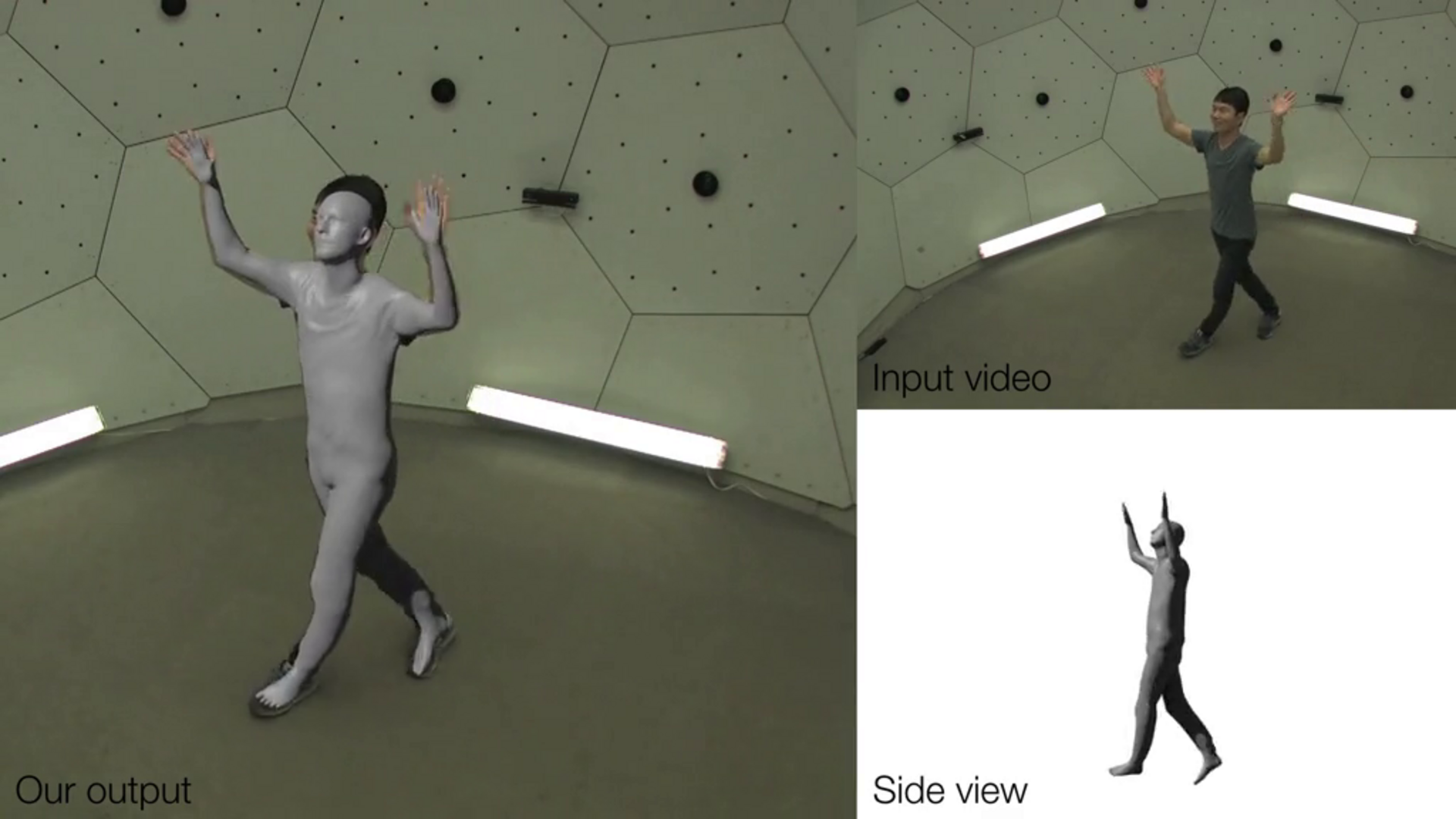}
    \caption*{\scriptsize{MonoClothCap\cite{xiang2020monocloth}}}
  \end{subfigure}
  \hspace*{\fill}
   \begin{subfigure}{{\imWidth}\linewidth}
    \includegraphics[trim={\cropOursLL} {\cropOursBB} {\cropOursRR} {\cropOursTT}, clip, width=\linewidth]{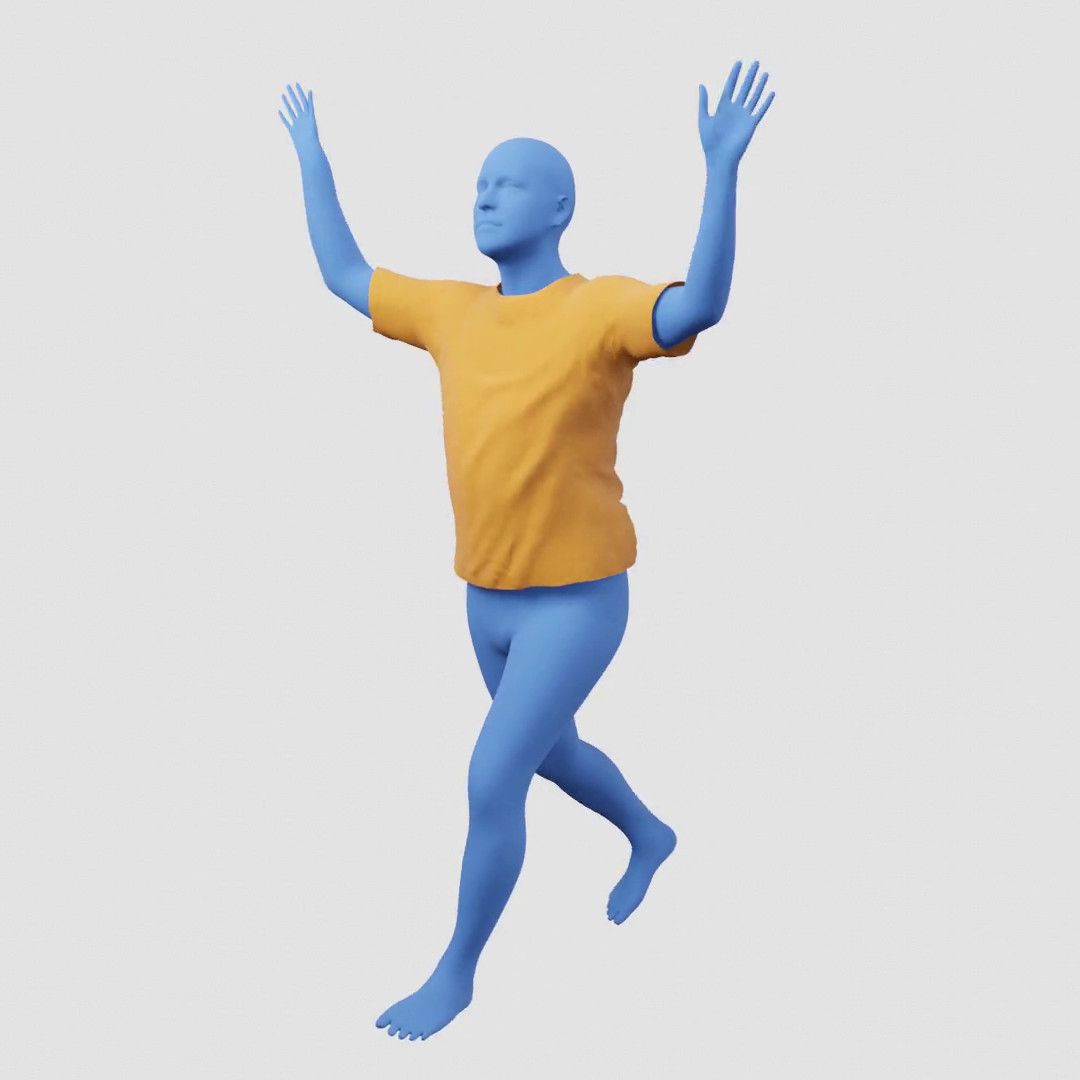}
    \caption*{\scriptsize{PERGAMO (Ours)}}
  \end{subfigure}
  \caption{Qualitative comparison to MonoClothCap \cite{xiang2020monocloth} in CMU Panoptic Dataset \cite{xiang2019monocular,joo2018total}}
  \label{fig:qualitative-panoptic}
\end{figure}
}

%% file: sections/figures/figure-oleksander.tex
{
\newcommand{\imWidth}{0.23}
\newcommand{\cropOursL}{80}
\newcommand{\cropOursB}{50}
\newcommand{\cropOursR}{100}
\newcommand{\cropOursT}{0}
\newcommand{\cropOursLL}{19}
\newcommand{\cropOursBB}{12}
\newcommand{\cropOursRR}{23}
\newcommand{\cropOursTT}{0}
\newcommand{\cropMPCL}{1270}
\newcommand{\cropMPCB}{50}
\newcommand{\cropMPCR}{250}
\newcommand{\cropMPCT}{570}
\newcommand{\cropMCCL}{1270}
\newcommand{\cropMCCB}{600}
\newcommand{\cropMCCR}{250}
\newcommand{\cropMCCT}{20}
\newcommand{\cropOriginalL}{620}
\newcommand{\cropOriginalB}{95}
\newcommand{\cropOriginalR}{500}
\newcommand{\cropOriginalT}{70}
\begin{figure}[t]
 \begin{subfigure}{{\imWidth}\linewidth}
    \includegraphics[trim={\cropOriginalL} {\cropOriginalB} {\cropOriginalR} {\cropOriginalT}, clip, width=\linewidth]{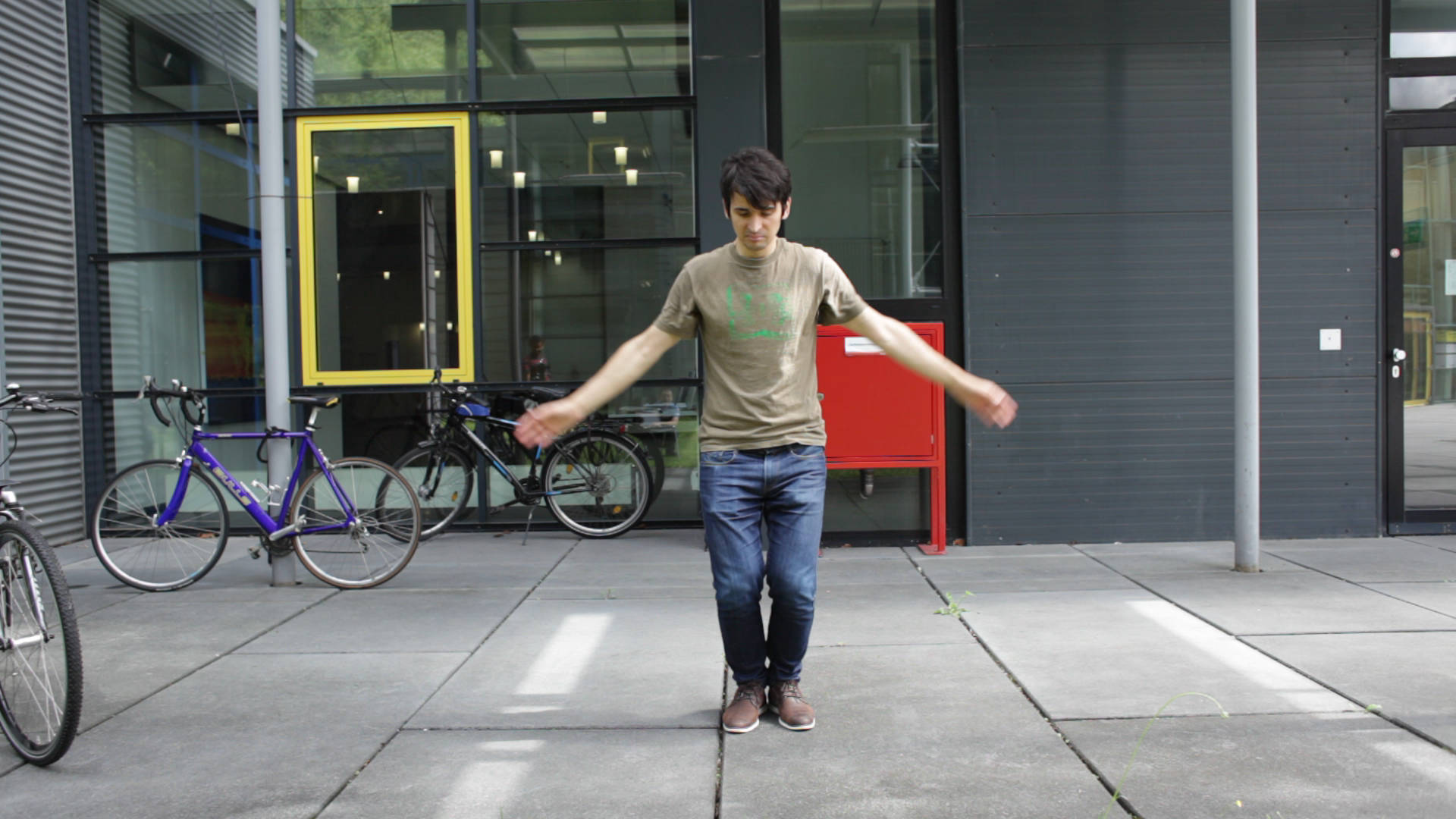}
  \end{subfigure}
  \hspace*{\fill}
 \begin{subfigure}{{\imWidth}\linewidth}
    \includegraphics[trim={\cropMPCL} {\cropMPCB} {\cropMPCR} {\cropMPCT}, clip, width=\linewidth]{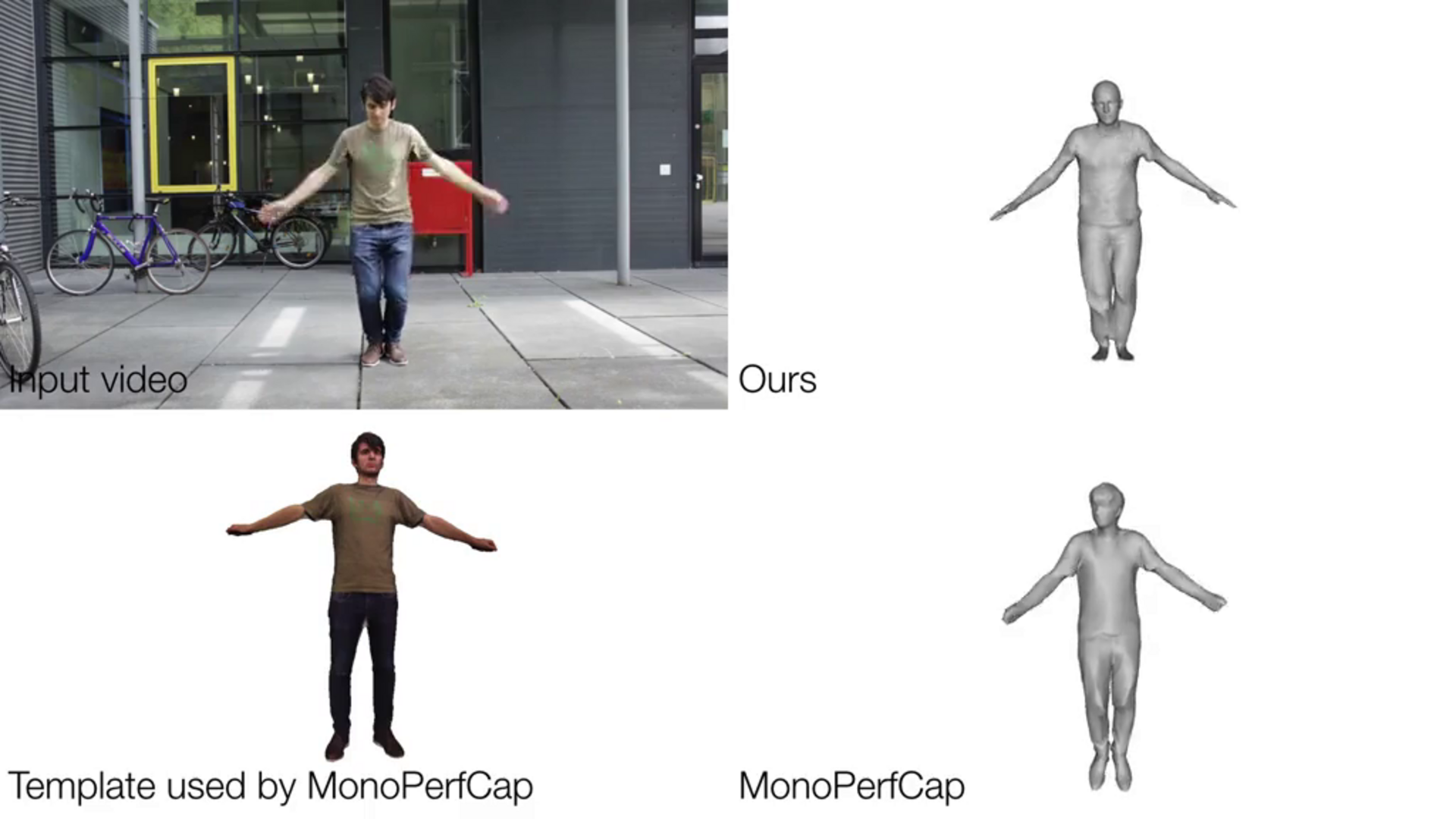}
  \end{subfigure}
  \hspace*{\fill}
 \begin{subfigure}{{\imWidth}\linewidth}
    \includegraphics[trim={\cropMCCL} {\cropMCCB} {\cropMCCR} {\cropMCCT}, clip, width=\linewidth]{images/results/MonoClothCap/Oleks/0018_others.png}
  \end{subfigure}
  \hspace*{\fill}
   \begin{subfigure}{{\imWidth}\linewidth}
    \includegraphics[trim={\cropOursLL} {\cropOursBB} {\cropOursRR} {\cropOursTT}, clip, width=\linewidth]{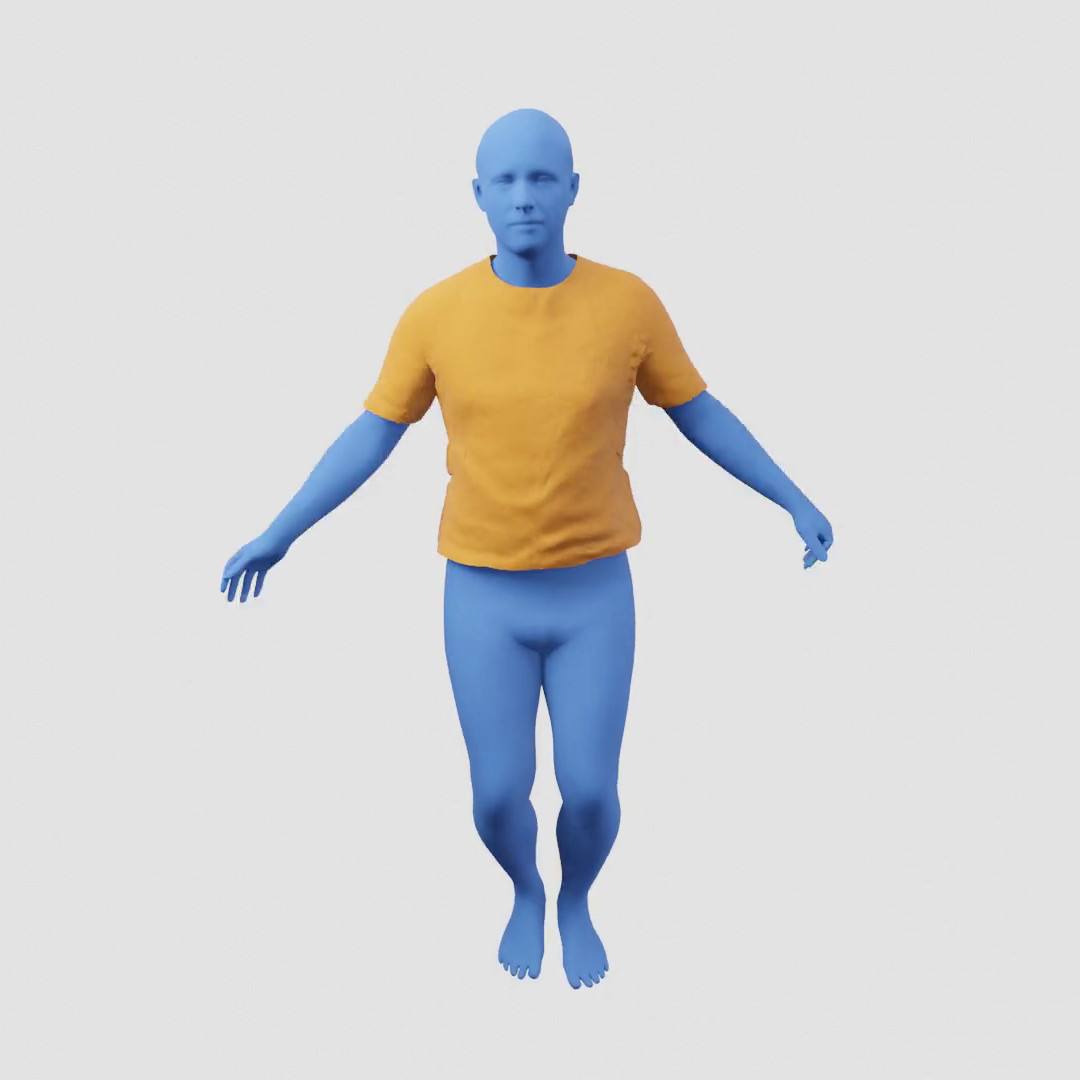}
  \end{subfigure}
 
   \begin{subfigure}{{\imWidth}\linewidth}
    \includegraphics[trim={\cropOriginalL} {\cropOriginalB} {\cropOriginalR} {\cropOriginalT}, clip, width=\linewidth]{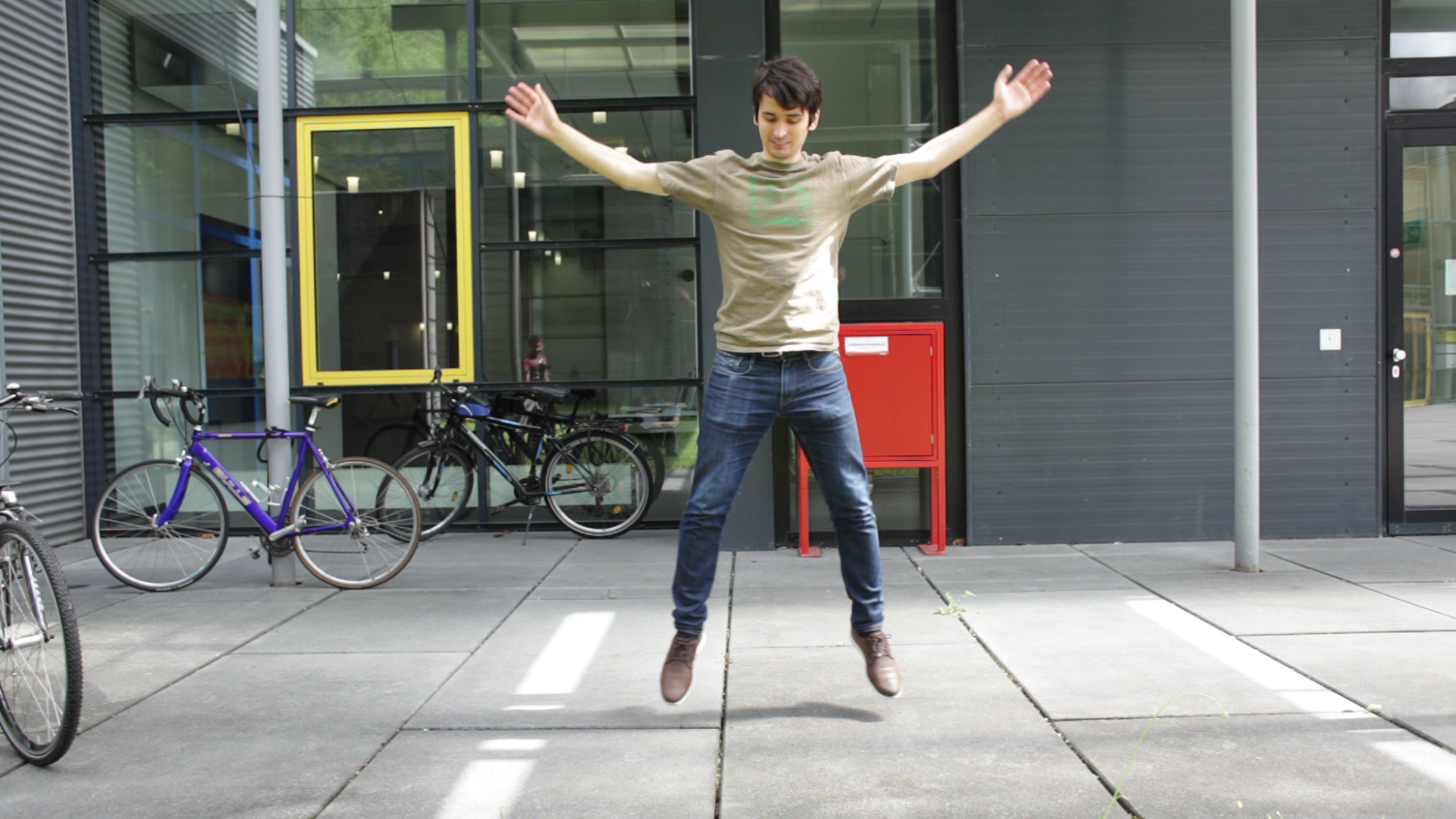}
  \end{subfigure}
  \hspace*{\fill}
 \begin{subfigure}{{\imWidth}\linewidth}
    \includegraphics[trim={\cropMPCL} {\cropMPCB} {\cropMPCR} {\cropMPCT}, clip, width=\linewidth]{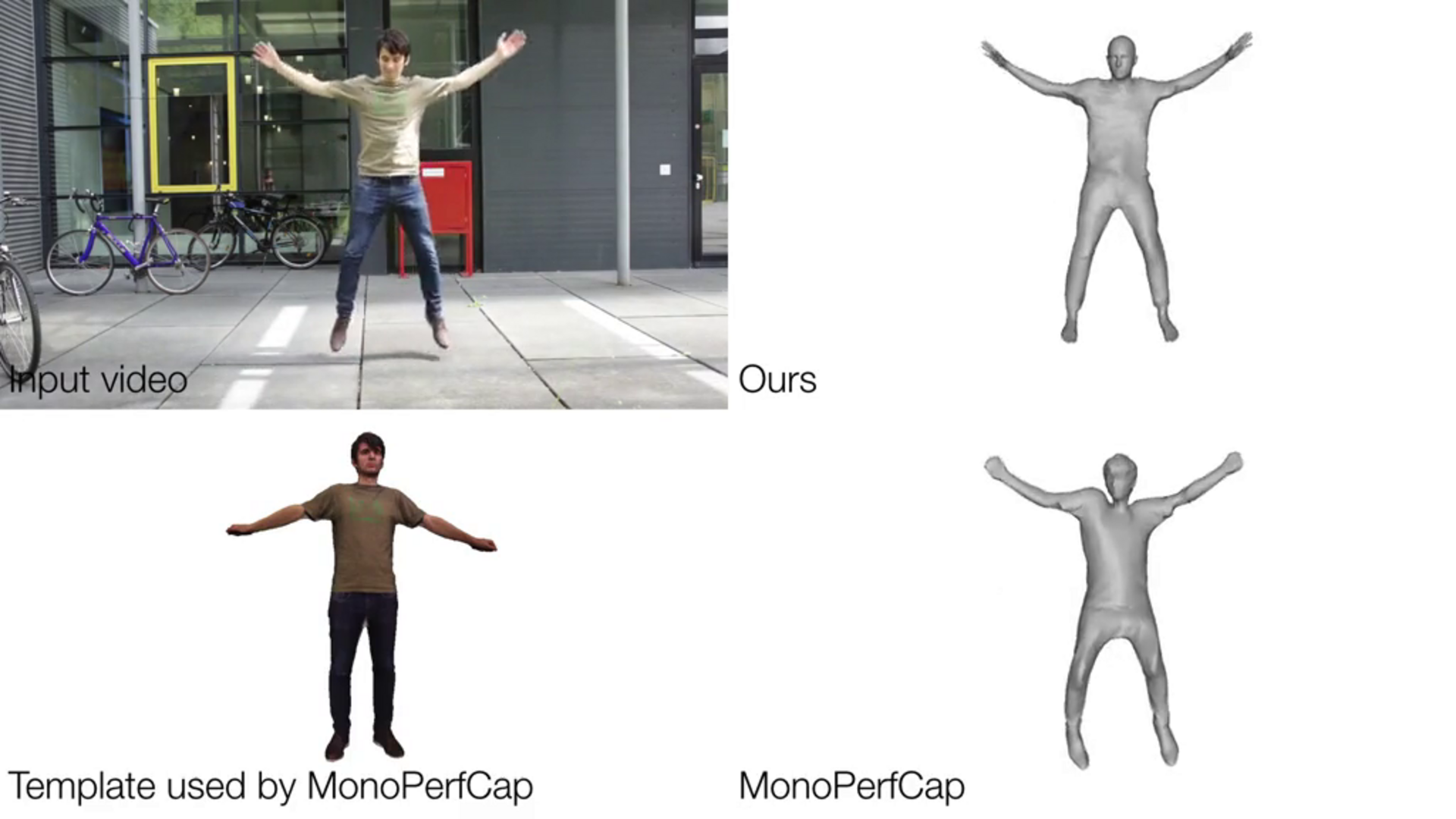}
  \end{subfigure}
  \hspace*{\fill}
 \begin{subfigure}{{\imWidth}\linewidth}
    \includegraphics[trim={\cropMCCL} {\cropMCCB} {\cropMCCR} {\cropMCCT}, clip, width=\linewidth]{images/results/MonoClothCap/Oleks/0111_others.png}
  \end{subfigure}
  \hspace*{\fill}
   \begin{subfigure}{{\imWidth}\linewidth}
    \includegraphics[trim={\cropOursLL} {\cropOursBB} {\cropOursRR} {\cropOursTT}, clip, width=\linewidth]{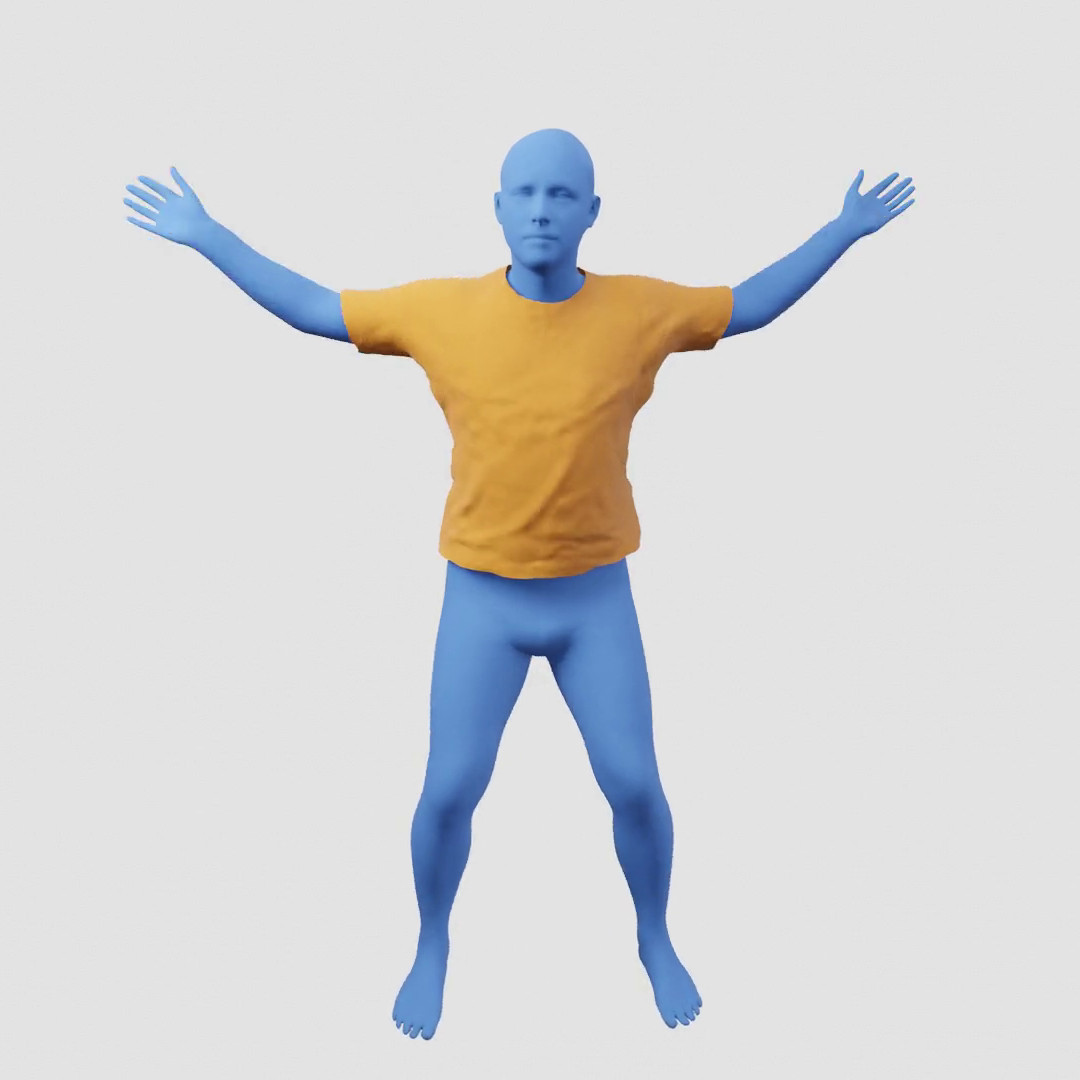}
  \end{subfigure}
  
   \begin{subfigure}{{\imWidth}\linewidth}
    \includegraphics[trim={\cropOriginalL} {\cropOriginalB} {\cropOriginalR} {\cropOriginalT}, clip, width=\linewidth]{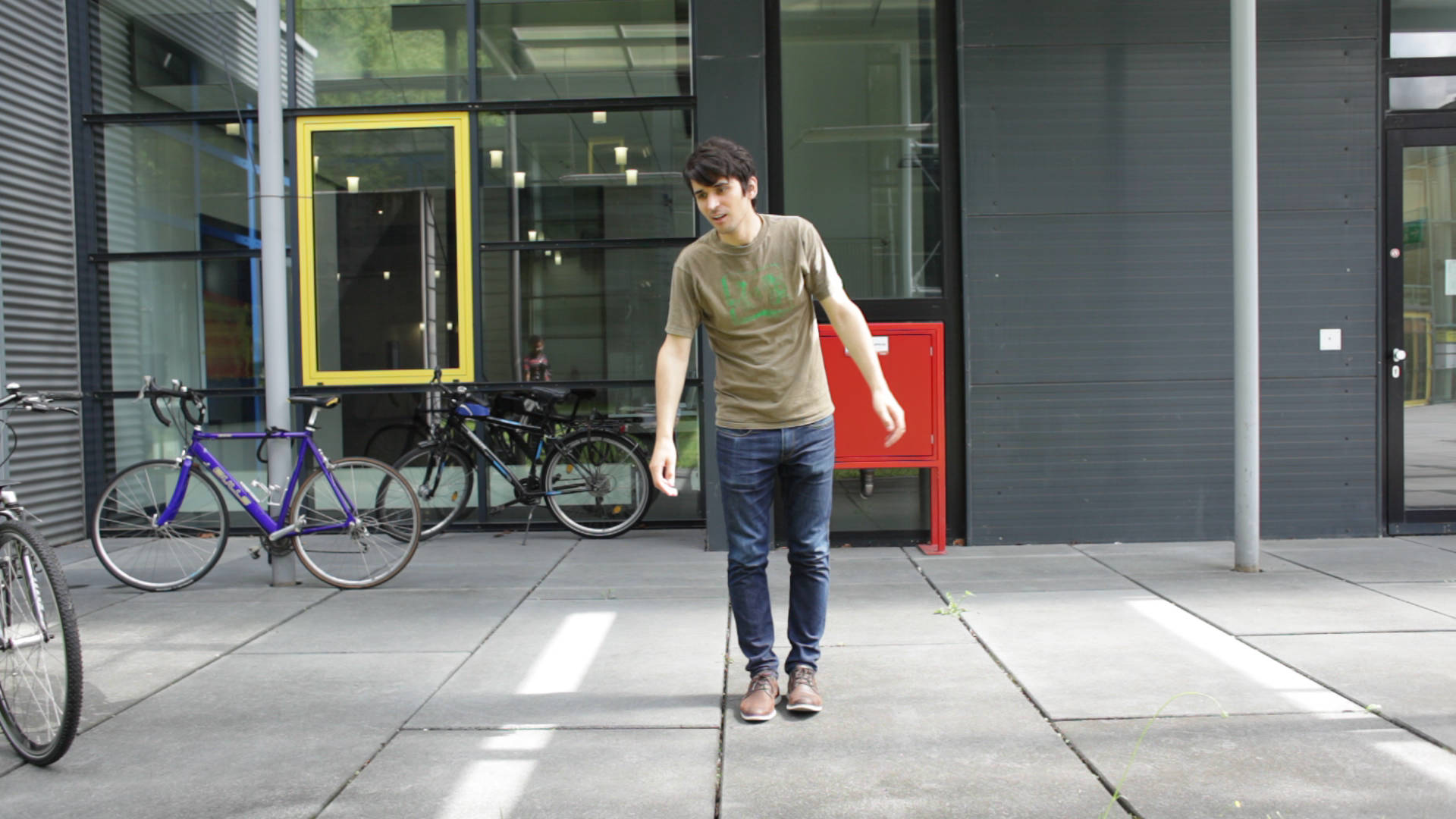}
      \caption*{\scriptsize{Input}}
  \end{subfigure}
  \hspace*{\fill}
 \begin{subfigure}{{\imWidth}\linewidth}
    \includegraphics[trim={\cropMPCL} {\cropMPCB} {\cropMPCR} {\cropMPCT}, clip, width=\linewidth]{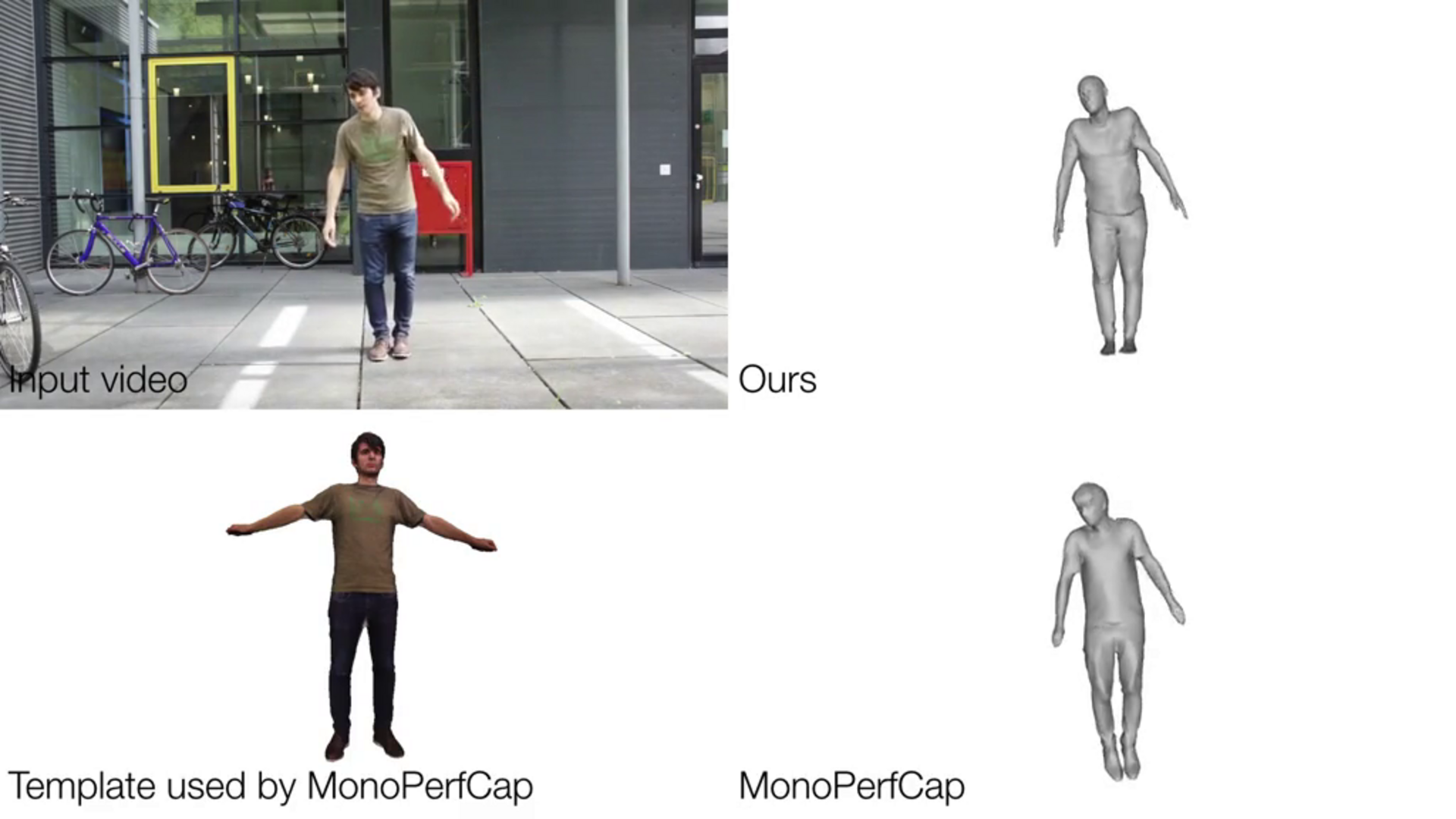}
     \caption*{\scriptsize{MonoPerfCap}}
  \end{subfigure}
  \hspace*{\fill}
 \begin{subfigure}{{\imWidth}\linewidth}
    \includegraphics[trim={\cropMCCL} {\cropMCCB} {\cropMCCR} {\cropMCCT}, clip, width=\linewidth]{images/results/MonoClothCap/Oleks/0194_others.png}
     \caption*{\scriptsize{MonoClothCap}}
  \end{subfigure}
  \hspace*{\fill}
   \begin{subfigure}{{\imWidth}\linewidth}
    \includegraphics[trim={\cropOursLL} {\cropOursBB} {\cropOursRR} {\cropOursTT}, clip, width=\linewidth]{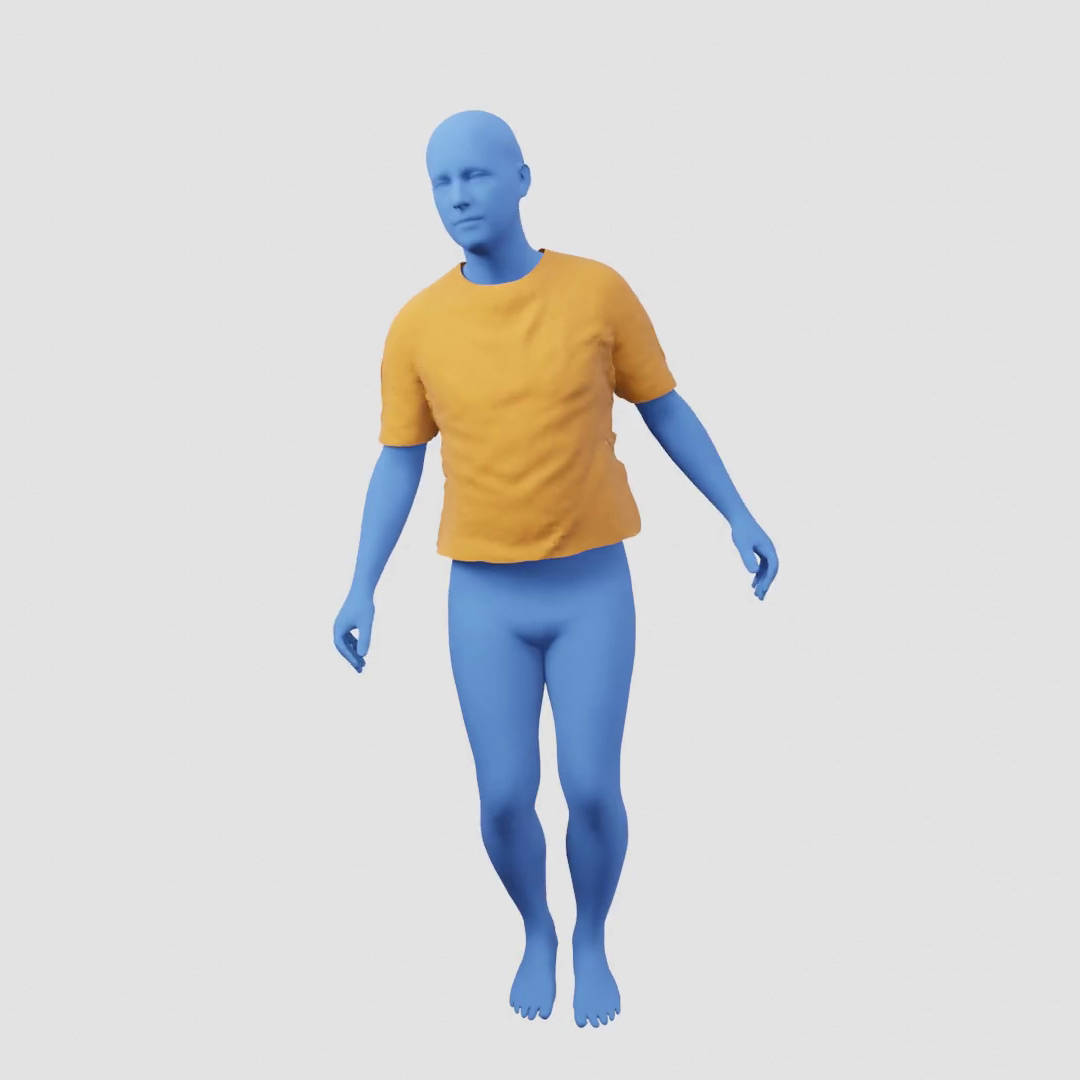}
       \caption*{\scriptsize{PERGAMO (Ours)}}
  \end{subfigure}
  \caption{Qualitative comparison to MonoClothCap \cite{xiang2020monocloth} and MonoPerfCap in \texttt{Oleksander\_outdoor} sequence from \cite{habermann2019livecap}}
  \label{fig:oleks}
\end{figure}
}

%% file: sections/figures/figure-regression.tex
{
\newcommand{\imWidth}{0.1925}
\newcommand{\cropOursL}{70}
\newcommand{\cropOursB}{20}
\newcommand{\cropOursR}{130}
\newcommand{\cropOursT}{0}
\newcommand{\cropOursLL}{19}
\newcommand{\cropOursBB}{12}
\newcommand{\cropOursRR}{23}
\newcommand{\cropOursTT}{0}
\newcommand{\cropMPCL}{1270}
\newcommand{\cropMPCB}{50}
\newcommand{\cropMPCR}{250}
\newcommand{\cropMPCT}{570}
\newcommand{\cropMCCL}{1270}
\newcommand{\cropMCCB}{600}
\newcommand{\cropMCCR}{250}
\newcommand{\cropMCCT}{20}
\newcommand{\cropOriginalL}{600}
\newcommand{\cropOriginalB}{45}
\newcommand{\cropOriginalR}{520}
\newcommand{\cropOriginalT}{70}
\begin{figure*}
 \begin{subfigure}{{\imWidth}\linewidth}
    \includegraphics[trim={\cropOriginalL} {\cropOriginalB} {\cropOriginalR} {\cropOriginalT}, clip, width=\linewidth]{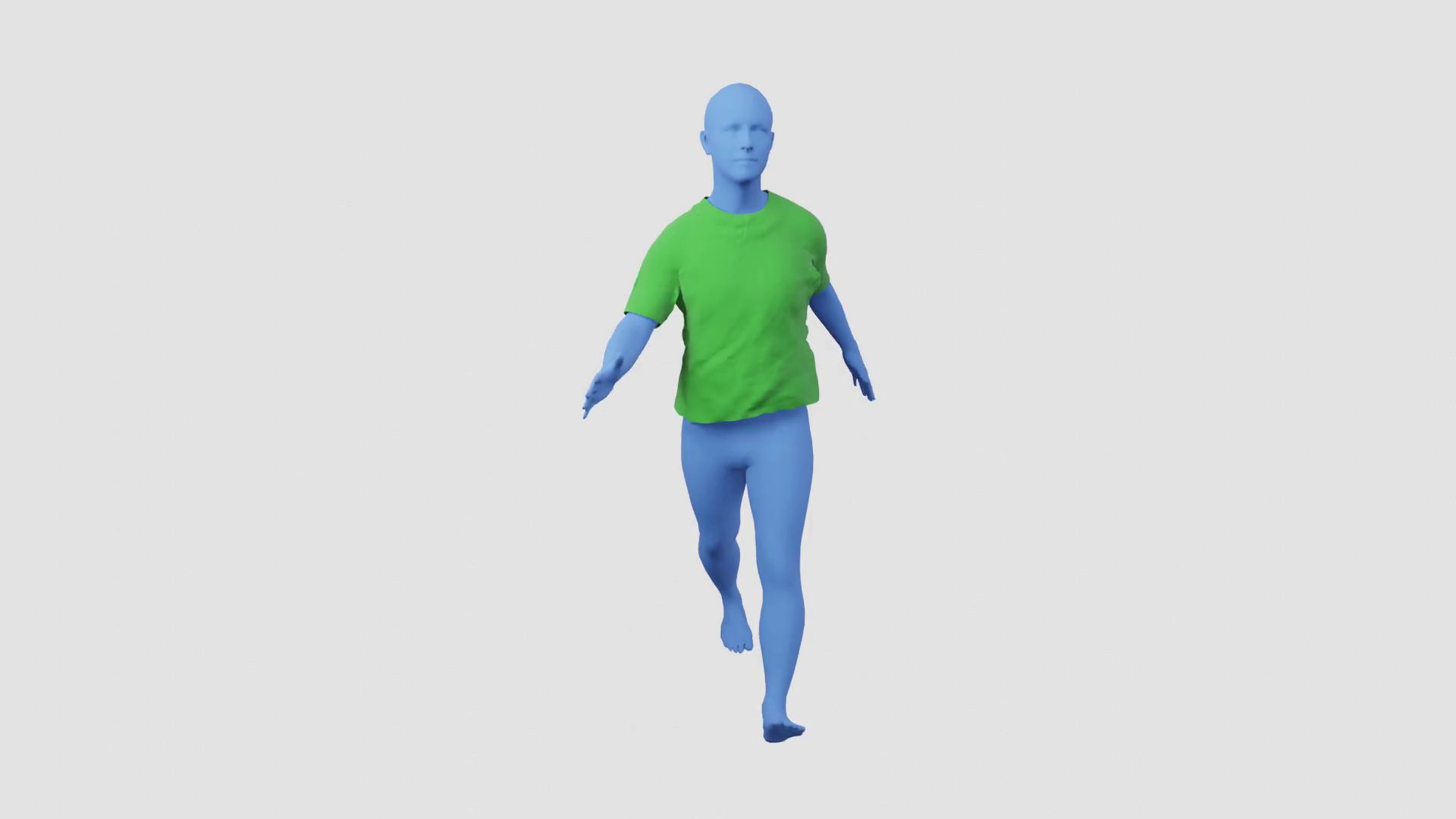}
  \end{subfigure}
  \hspace*{\fill}
 \begin{subfigure}{{\imWidth}\linewidth}
    \includegraphics[trim={\cropOriginalL} {\cropOriginalB} {\cropOriginalR} {\cropOriginalT}, clip, width=\linewidth]{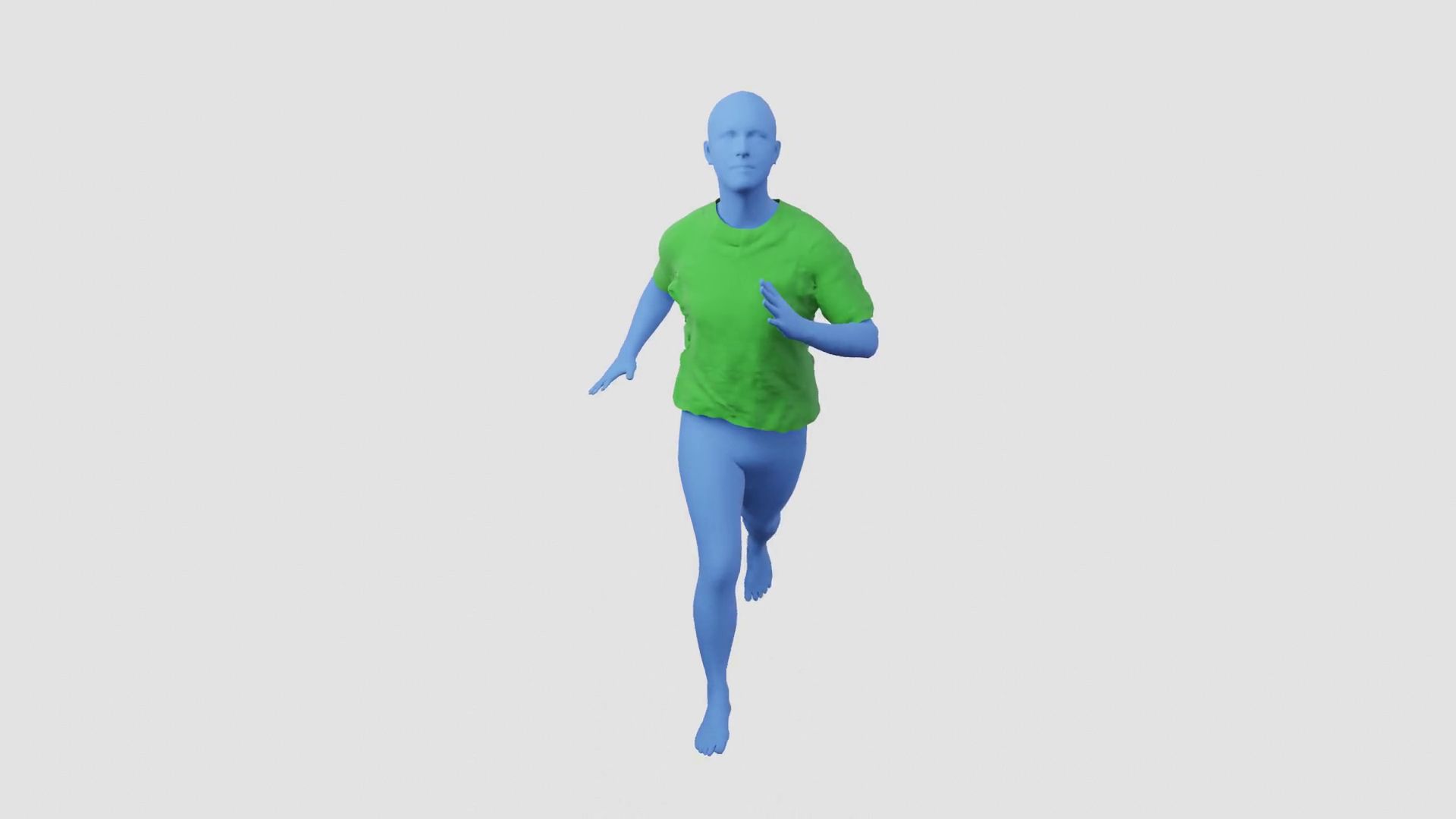}
  \end{subfigure}
 \hspace*{\fill}
   \begin{subfigure}{{\imWidth}\linewidth}
    \includegraphics[trim={\cropOriginalL} {\cropOriginalB} {\cropOriginalR} {\cropOriginalT}, clip, width=\linewidth]{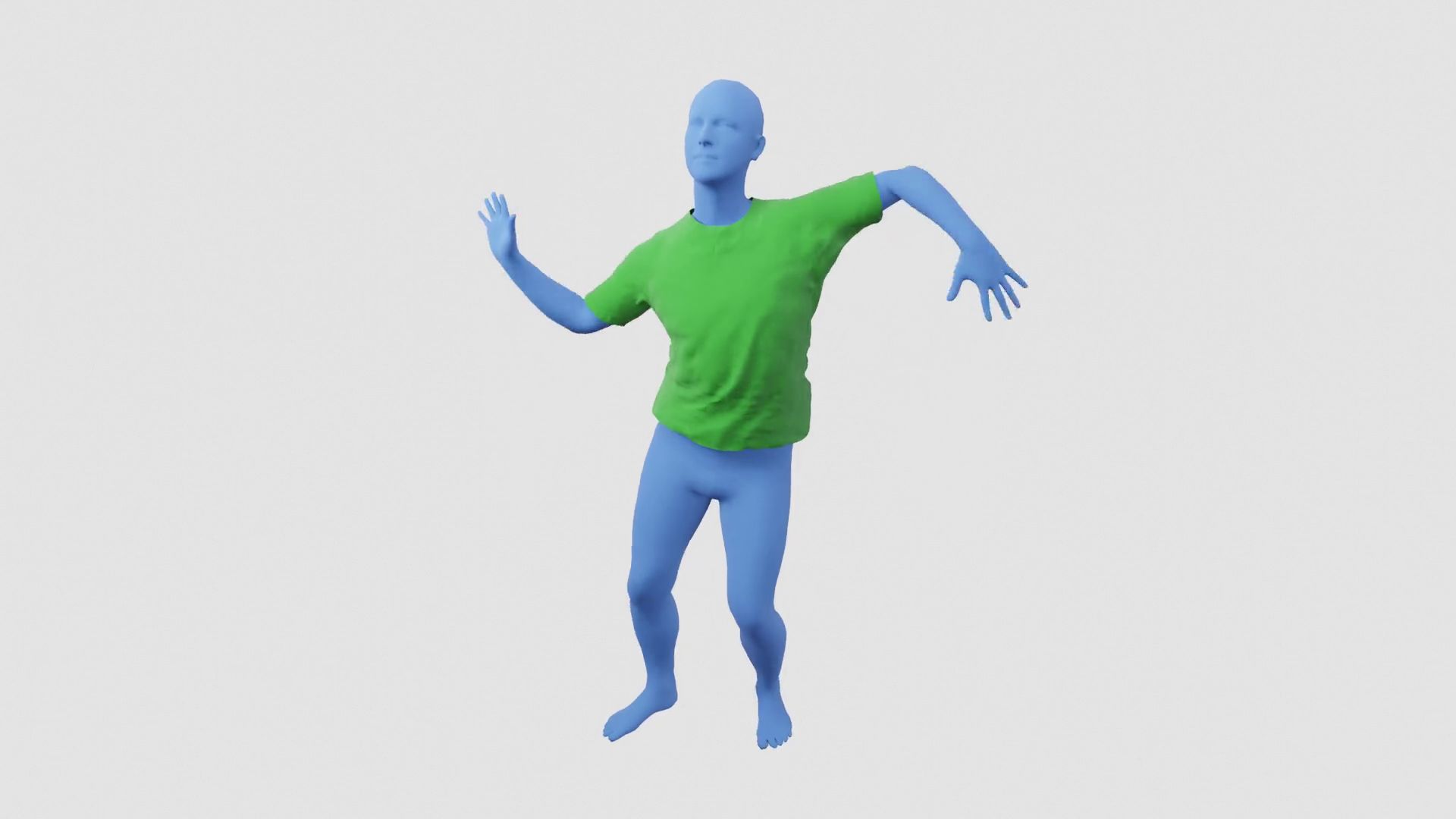}
  \end{subfigure}
 \hspace*{\fill}
  \begin{subfigure}{{\imWidth}\linewidth}
    \includegraphics[trim={\cropOriginalL} {\cropOriginalB} {\cropOriginalR} {\cropOriginalT}, clip, width=\linewidth]{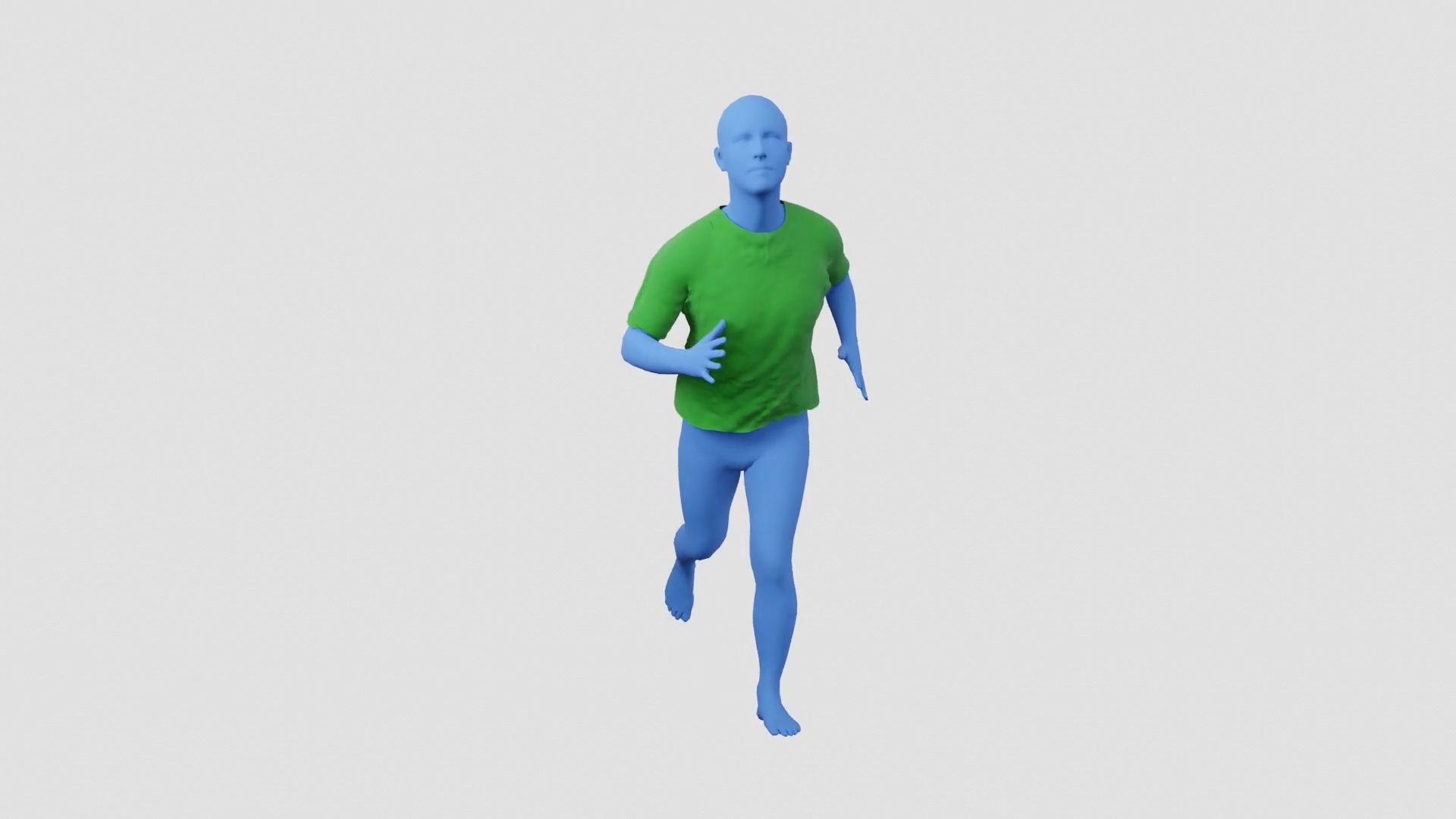}
  \end{subfigure}
  \hspace*{\fill}
 \begin{subfigure}{{\imWidth}\linewidth}
    \includegraphics[trim=400 66 380 46, clip, width=\linewidth]{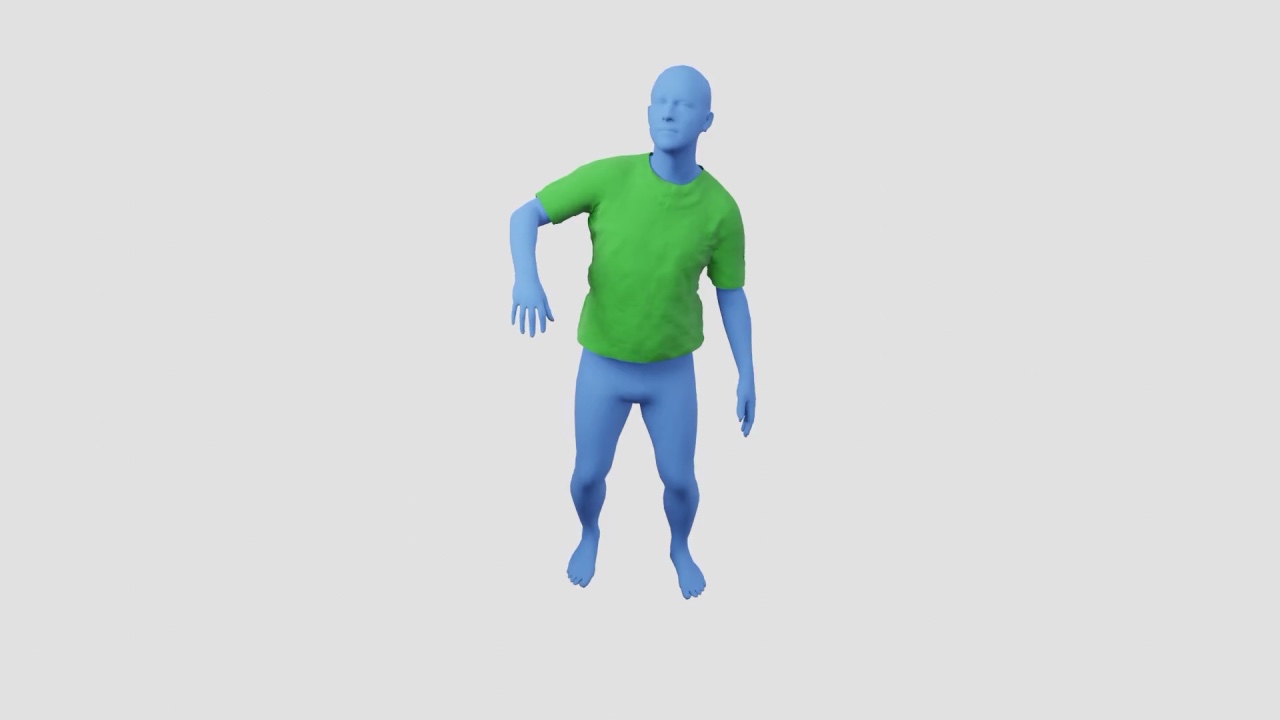}
  \end{subfigure}
  \caption{Garment regression results for \textit{test} motion sequences unseen at training time. PERGAMO is capable of inferring dynamic 3D garment details learning from just the monocular videos. Specifically, here we show test poses from AMASS \cite{AMASS:ICCV:2019} dataset sequences \texttt{08\_01}, \texttt{09\_01}, \texttt{S\_6\_F\_7}, \texttt{128\_02}, and BUFF \cite{zhang2017detailed}. See supplementary video for animated version of this figure.}
  \label{fig:qualitative-regression}
\end{figure*}
}

%% file: sections/conclussion.tex
\section{Conclusions}
We have introduced PERGAMO, a novel approach to learn a deformable model for 3D garments directly from monocular videos.
To the best of our knowledge, PERGAMO is the first method to reconstruct an explicit 3D garment layer from single view, \textit{and} use the reconstructed geometry to learn a deformable model.
Since PERGAMO uses training data that comes directly from real-world images, it circumvents the simulation-to-real gap issue that existing data-driven methods suffer.
Comparisons to existing methods demonstrate that our approach is capable of recovering finer wrinkle detail, and it generalizes well to unseen motions at train time.

\begin{figure}
    \centering
    \includegraphics[width=0.20\textwidth]{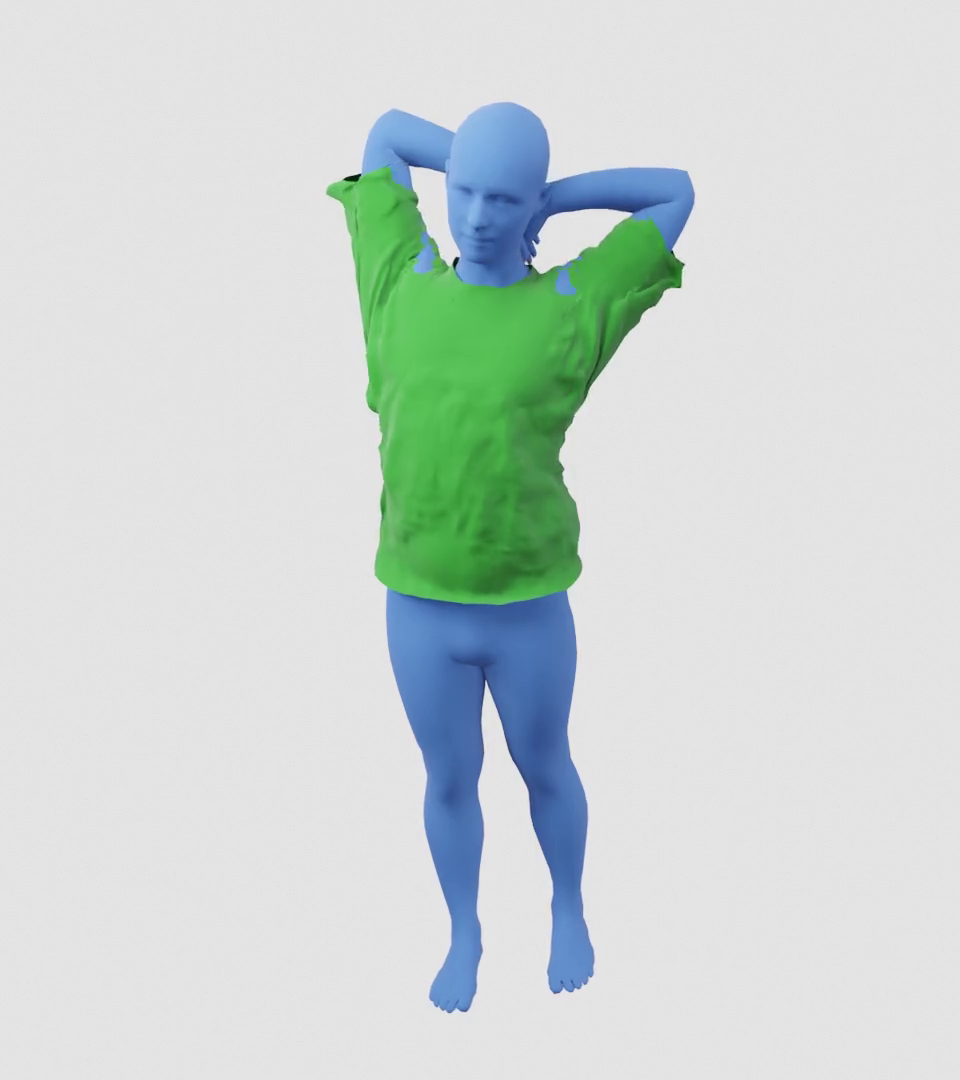}
    \includegraphics[width=0.20\textwidth]{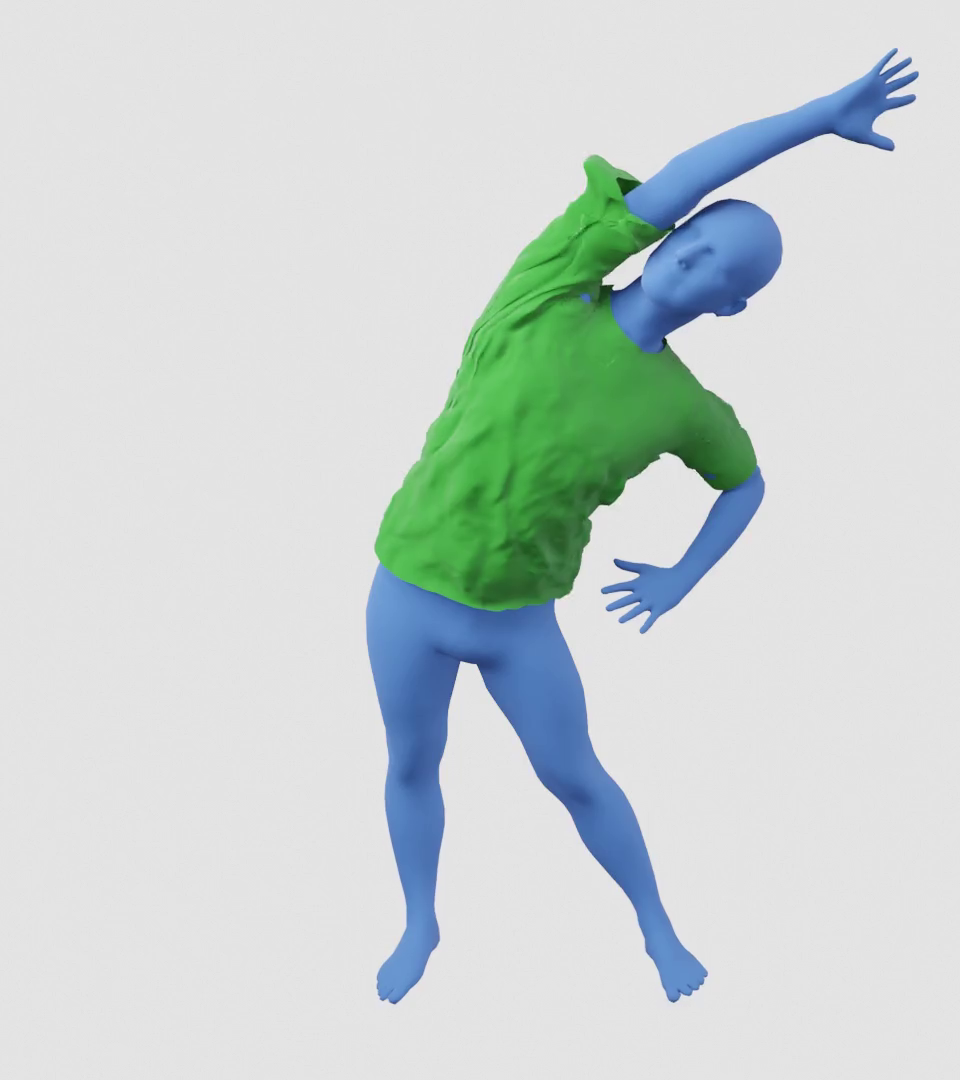}
    \caption{Examples of failure cases on extreme body poses.}
    \label{fig:failureCases}
\end{figure}

Despite the step forward that PERGAMO does in the field of 3D garment modeling and animation, it still suffer from a few limitations.
\Change{First, garment self-collisions are not explicitly modeled, therefore, despite that we regularize 3D deformations to avoid geometry artifacts, residual collisions can occur in highly deformed areas such as armpits.}
\Change{Second}, our approach is only capable of reconstruction garments that are close to the coarse deformable mesh used in our reconstructing step.
It remains open to future research how to generalize the reconstruction pipeline to a larger variety of clothing.
Additionally, at the moment we do not model body shape variations, which prevents PERGAMO to generalize to unseen subjects.
\Change{Third, our method is dependent on the quality of the information used as input. Noisy estimated normals, pose or segmentation can impact the quality of the reconstructed garments.}
\Change{And fourth, the regressor may produce unsatisfactory results when dealing with extreme poses as can be seen in Figure~\ref{fig:failureCases}.}
\section*{Acknowledgments}
This work has been partially funded by: the Comunidad de Madrid in the framework of the Multiannual Agreement with the Universidad Rey Juan Carlos in line of Action 1, "Encouragement of research for young PhD", project CaptHuRe (M2736); the Universidad Rey Juan Carlos through the Distinguished Researcher position INVESDIST-04 under the call from 17/12/2020; and by a Leonardo Fellowship from the Fundaci\'on BBVA.

\vspace{5mm}

\includegraphics[width=0.18\textwidth]{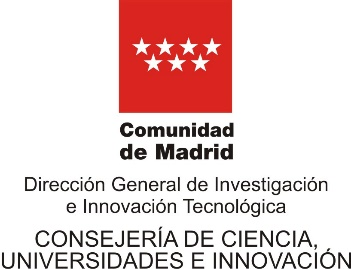}
\includegraphics[width=0.18\textwidth]{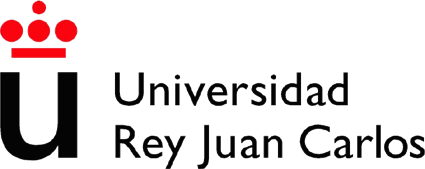}